\documentclass[11pt,letterpaper]{article}
\usepackage[letterpaper,margin=1in]{geometry}
\usepackage[utf8]{inputenc}
\IfFileExists{lmodern.sty}{\usepackage[T1]{fontenc}\usepackage{lmodern}}{}  %
\usepackage[expansion=false]{microtype}
\usepackage{amsmath,amssymb,amsthm,mathtools}
\usepackage{enumitem}
\usepackage{xcolor}
\usepackage[framemethod=TikZ]{mdframed}
\usepackage{tikz,pgfplots}
\usetikzlibrary{arrows.meta,positioning,decorations.pathreplacing,calc}
\pgfplotsset{compat=1.17}
\usepackage[numbers,sort&compress]{natbib}
\usepackage[bookmarksnumbered,bookmarksopen,bookmarksopenlevel=1,colorlinks=true,linkcolor=blue!55!black,citecolor=green!35!black,urlcolor=blue!55!black]{hyperref}

\allowdisplaybreaks
\setlist{nosep,leftmargin=1.6em}

\theoremstyle{plain}
\newtheorem{theorem}{Theorem}[section]
\newtheorem{lemma}[theorem]{Lemma}
\newtheorem{proposition}[theorem]{Proposition}
\newtheorem{corollary}[theorem]{Corollary}

\theoremstyle{definition}
\newtheorem{definition}[theorem]{Definition}
\theoremstyle{remark}
\newtheorem{remark}[theorem]{Remark}
\newtheorem*{remark*}{Remark}

\newmdenv[linecolor=blue!55!black,backgroundcolor=blue!5,linewidth=0.8pt,
  skipabove=8pt,skipbelow=8pt,innertopmargin=2pt,innerbottommargin=6pt,
  innerleftmargin=6pt,innerrightmargin=6pt]{mainresult}

\newif\ifdraft\drafttrue
\ifdraft
\newmdenv[linecolor=red!70!black,backgroundcolor=red!4,linewidth=0.8pt,
  skipabove=6pt,skipbelow=6pt,innertopmargin=4pt,innerbottommargin=4pt]{todobox}
\newcommand{\todo}[1]{\begin{todobox}\small\textbf{\textcolor{red!70!black}{To do.}} #1\end{todobox}}
\else
\newcommand{\todo}[1]{}
\fi

\newmdenv[topline=false,bottomline=false,rightline=false,
  linecolor=blue!55!black,linewidth=2pt,backgroundcolor=blue!4,
  innerleftmargin=10pt,innerrightmargin=10pt,innertopmargin=7pt,innerbottommargin=7pt,
  skipabove=10pt,skipbelow=10pt]{highlightbox}

\newcommand{\R}{\mathbb R}
\newcommand{\C}{\mathbb C}

\newcommand{\E}{\mathbb E}
\newcommand{\Rea}{\operatorname{Re}}
\newcommand{\Ima}{\operatorname{Im}}
\DeclareMathOperator{\Var}{Var}
\DeclareMathOperator{\Reg}{Reg}
\DeclareMathOperator{\sech}{sech}
\DeclareMathOperator{\osc}{osc}
\DeclareMathOperator{\KL}{KL}
\newcommand{\norm}[1]{\lVert#1\rVert}

\newcommand{\ip}[2]{\langle #1,#2\rangle}

\newcommand{\cG}{\mathcal G}
\newcommand{\cI}{\mathcal I}

\newcommand{\cP}{\mathcal P}
\newcommand{\cK}{\mathcal K}

\newcommand{\loss}{\boldsymbol\ell}
\newcommand{\simplex}{\Delta}
\newcommand{\tilt}[2]{\mathsf E_{#1}[#2]}
\newcommand{\eps}{\varepsilon}
\newcommand{\mem}{\varrho}

\title{Optimistic Hedge in General-Sum Games:\\
Fundamental Limits and Regret Bounds\\ via Temporal Approximation}
\author{%
  \begin{tabular}[t]{@{}c@{\hspace{2.2em}}c@{\hspace{2.2em}}c@{\hspace{2.2em}}c@{}}
   Yuval Dagan\thanks{School of Computer Science and AI, Tel Aviv University, Tel Aviv, Israel.} & Nadav Eisen\footnotemark[1] & Liad Erez\footnotemark[1] & Tomer Koren\footnotemark[1]~\thanks{Google Research, Tel Aviv, Israel.}
  \end{tabular}}
\date{}

\begin{document}

\pagenumbering{roman}
\maketitle
\thispagestyle{empty}
\begin{abstract}
\noindent
Optimistic Hedge, a multiplicative weight update with one step of extrapolation, is one of the most natural regret-minimizing dynamics for games. Nevertheless, the characterization of its regret has been a longstanding open question which has been a focus of considerable attention. In recent years, Daskalakis, Fishelson and Golowich \cite{DFG21} made significant progress: they proved that if all $m$ players in a finite general sum game follow Optimistic Hedge with a fixed learning rate, each player enjoys regret which is \emph{polylogarithmic} in the horizon $T$. Subsequently, Ha \cite{Ha2026constant} showed that in the same setting, an appropriate choice of learning rate results in each player's regret being bounded \emph{independently of} $T$. This bound, however, arises from a non-constructive argument, and its dependence on the number of actions $n$ is not explicit in the proof, and thus still leaves open the question of whether Optimistic Hedge can achieve a constant regret bound which is polynomial in the size of the game. 

We show that the answer to this question hinges on the number of players. In two-player games, there is a learning rate for which Optimistic Hedge guarantees $O(n\log n)$ regret independent of the horizon. In three-player games, however, we show that no such bound is possible: for any learning rate up to an absolute constant and for sufficiently large horizon $T$, there is a game with $n$ actions per player in which some player's regret is $2^{\Omega(n)}$, and more generally, for every $m\ge3$ we construct $m$-player games in which some player's regret is $2^{\Omega(mn)}$; and for learning rates above $2^{-\Omega(mn)}$, that player's regret even grows linearly in $T$. 

The proof of our upper bound is based on a temporal approximation framework we develop for the analysis of Optimistic Hedge. Starting from the variance-based RVU bound of \cite{DFG21}, we approximate the losses over a time-localized window by functions which are well-behaved, in the sense that their total variation is bounded by a small fraction of their cumulative variance. We also instantiate this framework in $m$-player games and obtain a significantly simplified analysis of Optimistic Hedge, matching the recent $O(\sqrt m\,\log n\,\log T)$ regret bound due to Ha \cite{Ha2026log}.
\end{abstract}

\newpage
\ifdraft
\thispagestyle{empty}
\setcounter{tocdepth}{2}
\tableofcontents
\newpage
\fi
\pagenumbering{arabic}

\section{Introduction}\label{sec:intro}

No-regret learning is a central paradigm for decision making under uncertainty and for learning in games. In each of $T$ rounds, a learner chooses a probability distribution $x^t$ over $n$ actions, observes a loss vector $\ell^t\in[0,1]^n$, and suffers the expected loss $\ip{x^t}{\ell^t}$. The learner's \emph{regret},
$
 \Reg(T)\;=\;\sum_{t=1}^T\ip{x^t}{\ell^t}-\min_{a\in[n]}\sum_{t=1}^T\ell^t(a),
$
compares their cumulative loss with that of the best fixed action in hindsight. The canonical no-regret algorithm is multiplicative weights, or Hedge~\cite{LW94,FS97}, which plays $x^t(a)\propto\exp(-\eta\sum_{s<t}\ell^s(a))$ and, for a learning rate $\eta\simeq\sqrt{\log n/T}$, guarantees $\Reg(T)=O(\sqrt{T\log n})$ against every sequence of losses. Algorithms of this kind underlie boosting~\cite{FS97}, fast approximate solvers for zero-sum games, linear programs and combinatorial problems~\cite{FS99,AHK12}, and counterfactual regret minimization, the engine behind superhuman poker programs~\cite{ZJBP07,BS18}; they also model bidder behavior in ad auctions~\cite{NST15}. In games, when every player has regret $o(T)$, the empirical distribution of play converges to the set of \emph{coarse correlated equilibria}~\cite{Hannan57,MV78}, at a rate governed by the largest average regret among the players; under the stronger requirement of vanishing internal regret, it converges to correlated equilibria~\cite{FV97,HM00}.

Against an \emph{adversary}, who may choose every loss vector with full knowledge of the learner's algorithm, the rate $\sqrt{T\log n}$ cannot be improved~\cite{CBL06}. A player in a game, however, faces other learners rather than an adversary. If they adapt gradually, the loss vector $\ell^{t+1}$ resembles $\ell^t$, and anticipating this may yield regret far below the adversarial rate. \emph{Optimistic Hedge}~\cite{RS13colt,RS13,SALS15} builds this anticipation into Hedge by counting the most recent loss vector twice, the second time as a prediction of the next one. Suitably tuned, it retains the adversarial guarantee of Hedge and requires no knowledge of the game and no coordination beyond a shared learning rate.

A long line of work has studied the regret of Optimistic Hedge when all $m$ players of a general-sum game, each with $n$ actions, run it with a common, fixed learning rate. Its starting point is the RVU (regret bounded by variation in utilities) property of Syrgkanis, Agarwal, Luo and Schapire~\cite{SALS15}, which bounds the regret by $\log n/\eta$, plus a term proportional to the variation of the losses, minus one proportional to the movement of the player's own strategy. Since strategies move by $O(\eta)$ per round, this can be shown to yield regret $O(T^{1/4})$. The exponent was lowered to $1/6$ for two players by Chen and Peng~\cite{CP20}, and the first polylogarithmic bound, $O(m\log n\log^4T)$, was obtained by Daskalakis, Fishelson and Golowich~\cite{DFG21}. Ha~\cite{Ha2026log} recently improved it to $O(\sqrt m\,\log n\,\log T)$, and later showed that, for a suitable learning rate, the regret is bounded by a constant $C(m,n)$ independent of $T$~\cite{Ha2026constant}.

This development settles the dependence of the regret on the horizon, but not on the size of the game. The proof of~\cite{Ha2026constant} is non-constructive: it derives an exact finite-order linear relation among the temporal differences of the trajectory from a Noetherianity theorem of Frisch~\cite{Frisch67}. This shows that $C(m,n)$ is finite, but gives no control over its dependence on $m$ and $n$. Were $C(m,n)$ exponential in $n$, the constant bound would be weaker than the logarithmic one at every horizon of practical interest. Meanwhile, more elaborate dynamics, based on higher-order or multiplicative forms of optimism, were recently shown to attain horizon-independent regret polynomial in $m$ and $n$, and even polylogarithmic in $n$ for external regret~\cite{LFO26,ALM26,SP26,Mai26}. As Optimistic Hedge remains the most natural and best studied of the optimistic dynamics, we ask whether its simple one-step optimism already suffices:
\begin{center}
 \emph{Does fixed-learning-rate Optimistic Hedge guarantee every player\\ regret $\mathrm{poly}(m,n)$, independently of the horizon?}
\end{center}
We show that, perhaps surprisingly, the answer is affirmative for two players and negative for three or more, although the polylogarithmic bounds above depend on $m$ only polynomially.

\subsection{Main results}\label{sec:results}

Throughout, all players run Optimistic Hedge with a common, fixed learning rate $\eta$, starting from the uniform distribution and observing their expected loss vectors. We obtain three results.
\begin{enumerate}[label=(\roman*),topsep=\medskipamount]
 \item \emph{Logarithmic regret for $m$ players.} We develop a temporal approximation framework for the analysis of Optimistic Hedge, and use it to reprove the logarithmic bound of Ha~\cite{Ha2026log}, with the same dependence on $m$ and $n$ and with explicit constants: in every $m$-player game, the learning rate $\eta=c/(\sqrt m\log T)$, for a small enough absolute constant $c>0$, guarantees each player regret $O(\sqrt m\,\log n\,\log T)$ (Theorem~\ref{thm:log}).
 \item \emph{Constant regret for two players.} There is an absolute constant $c>0$ such that in every two-player game, any learning rate $0<\eta\le c/n$ guarantees each player regret at most $\log n/\eta+\eta$ at every horizon $T$. In particular, $\eta=c/n$ gives regret $O(n\log n)$ for all $T$ (Theorem~\ref{thm:two}).
 \item \emph{An exponential barrier for three players.} There are absolute constants $C,\eta_0>0$ such that for every $n\ge C$ and every learning rate $0<\eta\le\eta_0$, some three-player game with $n$ actions per player forces regret $2^{\Omega(n)}$ on some player for all sufficiently large $T$. If moreover $\eta\ge Cn2^{-n/8}$, the game can be chosen so that this player's regret grows linearly in $T$ (Theorem~\ref{thm:lower}). More generally, for $m\ge3$ players the lower bound becomes $2^{\Omega(mn)}$ (Theorem~\ref{thm:lower-many}).
\end{enumerate}
Our proof of~(i) is short and significantly simplifies earlier polylogarithmic analyses of Optimistic Hedge~\cite{DFG21,Ha2026log}. The bound in~(ii) is explicit and polynomial in $n$, and the learning rate $\eta=c/n$ is oblivious to both the game and the horizon. It is also tight up to an additive constant, since even against constant losses the regret of Optimistic Hedge can reach $\log n/\eta-1$. The same elementary fact gives the first part of~(iii) when $\eta<Cn2^{-n/8}$, so the substance of~(iii) lies in its second part. Already for $\eta=c/n$, some three-player game forces linear regret once $n$ is large, so adding a single player turns a constant regret bound into a linear one. To our knowledge, (iii) is the first lower bound for Optimistic Hedge beyond two-player zero-sum games, and it separates the algorithm from the more elaborate dynamics mentioned above.

\paragraph{Techniques.}
Through the variance form of the RVU bound~\cite{SALS15,DFG21}, both upper bounds reduce to bounding the variation of a player's losses by their variance: along the trajectory of play, the variation ${\sum_t\Var_{x^t}(\ell^t-\ell^{t-1})}$ must be at most half the cumulative variance $\sum_t\Var_{x^t}(\ell^{t-1})$, up to an additive term. Where previous work relied on the high-order smoothness of the trajectory~\cite{DFG21,Ha2026log}, we approximate the losses on short windows of time by simple functions for which such an inequality is classical. Inequalities of Markov and Bernstein type bound the derivative of a polynomial or of an exponential sum by the size of the function itself; in the integral form relevant here, every $f$ in the approximating class $\mathcal F$ satisfies
\[
 \int_I f'(s)^2\,ds\;\le\;\frac{C_{\mathcal F}}{W^2}\int_0^W f(s)^2\,ds,
\]
where $I$ is the middle part of the interval $[0,W]$. We use a discrete analogue of this inequality on windows of $W$ rounds. For $m$ players, $\mathcal F$ consists of polynomials of degree $O(\log T)$, with $C_{\mathcal F}=O(\log^2T)$, and the approximation error is paid for by the horizon. For two players, freezing the losses of player~2 on a window turns those of player~1 into ratios of sums of $n$ exponentials, for which a Bernstein inequality of Borwein and Erd\'elyi~\cite{BE96} gives $C_{\mathcal F}=O(n^2)$. Crucially, the error of freezing is paid for by the variance itself, through a feedback loop: the losses of player~2 move only when player~1 moves, and player~1 moves only when their own losses vary. A third player breaks this loop, and~(iii) shows that the breakdown is not an artifact of the analysis.

The lower bound rests on the observation that optimism can also hurt. To leading order, a small periodic perturbation of frequency $\omega$ in a player's losses changes the slope of their regret in proportion to $\frac12-\cos\omega$. Slow oscillations are anticipated by the optimistic prediction and reduce the regret, whereas fast ones throw the prediction out of phase and make the regret grow linearly. In our game, a third player observes two players driven into a cycle of period four. Since a small learning rate only produces slow oscillations, this cycle is generated by a cascade of blocks of eight actions, each driven quadratically by its predecessor and hence oscillating twice as fast; the $\log(1/\eta)$ blocks this requires account for the exponential barrier.

\subsection{Related work}\label{sec:related}
\begingroup
\makeatletter
\renewcommand\paragraph{\@startsection{paragraph}{4}{\z@}{1ex \@plus.5ex \@minus.2ex}{-1em}{\normalfont\normalsize\bfseries}}
\makeatother

\paragraph{Optimism.}
The optimistic step goes back to Popov's modification of the saddle-point method of Arrow and Hurwicz~\cite{Popov80}, now viewed as a single-call variant of the extragradient method and as an explicit approximation of the proximal point method~\cite{Martinet70,Rockafellar76,MOP20}. Rakhlin and Sridharan~\cite{RS13colt,RS13} brought it to online learning, building on gradual-variation bounds~\cite{CYLMLJZ12}; with the entropic regularizer, it yields Optimistic Hedge.

\paragraph{Optimistic Hedge in games.}
The analysis of~\cite{DFG21} shows that the finite differences of the dynamics of every order $k$ up to $O(\log T)$ are of order $(m\eta)^k$, up to $k$-dependent factors, and converts them into control of the first-order differences through a Fourier-analytic interpolation inequality. The same scheme underlies the bound of~\cite{Ha2026log} and has been extended to swap regret, and hence to correlated equilibria~\cite{ADFFGS22}. More is known in two-player zero-sum games, where near-optimal no-regret dynamics were first designed~\cite{DDK11}: the sum of the two regrets is bounded by a constant~\cite{RS13}, and Optimistic Hedge's individual regret is logarithmic in the numbers of actions, with algorithm-specific lower bounds tight for the sum of the regrets~\cite{Tsu25}.

\paragraph{Other dynamics.}
Logarithmic regret is also achieved by optimistic follow-the-regularized-leader with logarithmic regularizers~\cite{AFKLLS22,FAL22} and by cautious optimism, which adapts the learning rate~\cite{SPF25stoc}, while clairvoyant multiplicative weights approach coarse correlated equilibria at rate $O(\log T/T)$~\cite{PSS22}. The horizon-independent bounds discussed above rely on higher-order predictors built from exponential moving averages~\cite{LFO26} or on a discounted $(m{+}1)$-th order extrapolation over a lifted strategy space~\cite{ALM26}, on multiplicatively optimistic regret matching~\cite{SP26}, and, for swap regret, on two-scale higher-order optimism~\cite{Mai26}.

\paragraph{Cycles.}
Multiplicative weights and related dynamics are known to cycle, or even diverge, in zero-sum games~\cite{MPP18,BP18}, where optimistic methods instead converge in the last iterate when the equilibrium is unique~\cite{DP19,WLZL21}, though not necessarily fast~\cite{cai2024fast}. Our construction shows that in general-sum games Optimistic Hedge itself may cycle, at a high cost in regret.

\endgroup

\section{Problem setting}\label{sec:setting}

\paragraph{Games.}
A finite (general sum) game with $m\ge2$ players is specified by action sets $[n_i]=\{1,\dots,n_i\}$ and loss functions $L_i:\prod_{j=1}^m[n_j]\to[0,1]$, one for each player $i\in[m]$. A mixed strategy of player $i$ is a probability vector $x_i\in\simplex_{n_i}$, and a profile $x=(x_1,\dots,x_m)$ of mixed strategies induces for each player the vector of expected losses of their actions,
\begin{equation}\label{eq:lossmap}
 \loss_i(x)(a_i)\;=\;\sum_{a_{-i}}L_i(a_i,a_{-i})\prod_{j\ne i}x_j(a_j),\qquad a_i\in[n_i],
\end{equation}
which is multilinear in the opponents' strategies. We write $\cG(m,n)$ for the class of games with $m$ players and $n$ actions per player.

\paragraph{Repeated play and feedback.}
We consider a general sum game played repeatedly for $T$ rounds. In round $t$ the players simultaneously choose a mixed strategy profile $x^t = (x_1^t,\dots,x_m^t)$, and player $i$ observes their entire expected loss vector $\ell_i^t=\loss_i(x^t)\in[0,1]^{n_i}$. This is the full-information, expected-loss feedback model of \cite{SALS15,DFG21}; the players observe exact expectations rather than the losses of sampled action profiles. The individual (external) regret of player $i$ after $T$ rounds is
\begin{equation}\label{eq:regret}
 \Reg_i(T)\;=\;\sum_{t=1}^T\ip{x_i^t}{\ell_i^t}-\min_{a\in[n_i]}\sum_{t=1}^T\ell_i^t(a).
\end{equation}

\paragraph{Optimistic Hedge.}
All players update their strategies using Optimistic Hedge with a common, fixed learning rate $\eta>0$. Each player $i$ starts from the uniform distribution and, having observed $\ell_i^t$, updates
\begin{equation}\label{eq:OH}
 x_i^{t+1}(a)\;=\;\frac{x_i^t(a)\exp\bigl[-\eta\bigl(2\ell_i^t(a)-\ell_i^{t-1}(a)\bigr)\bigr]}{\sum_{c \in [n_i]}x_i^t(c)\exp\bigl[-\eta\bigl(2\ell_i^t(c)-\ell_i^{t-1}(c)\bigr)\bigr]}, \qquad a \in [n_i], \qquad \ell_i^0=0 .
\end{equation}

\paragraph{Notation.} Given a distribution $p \in \Delta_n$ and $z \in \R^n$ we write $\tilt pz$ for the exponential tilt of $p$ by $z$, i.e. the probability vector with entries $\tilt pz (a) = p(a)e^{z(a)}/\sum_cp(c)e^{z(c)}$, so that for an Optimistic Hedge update, $x^{t+1}=\tilt{x^t}{-\eta(2\ell^t-\ell^{t-1})}$. We also denote the variance of $z$ with respect to $p$ by $\Var_p(z)=\sum_ap(a)(z(a)-\ip pz)^2$, the variance seminorm $\norm z_p=\sqrt{\Var_p(z)}$, and $\osc(z)=\max_{a \in [n]} z(a)-\min_{a \in [n]} z(a)$.

\section{Technical overview}\label{sec:overview}

This section states our results formally and outlines their proofs.\footnote{We assume for simplicity that all players have the same number of actions $n$; it is straightforward to show that our results can be generalized to obtain individual dependence on $n_i$.} We begin by introducing the temporal approximation framework we use to derive our two upper bounds. 

\subsection{Temporal approximation framework}\label{sec:ov-reduction}

Fix a player $i$ with $n$ actions who runs Optimistic Hedge against an arbitrary sequence $\ell_i^1,\dots,\ell_i^T\in[0,1]^n$ of loss vectors, with $\ell_i^0=0$, and let $x_i^t$ be their strategy in round $t$. Consider the total variation and the cumulative variance of the losses along the player's trajectory,
\begin{equation}\label{eq:DV}
 D_i(T)\;=\;\sum_{t=1}^T \norm{\ell_i^t - \ell_i^{t-1}}^2_{x_i^t},\qquad
 V_i(T)\;=\;\sum_{t=1}^T\norm{\ell_i^{t-1}}^2_{x_i^t}.
\end{equation}
Here, for a probability vector $p\in\Delta_n$ and $v\in\R^n$, we write $\norm{v}_p^2=\Var_p(v)=\sum_{a\in[n]}p(a)\bigl(v(a)-\ip pv\bigr)^2$ for the variance of $v$ under $p$. Throughout, $\lesssim$ hides absolute constants.

Our framework relies on the variance form of the RVU inequality of \cite{SALS15} (Lemma~4.1 in \cite{DFG21}), proven for completeness in Appendix~\ref{app:elementary}.

\begin{lemma}[Variance RVU \cite{DFG21}]\label{lem:potential}
For $0<\eta\le1/8$,
\begin{equation}\label{eq:potential}
 \Reg_i(T)\;\le\;\frac{\log n}{\eta}+\frac{2\eta}{3}D_i(T)-\frac{\eta}{3}V_i(T).
\end{equation}
\end{lemma}

\begin{corollary}\label{cor:stability}
If $0<\eta\le1/8$ and $D_i(T)\le\frac12V_i(T)+B$, then $\Reg_i(T)\le\log n/\eta+\frac23\eta B$.
\end{corollary}

Each of our regret upper bounds therefore culminates in a \emph{variation--variance inequality} $D_i(T)\le\frac12V_i(T)+B$ along the trajectory of play. To prove this inequality, we consider windows of $W$ consecutive rounds, and within each window (that is, locally) we approximate the Optimistic Hedge losses, for each action separately, (as functions of the local time $s=0,1,\ldots,W$ within the window) by a class $\mathcal F$ of simple functions for which the RVU property is straightforward to prove. Schematically, every $f \in \mathcal F$ satisfies
\begin{equation}
\label{eq:ov-rvu-simple}
    \sum_{s=0}^W \left( f(s+1) - f(s) \right)^2 \lesssim \frac{C}{W^2} \sum_{s=0}^W f(s)^2, 
\end{equation}
where $C$ is a constant which depends on the choice of $\mathcal F$, which would suffice for Corollary \ref{cor:stability} as long as $C/W^2$ can be bounded by a small constant. We remark that we ultimately need a version of Equation \eqref{eq:ov-rvu-simple} which holds in terms of the variance with respect to a distribution $p \in \Delta_n$, but this step only amounts to centering followed by taking expectation with respect to $p$, which are both straightforward. We also note that the quantity $C / W^2$ is not sufficiently small in general for the losses induced by Optimstic Hedge, an example being the construction of the lower bound given in Theorem \ref{thm:lower} which gives a loss sequence which oscillates frequently while being small in magnitude. The function class $\mathcal F$ approximates the true loss trajectory in the sense that for every window $\{t_0, t_0+1, \ldots, t_0 + W \}\subseteq [T]$, local time $0 \leq s \leq W$ and action $a \in [n]$ it holds that
\begin{equation}
\label{eq:ov-approx}
    \inf_{f \in \mathcal F} \left| \ell_i^{t_0+s}(a) - f(s) \right| \leq \eps_s,
\end{equation}
where $\eps_s$ measures the approximation error in local time $s$. By sliding all $W$-length windows over the horizon, Equations \eqref{eq:ov-rvu-simple} and \eqref{eq:ov-approx} ultimately result in 
\begin{equation}
\label{eq:ov-sliding}
    D_i(T) \lesssim \frac{C}{W^2} V_i(T) + W \left(1 + \sum_{t=1}^T \eps_t \right),
\end{equation}
which reduces the desired variation--variance inequality to choosing an approximating class $\mathcal F$ for which we can control the ratio $C / W^2$ and the approximation errors. In the two subsections that follow, we outline how this framework can be instantiated in two settings in order to prove regret bounds for $m$-player and two-player games, respectively.

\subsection{\texorpdfstring{$m$}{m} players: a logarithmic regret bound}\label{sec:ov-log}
Theorem~\ref{thm:log} gives a bound which matches the one of \cite{Ha2026log}, and is proved in full in Section~\ref{sec:logT}. The theorem follows from Corollary~\ref{cor:stability} once we show that for each player $i$ the loss trajectory satisfies
\begin{equation}\label{eq:ov-goal-log}
 D_i(T)\;\le\;\tfrac12 V_i(T)+O(\log T),
\end{equation}
which we prove using the temporal approximation framework outlined in Section \ref{sec:ov-reduction}. 

\begin{theorem}[Logarithmic regret for $m$ players]\label{thm:log}
Let $T\ge 2$. If all players of an $m$-player game, with $n$ actions for player $i$, run Optimistic Hedge with learning rate $0<\eta\le 1/(8704\sqrt m\,\log T)$, then the regret of every player is bounded via
\[
 \Reg_i(T)\ \le\ \frac{\log n}{\eta}+68\eta \log T.
\]
In particular, for $\eta=1/(8704\sqrt m\,\log T)$ every player achieves regret at most $O(\sqrt m\,\log n\,\log T)$.
\end{theorem}

\paragraph{Losses are locally polynomial in time.}
We consider windows of size $W = \Theta(\log T)$ and let 
\begin{equation*}
    \label{eq:ov-poly}
    \mathcal F = \left\{ q : \R \to \R \mid q \text{ is a polynomial of degree at most } d \right\},
\end{equation*}
where $d = \log_2 T$. The following approximation result holds (see Lemma~\ref{lem:approx}): if $\eta\lesssim 1/(\sqrt m\,(W+d))$, then for every $1 \leq t_0 \leq T-W$ and $a \in [n]$,
\begin{equation}
\label{eq:ov-approx-poly}
    \inf_{q \in \mathcal F} \max_{0 \leq s \leq W} \left| \ell_i^{t_0+s}(a) - q(s) \right| \leq 2^{-d},
\end{equation}
that is, the losses can be locally approximated by polynomials \emph{uniformly} over the window.
At a high level, the proof relies on expanding the loss trajectory over the given window as a power series in the variable $z$ which replaces the learning rate $\eta$, i.e., we write $\ell_i^{t_0 + s}(a; z) = \sum_{k=0}^\infty c_{i,a,k}(s) z^k$ where the coefficients are functions of local time within the window, and the true trajectory is obtained by substituting $z = \eta$. A crucial property of Optimistic Hedge dynamics is that under this expansion, each coefficient $c_{i,a,k}(s)$ coincides with a polynomial of degree at most $k$ in local time $s$. Moreover, having chosen $\eta$ sufficiently small, we can truncate this expansion after $d$ terms resulting in an error which is exponentially small in $d$, i.e. $|\ell_i^{t_0 + s}(a; z) - \sum_{k=0}^d c_{i,a,k}(s) z^k| \approx 2^{-d}$. In a bit more detail, if we expand $\ell_i^{t_0+s}(a;z)$ in powers of the \emph{real} variable $z$, the truncation error of the series is exponentially small in $d$ as long as $\sum_{k=0}^\infty |c_{i,a,k}(s)| z^k$ is uniformly bounded over $0 \leq z \leq \eta$, and indeed, this holds as long as $\eta \lesssim 1/(m(W+d))$ which would result in a regret bound which is linear in $m$. The $\sqrt{m}$ factor improvement comes from expanding $\ell_i^{t_0 + s}(a;z)$ in powers of the \emph{complex} variable $z$, and using standard techniques from complex analysis, it suffices to prove that $|\ell_i^{t_0+s}(a;z)|$ is uniformly bounded on the entire complex disk $|z| \leq \eta$, which in turn holds as long as $\eta \lesssim 1/(\sqrt{m}(W+d))$.

\paragraph{Polynomials obey a variation--variance inequality.} Having defined $\mathcal F$ as the class of polynomials of degree at most $d$, it is not hard to prove the specific version of Equation \eqref{eq:ov-rvu-simple} for this class, namely, for each $q \in \mathcal F$:
\begin{equation}\label{eq:ov-discrete}
 \sum_{s=W/4}^{3W/4-1}\bigl(q(s+1)-q(s)\bigr)^2\;\le\;\frac{16\,d(d+1)}{3W^2}\sum_{s=0}^W q(s)^2.
\end{equation}
Note that the left-hand side contains a summation over only the middle of the window, but this still suffices to obtain Equation \eqref{eq:ov-sliding} with $C \lesssim d^2$. The proof of Equation \eqref{eq:ov-discrete} relies on basic properties of the Legendre operator \cite{Szego75}, which in particular imply that if $q$ is a polynomial of degree at most $d$, then for every $0 < \gamma < \frac12$ it holds that
\begin{equation}\label{eq:legendre-interior}
 \int_\gamma^{1-\gamma}q'(x)^2\,dx\;\le\;\frac{d(d+1)}{\gamma(1-\gamma)}\int_0^1q(x)^2\,dx,
\end{equation}
which can be seen as a continuous analogue of Equation \eqref{eq:ov-discrete}.

\subsection{Two players: a constant regret bound}\label{sec:ov-two}

We now outline the proof of our two-player upper bound:

\begin{theorem}[Constant regret for two players]\label{thm:two}
There is an absolute constant $c>0$ such that in every two-player game $G\in\cG(2,n)$, for every learning rate $0<\eta\le c/n$ and every horizon $T\ge1$, if both players run Optimistic Hedge with learning rate $\eta$, then the regret of each player $i$ is bounded by
\[
 \Reg_i(T)\ \le\ \frac{\log n}{\eta}+\eta 
\]
In particular, for $\eta=c/n$, both players have regret $O(n\log n)$ for all $T$.
\end{theorem}

\paragraph{Setup.} Consider a two-player game with loss matrices $A,B\in[0,1]^{n\times n}$: when player~$1$ plays action $a$ and player~$2$ plays action $b$, their losses are $A_{ab}$ and $B_{ab}$. The players' strategies in round $t$ are $x_1^t,x_2^t$ and their loss vectors are $\ell_1^t=Ax_2^t$ and $\ell_2^t=B^\top x_1^t$; both players run Optimistic Hedge \eqref{eq:OH}. We analyze player~$1$ (the arguments are symmetric between the two players). 

Let $D_1(T)$ and $V_1(T)$ be as in \eqref{eq:DV}. By Corollary~\ref{cor:stability} it suffices to prove the variation--variance inequality $D_1(T)\le\frac12V_1(T)+\frac12$, which yields the bound $\Reg_1(T)\le\log n/\eta+\eta/3$; for $\eta\le c/n$ we in fact prove $D_1(T)\le\frac1{64}V_1(T)+\frac12$ (Lemma~\ref{lem:smooth}).
As in the $m$-player case, we use the temporal approximation framework of Section~\ref{sec:ov-reduction}, which relies on proving appropriate versions of Equations \eqref{eq:ov-rvu-simple} and \eqref{eq:ov-approx}.

\paragraph{Defining the class of simple functions $\mathcal{F}$: when the second player's loss is "frozen".}

Keeping with the framework of Section~\ref{sec:ov-reduction}, we define a class of simple functions $\mathcal{F}$. The functions in the class follow the losses of the first player that result assuming the second player plays Optimistic Hedge while ``freezing'' its own loss for a fixed window of $3W+2$ iterates, from round $t_0-1$ to $t_0+3W$, for some starting round $t_0$. We turn to examine the form of the losses for player~$1$ under this assumption. Let $x_2^{t_0}(b)$ denote player~$2$'s probability for action $b$ on round $t_0$, and $\ell_2^{t_0}(b)$ its frozen loss. We denote player~$1$'s losses over this window by $\bar\ell_1^{t_0+s}$ and player~$2$'s probability vector by $\bar x_2^{t_0+s}$, where the local time $s$ ranges from $0$ to $3W$ (the bar notation highlights that player~$2$'s losses are frozen). In this case, player~$2$'s score $2\ell_2^{t_0+s}-\ell_2^{t_0+s-1}$ equals $\ell_2^{t_0}$ in every round of the window, so its probability distribution on iterate $t_0+s$ is
\[
\bar x_2^{t_0+s}(b)\;=\;\frac{x_2^{t_0}(b)\,e^{-\eta \ell_2^{t_0}(b)s}}{\sum_{b'=1}^n x_2^{t_0}(b')\,e^{-\eta \ell_2^{t_0}(b')s}}\qquad(b\in[n]),
\]
which implies that player~$1$'s loss on each action $a\in[n]$ is
\begin{equation}\label{eq:ov-frozen}
\bar\ell_1^{t_0+s}(a)\;=\;(A\bar x_2^{t_0+s})(a)\;=\;\frac{\sum_{b=1}^nx_2^{t_0}(b)\,e^{-\eta \ell_2^{t_0}(b)s}\,A_{ab}}{\sum_{b=1}^nx_2^{t_0}(b)e^{-\eta \ell_2^{t_0}(b)s}}.
\end{equation}
Ignoring the denominator whose impact is negligible, for every action $a$ the losses $\bar\ell_1^{t_0+s}(a)$, $0\le s\le 3W$, are the values $f(s)$ of a linear combination $f$ of $n$ exponentials, with coefficients $\beta_b=x_2^{t_0}(b)A_{ab}$ and exponents $\lambda_b=-\eta \ell_2^{t_0}(b)$; that is, $f$ belongs to the class
\[
\mathcal{F} = \Bigl\{ f:\{0,1,\dots,3W\}\to\R \;\Big|\; f(s)=\sum_{b=1}^n \beta_b e^{\lambda_b s}~\text{for some}~\beta_b,\lambda_b\in\R \Bigr\}.
\]

\paragraph{A Bernstein inequality for exponential sums.}
The functions in $\mathcal F$ satisfy the following discrete Bernstein-type inequality, which is a variant of Eauation \eqref{eq:ov-rvu-simple}: for every $f\in\mathcal F$,
\begin{equation}\label{eq:ov-bernstein-disc}
 \sum_{s=W+1}^{2W}\bigl(f(s)-f(s-1)\bigr)^2\;\lesssim\;\frac{n^2}{W^2}\sum_{s=0}^{3W}f(s)^2 .
\end{equation}
It follows from the continuous Bernstein inequality for exponential sums of Borwein and Erd\'elyi \cite{BE96} (see Lemma~\ref{lem:bernstein}).
When we apply this to player 1's loss vectors, we get, for every action $a$ of player 1,
\begin{equation}\label{eq:ov-bernstein-action}
 \sum_{t=t_0+W+1}^{t_0+2W}\bigl(\bar\ell_1^t(a)-\bar\ell_1^{t-1}(a)\bigr)^2\;\lesssim\;\frac{n^2}{W^2}\sum_{t=t_0}^{t_0+3W}\bar\ell_1^t(a)^2 .
\end{equation}
In a similar way, while hiding some technicalities, one can get a similar bound, where instead of fixing an action $a$ we take variances over the probability distribution $x_1^t$, which yields:
\begin{equation}\label{eq:ov-bernstein-var}
 \sum_{t=t_0+W+1}^{t_0+2W}\norm{\bar\ell_1^t-\bar\ell_1^{t-1}}_{x_1^t}^2\;\lesssim\;\frac{n^2}{W^2}\sum_{t=t_0}^{t_0+3W}\norm{\bar\ell_1^t}_{x_1^t}^2 .
\end{equation}
We build on the fact that since the step size is selected small enough compared to the window that we sum over, $x_1^t$ does not change by much over this window. 
Therefore, Equation \eqref{eq:ov-bernstein-var} is obtained from Equation \eqref{eq:ov-bernstein-action} by summing over the actions $a$ of player $1$, with coefficients $x_1^{t_0}(a)$.

\paragraph{Accounting for the changing losses of Player 2.}
Equation~\eqref{eq:ov-bernstein-var} is taken consider player 2's losses are frozen. When one accounts for the true losses of player 2, one obtains a slightly different inequality:
\begin{equation}\label{eq:ov-local-true}
 \sum_{t=t_0+W+1}^{t_0+2W}\norm{\ell_1^t-\ell_1^{t-1}}_{x_1^t}^2\;\lesssim\;\frac{n^2}{W^2}\sum_{t=t_0}^{t_0+3W}\norm{\ell_1^t}_{x_1^t}^2+\eta^2n^2W^2\sum_{r=t_0}^{t_0+3W-1}\norm{\ell_2^r-\ell_2^{r-1}}_\infty^2.
\end{equation}

The error term is still in terms of player~$2$'s losses, but in a two-player game these move \emph{only if} player~$1$ moves: $\ell_2^r-\ell_2^{r-1}=B^\top(x_1^r-x_1^{r-1})$ with $B\in[0,1]^{n\times n}$,
\[
 \norm{\ell_2^r-\ell_2^{r-1}}_\infty\le\norm{x_1^r-x_1^{r-1}}_1.
\]
By definition of Optimistic Hedge the iterates of the first player move only if their (previous two) losses have variance with relation to their current mixed strategy distribution. Concretely, we may obtain the following bounds:
\[
 \norm{x_1^r-x_1^{r-1}}_1\le3\eta\bigl(\norm{\ell_1^{r-1}}_{x_1^{r-1}}+\norm{\ell_1^{r-2}}_{x_1^{r-2}}\bigr).
\]
Squaring and substituting it into \eqref{eq:ov-local-true} expresses the error in terms of player~$1$'s own variances, 
\begin{equation}\label{eq:ov-local-own}
 \sum_{t=t_0+W+1}^{t_0+2W}\norm{\ell_1^t-\ell_1^{t-1}}_{x_1^t}^2\;\lesssim\;\frac{n^2}{W^2}\sum_{t=t_0}^{t_0+3W}\norm{\ell_1^t}_{x_1^t}^2+\eta^4n^2W^2\sum_{r=t_0-1}^{t_0+3W-2}\bigl(\norm{\ell_1^r}_{x_1^r}^2+\norm{\ell_1^{r-1}}_{x_1^{r-1}}^2\bigr).
\end{equation}

Setting $\eta=\Theta(1/n)$ with a small enough constant and $W=\Theta(n)$ with can roughly obtain
\[
 \sum_{t=t_0+W+1}^{t_0+2W}\norm{\ell_1^t-\ell_1^{t-1}}_{x_1^t}^2\;\le\;\frac13 \sum_{t=t_0}^{t_0+3W}\norm{\ell_1^t}_{x_1^t}^2.
\]
After summing over $T/W$ overlapping windows of size $3W$ each, we approximately get $D_1(T)\le V_1(T)/2$, which is what needed in order to invoke Lemma 3.1. This yields the regret bound $R_1(T)\le O(\log(n)/\eta)=O(n\log n)$. By symmetry, this regret bound also holds for player 2.
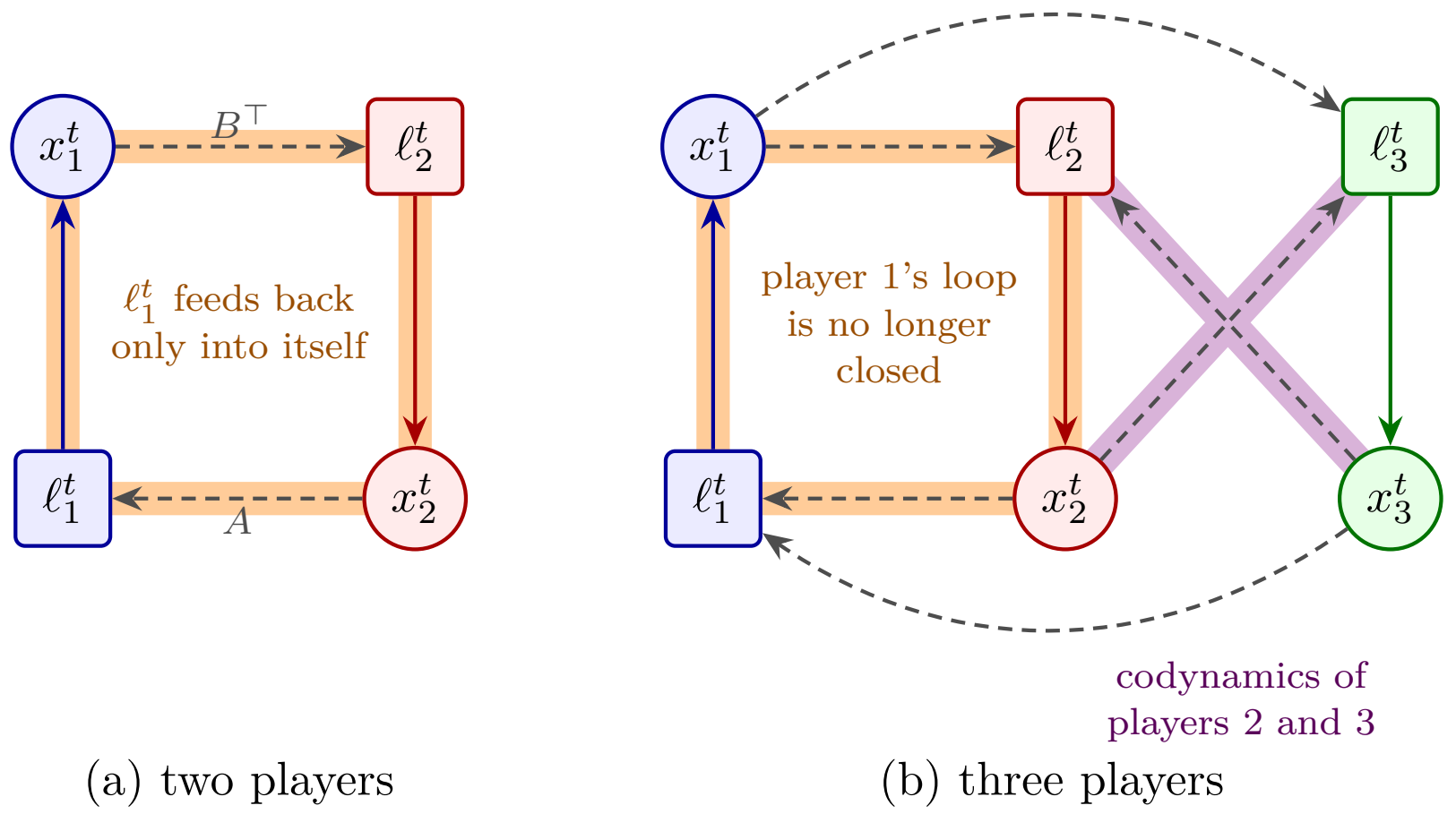
\begin{figure}[t]
\centering
\begin{tikzpicture}[
  >={Stealth[length=2.2mm,width=1.8mm]},
  font=\small,
  strat/.style={circle,draw,line width=0.8pt,minimum size=7.5mm,inner sep=1pt},
  loss/.style={rectangle,rounded corners=2pt,draw,line width=0.8pt,minimum size=7mm,inner sep=2pt},
  pone/.style={draw=blue!60!black,fill=blue!8},
  ptwo/.style={draw=red!65!black,fill=red!8},
  pthree/.style={draw=green!45!black,fill=green!10},
  eone/.style={->,line width=0.8pt,blue!60!black},
  etwo/.style={->,line width=0.8pt,red!65!black},
  ethree/.style={->,line width=0.8pt,green!45!black},
  couple/.style={->,line width=0.8pt,black!70,densely dashed},
  wt/.style={font=\scriptsize,inner sep=1.5pt},
  note/.style={font=\scriptsize,align=center},
  loop/.style={line width=7pt,orange!40,line cap=round,line join=round},
  codyn/.style={line width=7pt,violet!30,line cap=round},
]
\begin{scope}
\coordinate (Ax) at (0,2.6);  \coordinate (Ab) at (2.6,2.6);
\coordinate (Ay) at (2.6,0);  \coordinate (Al) at (0,0);
\draw[loop] (Ax) -- (Ab) -- (Ay) -- (Al) -- cycle;
\node[strat,pone] (ax) at (Ax) {$x_1^t$};
\node[loss,ptwo]  (ab) at (Ab) {$\ell_2^t$};
\node[strat,ptwo] (ay) at (Ay) {$x_2^t$};
\node[loss,pone]  (al) at (Al) {$\ell_1^t$};
\draw[eone]   (al) -- (ax);
\draw[couple] (ax) -- node[wt,above] {$B^{\top}$} (ab);
\draw[etwo]   (ab) -- (ay);
\draw[couple] (ay) -- node[wt,below] {$A$} (al);
\node[note,text=orange!60!black] at (1.3,1.3) {$\ell_1^t$ feeds back\\only into itself};
\node at (1.3,-2.1) {(a) two players};
\end{scope}
\begin{scope}[xshift=4.8cm]
\coordinate (Bx) at (0,2.6);    \coordinate (Bl) at (0,0);
\coordinate (Bb) at (2.6,2.6);  \coordinate (By) at (2.6,0);
\coordinate (Bc) at (5.0,2.6);  \coordinate (Bz) at (5.0,0);
\draw[loop]  (Bx) -- (Bb) -- (By) -- (Bl) -- cycle;
\draw[codyn] (By) -- (Bc);
\draw[codyn] (Bz) -- (Bb);
\node[strat,pone]   (bx) at (Bx) {$x_1^t$};
\node[loss,pone]    (bl) at (Bl) {$\ell_1^t$};
\node[loss,ptwo]    (bb) at (Bb) {$\ell_2^t$};
\node[strat,ptwo]   (by) at (By) {$x_2^t$};
\node[loss,pthree]  (bc) at (Bc) {$\ell_3^t$};
\node[strat,pthree] (bz) at (Bz) {$x_3^t$};
\draw[eone]   (bl) -- (bx);
\draw[couple] (bx) -- (bb);
\draw[couple] (bx) to[bend left=35] (bc);
\draw[etwo]   (bb) -- (by);
\draw[ethree] (bc) -- (bz);
\draw[couple] (by) -- (bl);
\draw[couple] (bz) to[bend left=35] (bl);
\draw[couple] (by) -- (bc);
\draw[couple] (bz) -- (bb);
\node[note,text=orange!60!black] at (1.3,1.3) {player 1's loop\\is no longer\\closed};
\node[note,text=violet!70!black] at (3.9,-1.5) {codynamics of\\players 2 and 3};
\node at (2.5,-2.1) {(b) three players};
\end{scope}
\end{tikzpicture}
\caption{Dependencies between the players' loss and mixed strategy iterates. (a)~With two players the only route from $\ell_1^t$ back to itself is the loop $\ell_1\to x_1\to\ell_2\to x_2\to\ell_1$ (orange). (b)~A third player, with strategy $x_3^t$ and loss $\ell_3^t$, also feeds $\ell_2^t$ and is fed by $x_2^t$, so the coupled dynamics of players 2 and 3 (violet) enter player~1's loop.}\label{fig:feedback}
\end{figure}
\begin{remark}
    For clarity, the sketch above works directly with the discrete iterates and applies the discrete Bernstein inequality \eqref{eq:ov-bernstein-disc} on each window. The formal proof differs in this respect: it works with continuous interpolations of the losses, obtained by running player~$2$'s exponential tilt continuously between consecutive rounds, so that the frozen losses become a ratio of exponential sums on the whole window. It then applies the continuous Bernstein inequality, and shows that the resulting integrals are close to the discrete sums that define $D_1(T)$ and $V_1(T)$.
\end{remark}

\paragraph{Why the bound for two players does not carry over to the $>2$ player case.}\label{subsec:notg2}

In order to prove that $D_1(T)\le\frac1{64}V_1(T)+\frac12$ , we use the fact that the (original) loss vector iterates are sufficiently close to the iterates when the second player "freezes" their loss. However, as we discussed, the drift between the faithful and approximate continuous curves is dependent solely on the same player's loss iterates (recall Equation \eqref{eq:ov-local-own}). It is \emph{this dependency on the same player's loss iterates}, rather than any other player's, which makes the "frozen" loss vector iterates sufficiently close to the original loss vector iterates.

The same-player dependency is due to the strictly coupled dynamics between the two players (Figure~\ref{fig:feedback}(a)). The iterates $\ell_2^t$ are dependent only on themselves and the iterates $x_1^t$, which are only dependent on themselves and the iterates $\ell_1^t$. As such, the change between keeping $\ell_2^t$ constant and allowing $\ell_2^t$ to progress is strictly dependent on how much the iterates $\ell_1^t$ change. Using a similar syllogism, we find that the iterates $\ell_1^t$ are also strictly dependent on themselves and $\ell_2^t$. Therefore, intuitively, it can be observed that the change in $\ell_1^t$ from the original dynamics to when $\ell_2^t$ is kept constant is strictly dependent upon the original $\ell_1^t$ iterates themselves. 

This is not the case, however, in games with more than two players (Figure~\ref{fig:feedback}(b)). Any player $i$'s loss iterates are dependent on the other player's mixed strategies, which are dependent on their own losses. However, if we were to try and "freeze" players' losses for a similar proof technique, we find instead that the other player's losses \emph{are dependent on each other}. This co-dependency means that the drift from the original loss player $i$ iterates to the iterates when the rest of the player's losses are "frozen" has to account for the codynamics between the other players in the original iterates; these codynamics add unavoidable terms \emph{dependent on the other players} to the drift. 

\subsection{Three players: oscillators and frequency doubling}\label{sec:ov-lower}

\begin{theorem}[An exponential barrier for three players]\label{thm:lower}
There are absolute constants $C_0$ and $\eta_0>0$ such that for every $n\ge C_0$ and every learning rate $\eta\in(0,\eta_0]$ there is a three-player game $G\in\cG(3,n)$ in which, when all players run Optimistic Hedge with learning rate $\eta$ starting from the uniform distribution,
\[
 \sup_T\,\Reg_3(T)\ \ge\ \frac{2^{n/8}\log n}{C_0\,n}-1\ =\ 2^{\Omega(n)} .
\]
In particular $\inf_{0<\eta\le\eta_0}\sup_{T\ge1}\,\sup_{G\in\cG(3,n)}\,\max_i\Reg_i(T)\ge2^{n/8}\log n/(C_0n)-1$. Moreover, if $\eta\ge C_0\,n\,2^{-n/8}$, then $G$ can be chosen so that player $3$'s regret grows linearly: $\Reg_3(T)\ge c\,T-C$ for all $T$, for some $c=c(G,\eta)>0$ and $C=C(G,\eta)$.
\end{theorem}

The two regimes are of different nature. For $\eta<C_0n2^{-n/8}$ the learning rate is so small that even a game with losses which are independent of the other players' actions already forces $\log n/\eta$ regret (Proposition~\ref{prop:constloss}). For larger $\eta$ we construct a game in which player $3$'s regret is unbounded; the rest of this subsection sketches this construction. We first describe the construction for three players, which gives the $2^{\Omega(n)}$ lower bound, and only at the end of this subsection explain how it extends to $m$ players, with a lower bound of $2^{\Omega(mn)}$.

\paragraph{The idea.}
The game has three players. Players $1$ and $2$ play against each other and ignore player $3$, and their utilities are designed so that Optimistic Hedge never converges: their strategies settle into a cycle and follow it forever. That learning dynamics in games may cycle rather than converge is well known \cite{MPP18,BP18}, and the last-iterate behavior of optimistic dynamics has been studied in detail \cite{DP19,WLZL21,cai2024fast}; to our knowledge, turning such a cycle into a regret lower bound is new. Player $3$, the \emph{observer}, has losses that depend only on player $1$'s action, so the losses they see oscillate too. Optimistic Hedge bets that the next loss will resemble the last one. If the oscillation is fast enough, this bet is systematically wrong, and the observer's regret grows linearly. The full proof is in Section~\ref{sec:lowerproof}.

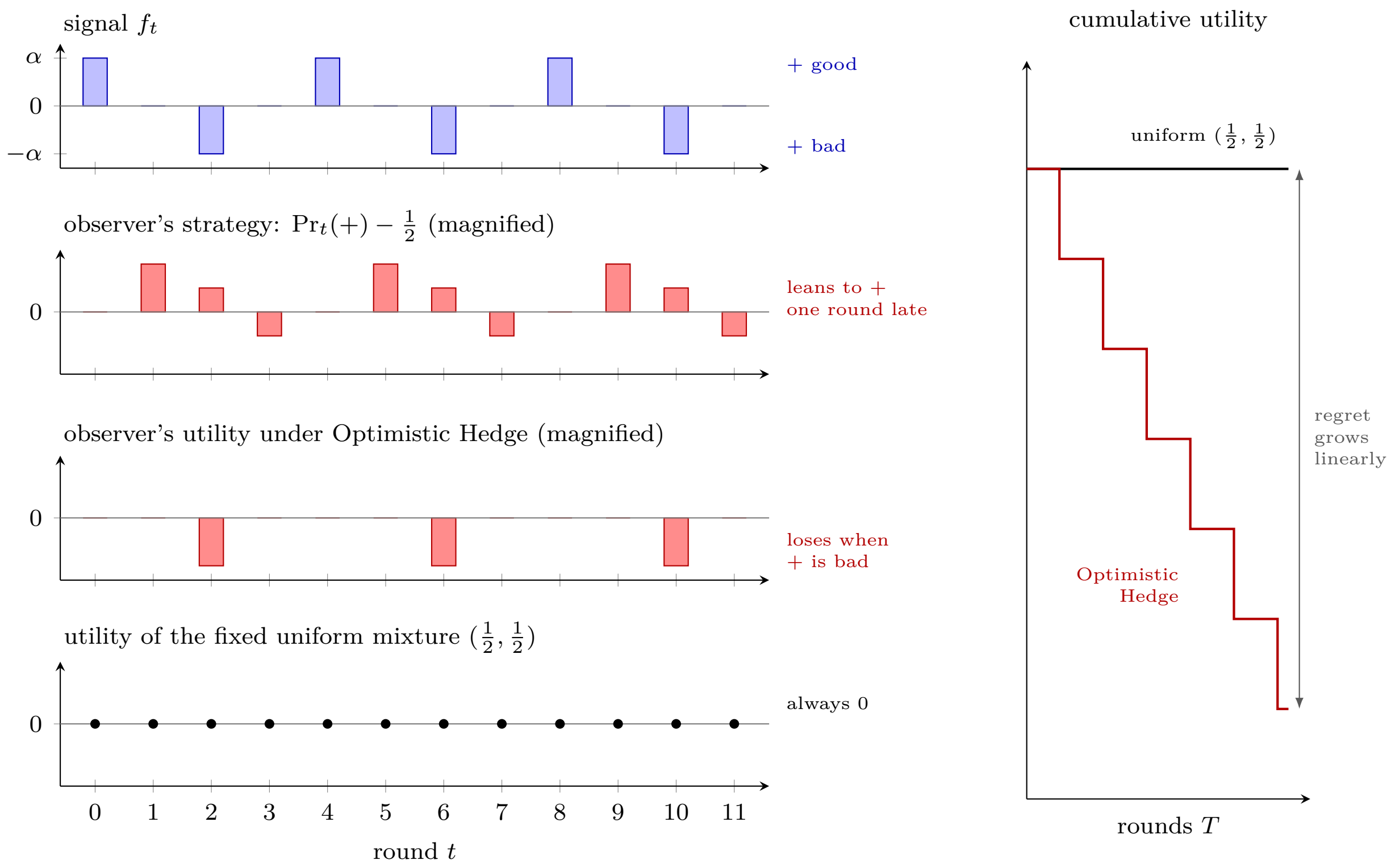
\begin{figure}[tb]
\centering
\pgfplotsset{ovrow/.style={scale only axis,width=0.5\textwidth,height=1.45cm,
   xmin=-0.6,xmax=11.6,xtick={0,...,11},xticklabels={},axis x line=bottom,axis y line=left,
   font=\scriptsize,ylabel style={font=\tiny},clip=false}}
\tikzset{rowtitle/.style={font=\scriptsize,anchor=south west,inner sep=1pt}}
\begin{tikzpicture}
 \begin{axis}[ovrow,name=r1,ymin=-1.3,ymax=1.3,ytick={-1,0,1},yticklabels={$-\alpha$,$0$,$\alpha$}]
  \node[rowtitle] at (rel axis cs:0,1.04) {signal $f_t$};
  \addplot[ybar,bar width=8pt,fill=blue!25,draw=blue!70!black] coordinates
   {(0,1) (1,0) (2,-1) (3,0) (4,1) (5,0) (6,-1) (7,0) (8,1) (9,0) (10,-1) (11,0)};
  \draw[gray] (axis cs:-0.6,0) -- (axis cs:11.6,0);
  \node[font=\tiny,anchor=west,blue!70!black] at (axis cs:11.7,0.85) {$+$ good};
  \node[font=\tiny,anchor=west,blue!70!black] at (axis cs:11.7,-0.85) {$+$ bad};
 \end{axis}
 \begin{axis}[ovrow,name=r2,at={(r1.south west)},anchor=north west,yshift=-0.95cm,
   ymin=-1.3,ymax=1.3,ytick={0}]
  \node[rowtitle] at (rel axis cs:0,1.04) {observer's strategy: $\Pr_t(+)-\frac12$ (magnified)};
  \addplot[ybar,bar width=8pt,fill=red!45,draw=red!70!black] coordinates
   {(0,0) (1,1) (2,0.5) (3,-0.5) (4,0) (5,1) (6,0.5) (7,-0.5) (8,0) (9,1) (10,0.5) (11,-0.5)};
  \draw[gray] (axis cs:-0.6,0) -- (axis cs:11.6,0);
  \node[font=\tiny,anchor=west,red!70!black,align=left] at (axis cs:11.7,0.3) {leans to $+$\\ one round late};
 \end{axis}
 \begin{axis}[ovrow,name=r3,at={(r2.south west)},anchor=north west,yshift=-0.95cm,
   ymin=-1.3,ymax=1.3,ytick={0}]
  \node[rowtitle] at (rel axis cs:0,1.04) {observer's utility under Optimistic Hedge (magnified)};
  \addplot[ybar,bar width=8pt,fill=red!45,draw=red!70!black] coordinates
   {(0,0) (1,0) (2,-1) (3,0) (4,0) (5,0) (6,-1) (7,0) (8,0) (9,0) (10,-1) (11,0)};
  \draw[gray] (axis cs:-0.6,0) -- (axis cs:11.6,0);
  \node[font=\tiny,anchor=west,red!70!black,align=left] at (axis cs:11.7,-0.7) {loses when\\ $+$ is bad};
 \end{axis}
 \begin{axis}[ovrow,name=r4,at={(r3.south west)},anchor=north west,yshift=-0.95cm,
   ymin=-1.3,ymax=1.3,ytick={0},
   xticklabels={0,...,11},xlabel={round $t$}]
  \node[rowtitle] at (rel axis cs:0,1.04) {utility of the fixed uniform mixture $(\frac12,\frac12)$};
  \addplot[only marks,mark=*,mark size=1.4pt,black] coordinates
   {(0,0) (1,0) (2,0) (3,0) (4,0) (5,0) (6,0) (7,0) (8,0) (9,0) (10,0) (11,0)};
  \draw[gray] (axis cs:-0.6,0) -- (axis cs:11.6,0);
  \node[font=\tiny,anchor=west] at (axis cs:11.7,0.4) {always $0$};
 \end{axis}
 \begin{axis}[at={(r1.north east)},anchor=north west,xshift=3.0cm,yshift=-0.2cm,
   scale only axis,width=0.2\textwidth,height=8.6cm,
   xmin=0,xmax=26,ymin=-7,ymax=1.2,xtick=\empty,ytick=\empty,
   axis x line=bottom,axis y line=left,font=\scriptsize,xlabel={rounds $T$},
   title={\scriptsize cumulative utility},clip=false]
  \addplot[black,thick] coordinates {(0,0) (24,0)};
  \node[font=\tiny,anchor=south east] at (axis cs:24,0.1) {uniform $(\frac12,\frac12)$};
  \addplot[red!70!black,thick] coordinates
   {(0,0) (3,0) (3,-1) (7,-1) (7,-2) (11,-2) (11,-3) (15,-3) (15,-4) (19,-4) (19,-5) (23,-5) (23,-6) (24,-6)};
  \node[font=\tiny,red!70!black,anchor=north east,align=right] at (axis cs:15,-4.3) {Optimistic\\ Hedge};
  \draw[{Latex[length=1.3mm]}-{Latex[length=1.3mm]},gray!70!black] (axis cs:25,0) -- (axis cs:25,-6);
  \node[font=\tiny,align=left,anchor=west,gray!70!black] at (axis cs:25.3,-3) {regret\\ grows\\ linearly};
 \end{axis}
\end{tikzpicture}
\caption{
Illustration of the induced loss sequence of the observer.
}
\label{fig:ov-observer}
\end{figure}

\paragraph{Fast oscillations hurt the observer.}
Split the observer's actions into two groups, $+$ and $-$, and suppose that the advantage $f_t$ of group $+$ over group $-$ oscillates with period four: $f_t=\alpha,0,-\alpha,0,\dots$. Optimistic Hedge reacts to the last loss twice, once through the cumulative loss and once more through its prediction. As a result the observer favors $+$ most one round \emph{after} it was good, and still favors it in the next round, when it is bad (Figure~\ref{fig:ov-observer}). She loses a fixed amount in every period, while either group alone breaks even, so their regret grows linearly. For slow oscillations the prediction is accurate and helps instead, which is what makes the upper bounds possible. It therefore suffices that players $1$ and $2$ oscillate with period four.

\begin{figure}[tb]
\centering
\begin{tikzpicture}[
  blk/.style={draw,rounded corners=2pt,minimum width=2.25cm,minimum height=1.05cm,align=center,font=\small},
  obs/.style={blk,fill=red!6,draw=red!60!black},
  arr/.style={-{Latex[length=2mm]},thick},
  lab/.style={font=\scriptsize,align=center}]
 \node[blk,fill=blue!6] (b1) {Block $1$\\[-1pt]\scriptsize oscillator};
 \node[blk,right=1.25cm of b1] (b2) {Block $2$\\[-1pt]\scriptsize doubler};
 \node[right=0.55cm of b2] (dots) {$\cdots$};
 \node[blk,right=0.55cm of dots] (bk) {Block $k$\\[-1pt]\scriptsize doubler};
 \node[obs,right=1.25cm of bk] (o) {Player $3$\\[-1pt]\scriptsize observer};
 \draw[arr] (b1) -- node[above,lab]{square} (b2);
 \draw[arr] (b2) -- (dots);
 \draw[arr] (dots) -- (bk);
 \draw[arr] (bk) -- node[above,lab]{signal $f_t$} (o);
 \foreach \n/\f/\a in {b1/{2\pi/M}/a, b2/{4\pi/M}/{a^2}, bk/{\pi/2}/{a^{2^{k-1}}}}{
   \node[lab,below=2pt of \n] {freq.\ $\f$\\ amplitude $\a$\\ 8 actions};}
 \node[lab,below=2pt of o] {linear regret\\ $\Reg_3(T)=c_{\mathrm{obs}}T+O(1)$};
 \draw[decorate,decoration={brace,amplitude=5pt},thick]
   ([yshift=6pt]b1.north west) -- node[above=6pt,lab]{players $1$ and $2$: $k$ blocks, one shared normalizer, $8k+1\le n$ actions}
   ([yshift=6pt]bk.north east);
\end{tikzpicture}
\caption{The three-player game. Block $1$ of players $1$ and $2$ rotates slowly, with period $M=2^{k+1}$ (Section~\ref{sec:lower-first}). Each later block sees the square of its predecessor's displacement and so rotates twice as fast (Section~\ref{sec:lower-second}). Block $k$ has period $4$, fast enough to give the observer linear regret (Section~\ref{sec:lower-observer}). All blocks live in the same two players (Section~\ref{sec:lower-joint}).}
\label{fig:cascade}
\end{figure}
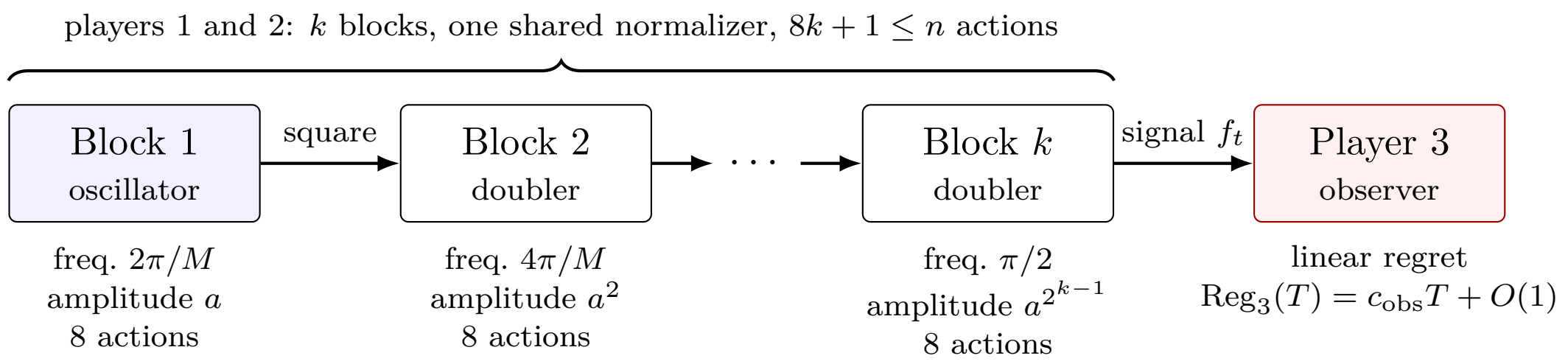

\paragraph{A cascade of blocks.}
Getting players $1$ and $2$ to oscillate this fast is the main difficulty. A small learning rate changes the probabilities only slightly in each round, so an oscillation of two players is necessarily slow, with a period of order at least $1/\eta$. We therefore build the fast oscillation in stages (Figure~\ref{fig:cascade}). The actions of players $1$ and $2$ are split into $k$ \emph{blocks} of eight actions. Block $1$ oscillates on its own, slowly. Each later block is driven by the block before it and oscillates with half its period, so after about $\log_2(1/\eta)$ blocks the period is four, and the last block drives the observer. Each halving costs eight actions, which is the source of the exponential dependence on $n$.

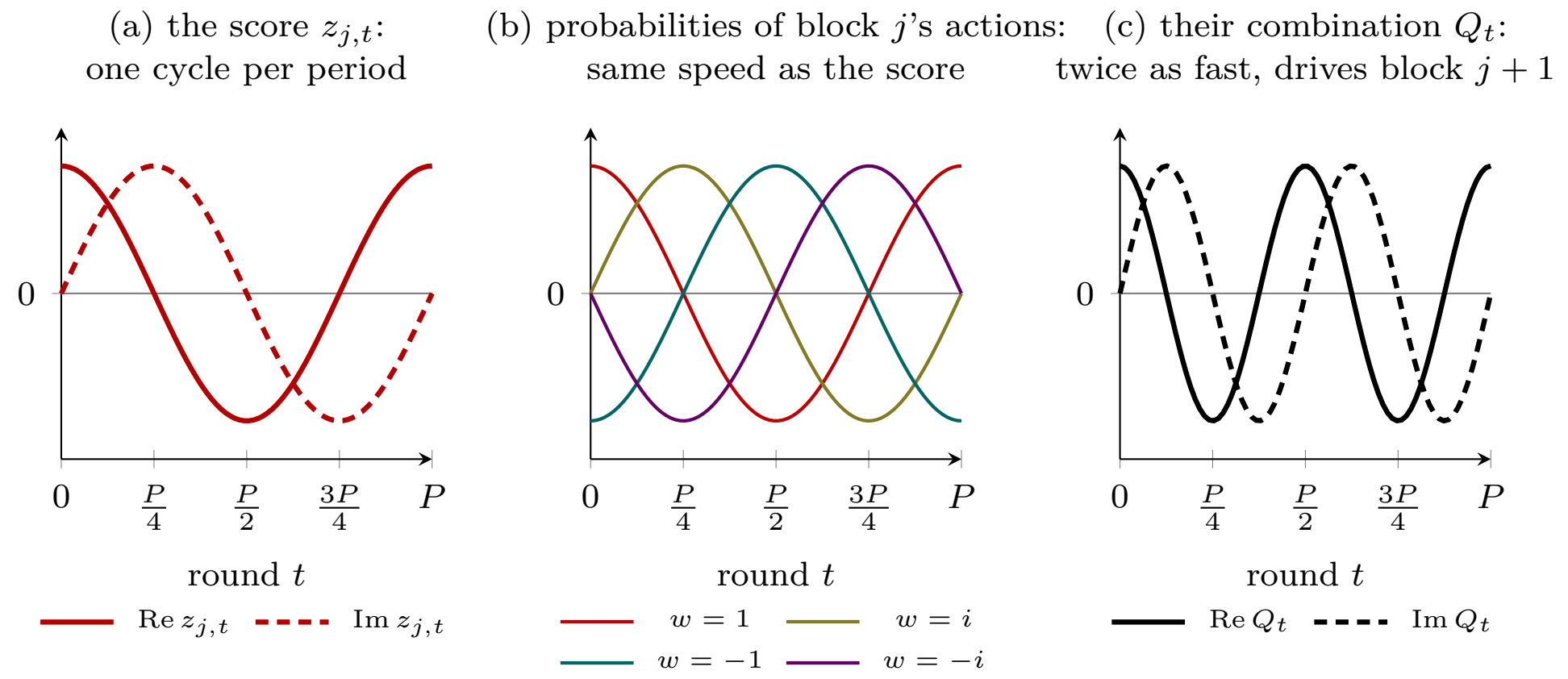
\begin{figure}[htbp]
\centering
\pgfplotsset{cycax/.style={width=0.28\textwidth,height=4.3cm,
   xmin=0,xmax=6.2832,ymin=-1.3,ymax=1.3,ytick={0},axis lines=left,font=\scriptsize,
   xtick={0,1.5708,3.1416,4.7124,6.2832},xticklabels={$0$,$\frac P4$,$\frac P2$,$\frac{3P}4$,$P$},
   xlabel={round $t$},clip=false,title style={align=center},
   legend style={at={(0.5,-0.4)},anchor=north,font=\tiny,legend columns=2,draw=none,column sep=3pt}}}
\begin{tikzpicture}
 \begin{axis}[cycax,name=pa,
   title={\scriptsize (a) the score $z_{j,t}$:\\ \scriptsize one cycle per period}]
  \draw[gray] (axis cs:0,0) -- (axis cs:6.2832,0);
  \addplot[domain=0:6.2832,samples=80,very thick,red!70!black] {cos(deg(x))};
  \addlegendentry{$\Rea z_{j,t}$}
  \addplot[domain=0:6.2832,samples=80,very thick,red!70!black,densely dashed] {sin(deg(x))};
  \addlegendentry{$\Ima z_{j,t}$}
 \end{axis}
 \begin{axis}[cycax,name=pb,anchor=west,at={($(pa.east)+(1.3cm,0)$)},
   title={\scriptsize (b) probabilities of block $j$'s actions:\\ \scriptsize same speed as the score}]
  \draw[gray] (axis cs:0,0) -- (axis cs:6.2832,0);
  \addplot[domain=0:6.2832,samples=80,thick,red!75!black] {cos(deg(x))};
  \addlegendentry{$w=1$}
  \addplot[domain=0:6.2832,samples=80,thick,olive!80!black] {sin(deg(x))};
  \addlegendentry{$w=i$}
  \addplot[domain=0:6.2832,samples=80,thick,teal!80!black] {-cos(deg(x))};
  \addlegendentry{$w=-1$}
  \addplot[domain=0:6.2832,samples=80,thick,violet!80!black] {-sin(deg(x))};
  \addlegendentry{$w=-i$}
 \end{axis}
 \begin{axis}[cycax,name=pc,anchor=west,at={($(pb.east)+(1.3cm,0)$)},
   title={\scriptsize (c) their combination $Q_t$:\\ \scriptsize twice as fast, drives block $j+1$}]
  \draw[gray] (axis cs:0,0) -- (axis cs:6.2832,0);
  \addplot[domain=0:6.2832,samples=80,very thick,black] {cos(2*deg(x))};
  \addlegendentry{$\Rea Q_t$}
  \addplot[domain=0:6.2832,samples=80,very thick,black,densely dashed] {sin(2*deg(x))};
  \addlegendentry{$\Ima Q_t$}
 \end{axis}
\end{tikzpicture}
\caption{Frequency doubling from block $j$ to block $j+1$ (Section~\ref{sec:ov-lower}), drawn schematically over one period $P$ of block $j$; the actual oscillations are much smaller. (a) The score $z_{j,t}$. (b) The probabilities of the actions $\pm1,\pm i$; the four corner actions behave alike. (c) The combination
$Q_t=\big[p_t(1)+p_t(-1)\big]-\big[p_t(i)+p_t(-i)\big]+\frac i2\big[p_t(1+i)+p_t(-1-i)\big]-\frac i2\big[p_t(1-i)+p_t(-1+i)\big]\approx z_{j,t}^2/n$,
which follows by expanding $p_t(w)\propto e^{\Rea(\overline w z_{j,t})}$ to second order.}
\label{fig:ov-cycle}
\end{figure}

\paragraph{One block's rotation.}
Fix a block $j$. Its eight actions are identified with points $w$ of the complex plane: the four corners and the four edge midpoints of a square around $0$. The game is designed so that the players' strategy on block $j$ is described by a single complex number, the block's \emph{score} $z_{j,t}$: action $w$ is played with probability proportional to $e^{\Rea(\overline wz_{j,t})}$, which is larger the more $z_{j,t}$ points toward $w$. Along the cycle, the score rotates around a small circle (Figure~\ref{fig:ov-cycle}(a)). As it turns, it points toward each action in turn, so each action's probability rises and falls at the same speed as the score rotates (Figure~\ref{fig:ov-cycle}(b)). In block $1$ this rotation is produced by the block's own utilities; in every later block it is driven by the block before it, as follows.

\paragraph{Halving the period.}
Opposite actions rise and fall in antiphase, so adding up their probabilities cancels the oscillation to first order, and what remains oscillates twice as fast. In terms of the score, a fixed linear combination of the probabilities of block $j$ equals the square of the score, to leading order, and squaring a rotation doubles its speed (Figure~\ref{fig:ov-cycle}(c)). The utilities of block $j+1$ depend on the actions of block $j$ through exactly this combination. The score of block $j+1$ is therefore pushed around a circle twice as fast as the score of block $j$, so its period is half as long.

\paragraph{Technical subtleties.}
This picture would be exact if the dynamics of the scores were linear. The softmax makes the actual dynamics nonlinear, and the proof has to deal with three issues.
\begin{itemize}
 \item \emph{An exact cycle.} The first block only rotates along a neutral direction, where small nonlinear errors are not damped and would accumulate over time. The implicit function theorem gives an exactly periodic orbit, after a small adjustment of the block's utilities.
 \item \emph{Convergence from the start.} The cycle does not attract every nearby starting point. The stable manifold theorem gives a manifold of starting points, the \emph{stable manifold} of the cycle, from which the dynamics converge to the cycle. Using the implicit function theorem again, we adjust the game slightly, by shifting the center of the cycle, so that the uniform starting point lies on this stable manifold.
 \item \emph{The later blocks.} Driving a block only by its predecessor does not make it converge to a cycle. We therefore modify its losses slightly: besides the term coming from the previous block, each block has a small self-interaction term that depends on its own actions. This term makes the block's dynamics contracting, so it settles into an exact cycle even though its input is only approximately periodic, and the remaining nonlinear errors do not accumulate.
\end{itemize}

\paragraph{More than three players.}
With $m$ players the same mechanism gives a barrier $2^{\Omega(mn)}$ (Theorem~\ref{thm:lower-many}). All players except the last are grouped into $\lfloor(m-1)/2\rfloor$ pairs, and each pair carries about $n/8$ blocks of its own, so the cascade has about $mn/16$ blocks in total. The first block of each pair is driven by the last block of the previous pair, in the same way as a block is driven by its predecessor within a pair, and player $m$ observes the last block of the last pair. Since a pair is never influenced by the pairs after it, the convergence argument carries over unchanged, and every additional pair halves the period about $n/8$ more times.

\section{Proof of the \texorpdfstring{$O(\sqrt{m}\log n\log T)$}{O(sqrt(m) log n log T)} regret upper bound for \texorpdfstring{$m$}{m} players}\label{sec:logT}

This section includes a fool proof of Theorem~\ref{thm:log}. The proof consists of three main parts: a local polynomial approximation lemma (Section~\ref{sec:log-approx}), the discrete variation--variance inequality for polynomials (Section~\ref{sec:log-energy}), and a sliding-window lemma which gives the variation--variance inequality over the entire horizon (Section~\ref{sec:log-windows}). Section~\ref{sec:log-completion} completes the proof.

\subsection{Local polynomial approximation}\label{sec:log-approx}

\begin{lemma}[Local approximation]\label{lem:approx}
Let $d\ge0$ and $W\ge1$ be integers, and suppose that all players run \eqref{eq:OH} with
\[
 0<\eta\le\frac1{256\sqrt m\,(W+d)}.
\]
Then for every player $i$, every integer $r$ with $d+1\le r\le T-W$ and every action $a \in [n]$ there is a polynomial $q_{i,r,a}:\R\to\R$ of degree at most $d$ such that
\begin{equation}\label{eq:approx}
 \max_{s\in\{0,\dots,W\}}\left|\ell_i^{r+s}(a)-q_{i,r,a}(s) \right| \le 2^{-d}.
\end{equation}
\end{lemma}

\begin{proof}
Fix $r$, put $t_0=r-d\ge1$ and $N=W+d$, and freeze the state of the dynamics at round $t_0$: let $p_j=x_j^{t_0}$ and $v_j=\ell_j^{t_0-1}$ for every player $j$. 

\emph{Step 1: defining the auxiliary trajectories.}
For a complex number $z$ define, simultaneously for all players and for local times $u=0,1,\dots,N$,
\begin{equation}\label{eq:localprocess}
 \begin{gathered}
 Y_j^0=0,\qquad \Lambda_j^{-1}=v_j,\qquad
 D_j^u(z)=\sum_ap_j(a)e^{Y_j^u(a;z)},\qquad
 X_j^u(a;z)=\frac{p_j(a)e^{Y_j^u(a;z)}}{D_j^u(z)},\\[2pt]
 \Lambda_j^u(z)=\loss_j\bigl(X^u(z)\bigr),\qquad
 Y_j^{u+1}(z)=Y_j^u(z)-z\bigl(2\Lambda_j^u(z)-\Lambda_j^{u-1}(z)\bigr)\quad(u<N).
 \end{gathered}
\end{equation}
Since $D_j^u(0)=1$, induction over $u$ shows that all these quantities are analytic functions of $z$ on some neighborhood of $0$, and Step~3 identifies a disk on which they are all defined. Substituting the score update into the formula for $X_j^{u+1}$ shows that $X_j^{u+1}=\tilt{X_j^u}{-z(2\Lambda_j^u-\Lambda_j^{u-1})}$, which for $z=\eta$ is precisely the update \eqref{eq:OH} with the same losses. Since $X_j^0=p_j=x_j^{t_0}$ and $\Lambda_j^{-1}=v_j=\ell_j^{t_0-1}$, induction gives
\begin{equation}\label{eq:match}
 X_j^u(\eta)=x_j^{t_0+u},\qquad\Lambda_j^u(\eta)=\ell_j^{t_0+u}\qquad(0\le u\le N),
\end{equation}
that is, at $z=\eta$ we obtain the actual trajectory exactly. At $z=0$ nothing moves: $X_j^u(0)=p_j$ and $\Lambda_j^u(0)=\loss_j(p)$ for all $u\ge0$.

\emph{Step 2: the coefficient of $z^k$ is a polynomial of degree $k$ in local time.}
Write
\[
 Y_j^u(z)=\sum_{k\ge1}y_{j,k}(u)z^k,\qquad X_j^u(z)=\sum_{k\ge0}\xi_{j,k}(u)z^k,\qquad \Lambda_j^u(z)=\sum_{k\ge0}c_{j,k}(u)z^k .
\]
We claim that for each $k\le d$ there are polynomials $\hat y_{j,k},\hat\xi_{j,k},\hat c_{j,k}$ of degree at most $k$ (coordinatewise) that agree with $y_{j,k},\xi_{j,k},c_{j,k}$ at every integer $u$ with $k\le u\le N$. We prove this by induction on $k$, simultaneously for all players. For $k=0$ the coefficients are the constants $p_j$ and $\loss_j(p)$. Suppose the claim holds below order $k$. Extracting the coefficient of $z^k$ from \eqref{eq:localprocess} gives
\begin{equation}\label{eq:coefrec}
 y_{j,k}(u+1)-y_{j,k}(u)\;=\;-2c_{j,k-1}(u)+c_{j,k-1}(u-1).
\end{equation}
For $k\le u<N$ both arguments on the right are at least $k-1$, so the right-hand side coincides with the polynomial $H(u)=-2\hat c_{j,k-1}(u)+\hat c_{j,k-1}(u-1)$ of degree at most $k-1$. Expand $H$ in the binomial basis, $H(u)=\sum_{h<k}\alpha_h\binom uh$, and set
\[
 \hat y_{j,k}(u)\;=\;y_{j,k}(k)+\sum_{h<k}\alpha_h\Bigl[\binom u{h+1}-\binom k{h+1}\Bigr].
\]
By Pascal's rule $\binom{u+1}{h+1}-\binom u{h+1}=\binom uh$, this polynomial has degree at most $k$, increments $H(u)$, and the correct value at $u=k$; telescoping \eqref{eq:coefrec} from $u=k$ shows that it agrees with $y_{j,k}(u)$ for all integers $k\le u\le N$. 

The coefficient of $z^k$ in $e^{Y_j^u(a;z)}$ is 
$$
A_{j,k}(u,a) = \sum_{q\ge1}\frac1{q!}\sum_{h_1+\dots+h_q=k,\ h_\nu\ge1}\prod_\nu y_{j,h_\nu}(u,a),
$$ and replacing each $y_{j,h}$ by $\hat y_{j,h}$ gives a polynomial of degree at most $k$ that agrees with it for $u\ge k$, since every $h_\nu\le k\le u$. Setting
$$
B_{j,k}(u) = \sum_a p_j(a) A_{j,k}(u,a), \qquad B_{j,0}=1,
$$
then $B_{j,k}$ has degree at most $k$ and agrees with the coefficient of $z^k$ in $D_j^u$ for $u \geq k$. Defining recursively
\begin{align*}
    E_{j,k}(u) = - \sum_{h=1}^k B_{j,h}(u) E_{j,k-h}(u), \qquad E_{j,0} = 1,
\end{align*}
by the recursive definition of the formal coefficients of $(D_j^u)^{-1}$ we deduce that $E_{j,k}$ has degree at most $k$ and agrees with the coefficient of $z^k$ in the expansion of $(D_j^{u})^{-1}$ for all $u \geq k$. Therefore, defining
\begin{align*}
    \hat{\xi}_{j,k}(u,a) = p_j(a) \sum_{h=0}^k A_{j,h}(u,a)E_{j,k-h}(u),
\end{align*}
gives a polynomial of degree at most $k$ which agrees with $\xi_{j,k}(u,a)$ for all $k \leq u \leq N$. To obtain the coefficients for the losses, define 
\begin{align*}
    \hat{c}_{i,k}(u,a) = \sum_{a_{-i}}L_i(a,a_{-i}) \sum_{h_j} \prod_{j \neq i} \hat{\xi}_{j,h_j}(u,a_j),
\end{align*}
where the second summation runs over $h_j \geq 0$ ($j \neq i$) with $\sum_{j \neq i} h_j = k$. It is straightforward to see that this is a polynomial of degree at most $k$ which agrees with the loss coefficients for all $u \geq k$.

\emph{Step 3: a disk of analyticity and the Taylor tail.}
Let $R=1/(64\sqrt m\,N)$, $b=1/(8\sqrt m)$ and $M=e^{1/64}<2$. We show by induction on $u\le N$ that all quantities in \eqref{eq:localprocess} up to local time $u$ are analytic on a neighborhood of the closed disk $\{|z|\le R\}$ and satisfy
\begin{equation}\label{eq:diskbound}
 \max_{j,a}\bigl|Y_j^u(a;z)\bigr|\le6uR<b,\qquad\max_{j,a}\bigl|\Lambda_j^u(a;z)\bigr|\le M .
\end{equation}
At $u=0$ all quantities are constant, and the losses, including $\Lambda_j^{-1}=v_j$, have entries in $[0,1]$. Suppose $\bigl|Y_j^u\bigr|\le6uR<b$ holds at time $u$. Then $|\Ima Y_j^u|\le b\le\frac18$, so by Lemma~\ref{lem:normalization} $D_j^u(z)\ne0$ on the disk; more precisely $\Rea D_j^u\ge e^{-b}\cos b>0$, so by continuity and compactness $D_j^u$ is nonzero on a neighborhood of the disk, on which $X_j^u$ and $\Lambda_j^u$ are therefore analytic, and Lemma~\ref{lem:normalization} gives $|\Lambda_j^u|\le e^{(m-1)b^2}\le e^{1/64}=M$. The score update then gives $|Y_j^{u+1}|\le|Y_j^u|+|z|(2|\Lambda_j^u|+|\Lambda_j^{u-1}|)\le6uR+6R$, and $6NR=\frac{3}{32\sqrt m}<b$ closes the induction. Intersecting the finitely many neighborhoods involved, all quantities are analytic on one neighborhood of the disk. By Cauchy's estimate, $|c_{i,k}(u,a)|\le2R^{-k}$ for all $k$, and since $\eta\le1/(256\sqrt mN)=R/4$, Lemma \ref{lem:cauchytail} gives, uniformly in $u\le N$ and $a$,
\begin{equation}\label{eq:localtail}
 \Bigl|\Lambda_i^u(a;\eta)-\sum_{k=0}^d c_{i,k}(u,a)\eta^k\Bigr|\;\le\;2\sum_{k>d}4^{-k}\;=\;\tfrac23\cdot4^{-d} \leq 2^{-d}.
\end{equation}

\emph{Step 4: the polynomial.}
Define $q_{i,r,a}(s)=\sum_{k\le d}\eta^k\hat c_{i,k}(d+s,a)$, which is a polynomial of degree at most $d$ in $s$ with real coefficients. For every integer $0\le s\le W$ the local time $u=d+s$ satisfies $k\le u\le N$ for all $k\le d$, so $q_{i,r,a}(s)=\sum_{k\le d}c_{i,k}(d+s,a)\eta^k$, while $\ell_i^{r+s}=\Lambda_i^{d+s}(\eta)$ by \eqref{eq:match} because $t_0+d+s=r+s$. Inequality \eqref{eq:localtail} completes the proof.
\end{proof}

\subsection{Polynomial variation--variance inequality}\label{sec:log-energy}

\begin{lemma}[Variation--variance inequality for polynomials]\label{lem:polyenergy}
Let $W\ge4$ be divisible by four, let $0\le d\le W$, and let $\cI_W=\{W/4,W/4+1,\dots,3W/4-1\}$. Then every real polynomial $q : \R \to \R$ of degree at most $d$ satisfies
\begin{equation}\label{eq:polyenergy}
 \sum_{s\in\cI_W}\bigl(q(s+1)-q(s)\bigr)^2\;\le\;\frac{16\,d(d+1)}{3W^2}\sum_{s=0}^Wq(s)^2.
\end{equation}
In particular, if each entry of $q: \R \to \R^n$ is a polynomial of degree at most $d$ and $p \in \Delta_n$,
\begin{equation}\label{eq:polyenergy-var}
 \sum_{s\in\cI_W}\norm{q(s+1) - q(s)}^2_p\;\le\;\frac{16\,d(d+1)}{3W^2}\sum_{s=0}^W\norm{q(s)}^2_p.
\end{equation}
\end{lemma}

\begin{proof}
Let $w_s=(s+1)(W-s)$ for $0\le s<W$ and $w_{-1}=w_W=0$; Define the symmetric operator $K$ on $\R^{W+1}$ by $(Kv)_s=w_s(v_s-v_{s+1})+w_{s-1}(v_s-v_{s-1})$, where terms with indices outside $\{0,\dots,W\}$ are omitted. Summation by parts gives
\[
 \ip v{Kv}\;=\;\sum_{s=0}^{W-1}w_s(v_{s+1}-v_s)^2 .
\]
Let $\cP_d\subseteq\R^{W+1}$ be the space of vectors $(q(0),\dots,q(W))$ with $q$ a polynomial of degree at most $d$; since $d\le W$, such a vector determines $q$. We claim that $K$ maps $\cP_d$ into itself and acts on monomials by $K(s^k)=k(k+1)s^k+(\text{a polynomial of degree}\le k-1)$ for $k\ge1$, and $K1=0$. Indeed, writing $(Kq)(s)=-w_s\bigl(q(s+1)-q(s)\bigr)+w_{s-1}\bigl(q(s)-q(s-1)\bigr)$ and expanding $q(s\pm1)=s^k\pm ks^{k-1}+\binom k2s^{k-2}\pm\cdots$, the first-order terms contribute $ks^{k-1}(w_{s-1}-w_s)=ks^{k-1}(2s-W)$, with leading coefficient $2k$, and the second-order terms contribute $-\binom k2s^{k-2}(w_s+w_{s-1})$, whose leading coefficient is $k(k-1)$ because the coefficient of $s^2$ in $w_s+w_{s-1}$ is $-2$; the remaining terms have degree at most $k-1$, and the vanishing of $w_{-1}$ and $w_W$ makes the computation valid at the endpoints. Hence $K$ restricted to $\cP_d$ is triangular in the basis of monomials with diagonal entries $k(k+1)$, $0\le k\le d$, which are therefore its eigenvalues on $\cP_d$. Being symmetric, $K$ restricted to the invariant subspace $\cP_d$ is self-adjoint, so $\ip v{Kv}\le d(d+1)\ip vv$ for $v\in\cP_d$, that is,
\[
 \sum_{s=0}^{W-1}w_s\bigl(q(s+1)-q(s)\bigr)^2\;\le\;d(d+1)\sum_{s=0}^Wq(s)^2 .
\]
For $s\in\cI_W$ the number $\theta_s=(s+1)/(W+1)$ lies in $[\frac14,\frac34]$, so $w_s=(W+1)^2\theta_s(1-\theta_s)\ge\frac3{16}(W+1)^2\ge\frac3{16}W^2$; dropping the terms with $s\notin\cI_W$ and dividing by $\frac3{16}W^2$ gives \eqref{eq:polyenergy}. For \eqref{eq:polyenergy-var}, apply \eqref{eq:polyenergy} to each centered coordinate $s\mapsto q_a(s)-\ip p{q(s)}$, which has degree at most $d$, multiply by $p(a)$ and sum over $a$.
\end{proof}

\subsection{Sliding windows}\label{sec:log-windows}

The next lemma converts a temporal (or local) approximation by polynomials into the global variation--variance inequality.

\begin{lemma}[Sliding-window comparison]\label{lem:windows}
Let $T\ge1$, $d\ge0$ and $r_0\ge1$ be integers, let $W\ge4(d+1)$ be divisible by four, and let $\eps\ge0$. Let $Z^0,\dots,Z^T\in[0,1]^n$ and let $P^1,\dots,P^T$ be probability vectors with full support such that, for some $\kappa\ge0$ with $e^{2\kappa W}\le2$,
\begin{equation}\label{eq:ratios}
 e^{-\kappa}\le\frac{P^{t+1}(a)}{P^t(a)}\le e^{\kappa}\qquad(1\le t<T,\ a\in[n]).
\end{equation}
Suppose that for every integer $r$ with $r_0\le r\le T-W$ there exists $q_r : \R \to \R^n$ in which every entry is a polynomial of degree at most $d$ with $\max_{0\le s\le W}\norm{Z^{r+s}-q_r(s)}_\infty\le\eps$. Then
\begin{equation}\label{eq:windows}
 \sum_{t=1}^T\norm{Z^t - Z^{t-1}}^2_{P^t}\;\le\;\frac{256\,d(d+1)(W+1)}{3W^3}\Bigl(\sum_{t=1}^T\norm{Z^{t-1}}_{P^t}^2+\frac14\Bigr)+r_0+2W+32T\eps^2 .
\end{equation}
\end{lemma}

\begin{proof}
Write $A_t=\norm{Z^t - Z^{t-1}}^2_{P^t}$, $V_t=\norm{Z^{t-1}}^2_{P^t}$, $A=\sum_{t\le T}A_t$, $V=\sum_{t\le T}V_t$ and $\lambda=16d(d+1)/(3W^2)$. The hypothesis $W\ge4(d+1)$ gives $d(d+1)\le W^2/16$ and hence $\lambda(W+1)\le(W+1)/3\le W$. For bookkeeping set $P^{T+1}=P^T$ and $V_{T+1}=\norm{Z^T}_{P^t}^2\le\frac14$.

\emph{One window with frozen weights.} Fix $b$ with $r_0+1\le b\le T-W+1$ and let $z_s=Z^{b-1+s}$, $q_s=q_{b-1}(s)$ and $e_s=z_s-q_s$ for $0\le s\le W$, so that $\norm{e_s}_\infty\le\eps$. For a fixed probability vector $p$, the inequality $\norm{u+v}^2_p \le2\norm{u}^2_p+2\norm{v}_p^2$, Lemma~\ref{lem:polyenergy} and $|\cI_W|=W/2$ give
\begin{align*}
 \sum_{s\in\cI_W}\norm{z_{s+1} - z_s}^2_p&\le2\sum_{s\in\cI_W}\norm{q_{s+1}-q_s}^2_p+4W\eps^2
 \le2\lambda\sum_{s=0}^W\norm{q_s}^2_p+4W\eps^2\\
 &\le4\lambda\sum_{s=0}^W\norm{z_s}^2_p+\bigl(4\lambda(W+1)+4W\bigr)\eps^2 .
\end{align*}
\emph{Restoring the time-varying weights.} Take $p=P^b$. By \eqref{eq:ratios}, the coordinates of $P^b$ and $P^{b+s}$ differ by a factor at most $\Gamma=e^{\kappa W}$ for $0\le s\le W$, and $\Gamma^2\le2$. Since $z_{s+1}-z_s=Z^{b+s}-Z^{b+s-1}$ and $z_s=Z^{b+s-1}$ are the vectors whose variances under $P^{b+s}$ are $A_{b+s}$ and $V_{b+s}$, the fact that $\Var_{p'}\le\Gamma\Var_p$, applied once on each side, gives
\begin{equation}\label{eq:movingwindow}
 \sum_{s\in\cI_W}A_{b+s}\;\le\;4\lambda\Gamma^2\sum_{s=0}^WV_{b+s}+\Gamma\bigl(4\lambda(W+1)+4W\bigr)\eps^2\;\le\;8\lambda\sum_{s=0}^WV_{b+s}+16W\eps^2 ,
\end{equation}
using $\lambda(W+1)\le W$ in the last step.

\emph{Summing over windows.} Each $A_t$ is trivially at most $1$. If $T<r_0+2W$ then $A\le T<r_0+2W$ and we are done, so assume $T\ge r_0+2W$ and sum \eqref{eq:movingwindow} over all admissible $b$. A time step $t$ with $r_0+\frac34W\le t\le T-\frac34W+1$ occurs on the left-hand side once for every $s\in\cI_W$, with $b=t-s$, and these $W/2$ values of $b$ are all admissible; every other $t$ occurs at most $W/2$ times, and there are at most $r_0+\frac32W\le r_0+2W$ such $t$, contributing at most $r_0+2W$ to $A$ in total. Hence the left-hand sides add up to at least $\frac W2(A-r_0-2W)$. On the right, each $V_t$ appears in at most $W+1$ windows and there are at most $T$ windows, so
\[
 \frac W2\bigl(A-r_0-2W\bigr)\;\le\;8\lambda(W+1)\Bigl(V+\frac14\Bigr)+16WT\eps^2 .
\]
Dividing by $W/2$ and substituting $\lambda$ gives \eqref{eq:windows}.
\end{proof}

\subsection{Completion of the proof}\label{sec:log-completion}

\begin{proof}[Proof of Theorem~\ref{thm:log}]
Fix a player $i$. The vector $2\ell_i^t-\ell_i^{t-1}$ has entries in $[-1,2]$, so by \eqref{eq:OH} consecutive strategies satisfy $e^{-3\eta}\le x_i^{t+1}(a)/x_i^t(a)\le e^{3\eta}$. We apply Lemma~\ref{lem:windows} with $Z^t=\ell_i^t$, $P^t=x_i^t$, $\kappa=3\eta$, $r_0=d+1$, $\eps=2^{-d}$ and the polynomials of Lemma~\ref{lem:approx}, whose hypothesis is the learning-rate condition of the theorem. The window $W=32(d+1)$ is divisible by four and satisfies $W\ge4(d+1)$, and
\[
 2\kappa W=6\eta W\le\frac{6}{256\sqrt m}\cdot\frac{W}{W+d}<\log2 ,
\]
so \eqref{eq:ratios} holds. Lemma~\ref{lem:windows} yields
\[
 D_i(T)\;\le\;\beta\Bigl(V_i(T)+\frac14\Bigr)+(d+1)+2W+32T4^{-d},\qquad
 \beta=\frac{256\,d(d+1)(W+1)}{3W^3},
\]
and the choice $W=32(d+1)$ gives $\beta=\frac{d}{12(d+1)}\bigl(1+\frac1{32(d+1)}\bigr)<\frac1{12}$.
Since $4^d\ge(T+1)^2$ we have $32T4^{-d}\le32$, and since $d\ge1$ we have $(d+1)+32+\frac1{48}\le32(d+1)=W$; therefore
\begin{equation}\label{eq:AVcomparison}
 D_i(T)\;\le\;\tfrac14V_i(T)+3W .
\end{equation}
The learning-rate condition implies $\eta\le1/8$, so Lemma~\ref{lem:potential} and \eqref{eq:AVcomparison} give
\[
 \Reg_i(T)\;\le\;\frac{\log n_i}{\eta}+\frac{2\eta}{3}\Bigl(\frac14V_i(T)+3W\Bigr)-\frac\eta3V_i(T)
 \;=\;\frac{\log n_i}{\eta}-\frac\eta6V_i(T)+2\eta W\;\le\;\frac{\log n_i}{\eta}+2\eta W .
\]
For $\eta=1/(256\sqrt m(W+d))$ we have $2\eta W\le1/(128\sqrt m)\le1$ and $W+d=33d+32=O(\log(T+2))$, which gives the stated rate.
\end{proof}

The following is an auxiliary section containing useful lemmas from complex analysis which allow us to bound the truncation error of the loss expansions.

\subsection{Complex exponential weights}\label{sec:log-complex}

The approximation lemma expands the dynamics in a complex learning rate. The following estimate is the reason the expansion converges on a disk whose radius degrades only as $m^{-1/2}$ with the number of players.

\begin{lemma}[Complex normalization]\label{lem:normalization}
Let $p\in\simplex_n$ and let $y\in\C^n$ satisfy $|\Ima y(a)|\le b\le\frac12$ for every $a$. Then $Z=\sum_ap(a)e^{y(a)}$ is nonzero, and the complex vector $q(a)=p(a)e^{y(a)}/Z$ satisfies
\begin{equation}\label{eq:complexnorm}
 \sum_a|q(a)|\;\le\;\frac1{\cos b}\;\le\;e^{b^2}.
\end{equation}
If the strategy of every opponent $j\ne i$ is a complex vector $q_j$ of this form, with the same $b$, then the loss map \eqref{eq:lossmap} extended multilinearly satisfies $\norm{\loss_i(q)}_\infty\le\prod_{j\ne i}\sum_a|q_j(a)|\le e^{(m-1)b^2}$.
\end{lemma}

\begin{proof}
Write $y(a)=\alpha_a+i\theta_a$. Then $\Rea Z=\sum_ap(a)e^{\alpha_a}\cos\theta_a\ge\cos b\sum_ap(a)e^{\alpha_a}>0$, so $Z\ne0$ and
\[
 \sum_a|q(a)|=\frac{\sum_ap(a)e^{\alpha_a}}{|Z|}\le\frac{\sum_ap(a)e^{\alpha_a}}{\Rea Z}\le\frac1{\cos b}.
\]
For $0\le t\le\frac12$ we have $\sin t\le t$ and $\cos t\ge1-t^2/2\ge\frac78$, hence $\tan t\le2t$ and $-\log\cos b=\int_0^b\tan t\,dt\le b^2$. For the loss map, $|\loss_i(q)(a_i)|\le\sum_{a_{-i}}L_i(a_i,a_{-i})\prod_{j\ne i}|q_j(a_j)|\le\prod_{j\ne i}\sum_{a_j}|q_j(a_j)|$, where the last step distributes the product over the opponents' action sets and uses $L_i\le1$.
\end{proof}

The following is a classical result from complex analysis which follow from the Cauchy integral estimate of analytic functions:

\begin{lemma}[Analytic function tail bound]
\label{lem:cauchytail}
    Let $f : \C \to \R$ be analytic in a neighborhood of the closed disk $\{ |z| \leq R \}$ and let $M>0$ be a constant for which $|f(z)| \leq M$ for all $|z| \leq R$. Then the Taylor expansion $f(z) = \sum_{k=0}^\infty f_k z^k$ satisfies, for all $0 \leq \eta < R$:
    \begin{equation}
         \Bigl|f(\eta)-\sum_{k=0}^df_k\eta^k\Bigr| \le 
         \frac{M(\eta/R)^{d+1}}{1-\eta/R}.
    \end{equation}
\end{lemma}

\section{Proof of the constant regret upper bound for two players}\label{sec:two}

\paragraph{Setting and notation.}
Throughout this section there are two players with $n\ge2$ actions each and loss matrices $A,B\in[0,1]^{n\times n}$: when player~$1$ plays $a$ and player~$2$ plays $b$, their losses are $A_{ab}$ and $B_{ab}$. As in Section~\ref{sec:ov-two}, we write $x_1^t,x_2^t\in\simplex_n$ for the strategies of players~$1$ and~$2$ in round $t$ and
\[
 \ell_1^t=Ax_2^t,\qquad \ell_2^t=B^\top x_1^t
\]
for the loss vectors, so that the dynamics are $x_1^{t+1}=\tilt{x_1^t}{-\eta(2\ell_1^t-\ell_1^{t-1})}$ and $x_2^{t+1}=\tilt{x_2^t}{-\eta(2\ell_2^t-\ell_2^{t-1})}$, with $x_1^1=x_2^1$ uniform and $\ell_1^0=\ell_2^0=0$. We prove the regret bound for player~$1$; player~$2$ is treated symmetrically. For a fixed horizon $T$ the argument is organised around three sums: the variance of the \emph{previous} losses $\sum_{t=1}^T\Var_{x_1^t}(\ell_1^{t-1})$ and the variance of the \emph{increments} $\sum_{t=1}^T\Var_{x_1^t}(\ell_1^t-\ell_1^{t-1})$, which are the two quantities $V_1(T)$ and $D_1(T)$ appearing in Lemma~\ref{lem:potential}, and the variance of the \emph{current} losses $\sum_{t=1}^T\Var_{x_1^t}(\ell_1^t)$, convenient because it is what controls the movement of player~$1$. We write $v^t=\norm{\ell_1^t}_{x_1^t}=\sqrt{\Var_{x_1^t}(\ell_1^t)}$, with $v^0=0$. Elementary facts about exponential tilting and variances are collected in Lemmas~\ref{lem:tilt} and~\ref{lem:varfacts} of Appendix~\ref{app:elementary}; we use them freely. The letters $C,C',C''$ denote absolute constants.

\subsection{Preliminary inequalities}\label{sec:two-movement}

\begin{lemma}[Movement and variance shift]\label{lem:movement}
Let $0<\eta\le1/100$. For every $t\ge1$,
\begin{equation}\label{eq:ratio}
 e^{-3\eta}\le\frac{x_1^{t+1}(a)}{x_1^t(a)}\le e^{3\eta}\quad(a\in[n]),\qquad
 \norm{x_1^{t+1}-x_1^t}_1\le3\eta\,(v^t+v^{t-1}),
\end{equation}
and
\begin{equation}\label{eq:shift}
 \sum_{t=1}^T\Var_{x_1^t}(\ell_1^t)\;\le\;e^{3\eta}\sum_{t=1}^T\Var_{x_1^t}(\ell_1^{t-1})+\frac14 .
\end{equation}
\end{lemma}

\begin{proof}
The vector $z=-\eta(2\ell_1^t-\ell_1^{t-1})$ satisfies $\max_{a,b\in [n]}z(a)-z(b)\leq 3\eta$, so the density bound \ref{it:density} of Lemma~\ref{lem:tilt} gives the ratio bound. The movement bound \ref{it:movement} of Lemma~\ref{lem:tilt} gives
\[
 \norm{x_1^{t+1}-x_1^t}_1\le e^{3\eta/2}\,\eta\,\norm{2\ell_1^t-\ell_1^{t-1}}_{x_1^t}\le\eta e^{3\eta/2}\bigl(2v^t+\norm{\ell_1^{t-1}}_{x_1^t}\bigr).
\]
For $t\ge2$ the ratio bound between $x_1^t$ and $x_1^{t-1}$ and the comparison fact for variances (Lemma~\ref{lem:varfacts}) give $\norm{\ell_1^{t-1}}_{x_1^t}\le e^{3\eta/2}\norm{\ell_1^{t-1}}_{x_1^{t-1}}=e^{3\eta/2}v^{t-1}$, while for $t=1$ the term vanishes. Hence $\norm{x_1^{t+1}-x_1^t}_1\le2\eta e^{3\eta/2}v^t+\eta e^{3\eta}v^{t-1}\le3\eta(v^t+v^{t-1})$ for $\eta\le1/100$. Finally, for $t<T$ the ratio bound gives $\Var_{x_1^t}(\ell_1^t)\le e^{3\eta}\Var_{x_1^{t+1}}(\ell_1^t)$, and summing over $t<T$ yields
\[
 \sum_{t<T}\Var_{x_1^t}(\ell_1^t)\;\le\;e^{3\eta}\sum_{t<T}\Var_{x_1^{t+1}}(\ell_1^t)\;\le\;e^{3\eta}\sum_{t=1}^T\Var_{x_1^t}(\ell_1^{t-1});
\]
the remaining term $\Var_{x_1^T}(\ell_1^T)$ is at most $\frac14$ because $\ell_1^T\in[0,1]^n$.
\end{proof}

Since $\ell_2^t=B^\top x_1^t$ with entries of $B$ in $[0,1]$, the lemma gives the two-player feedback estimate
\begin{equation}\label{eq:feedback}
 \norm{\ell_2^{t+1}-\ell_2^t}_\infty\;\le\;\norm{x_1^{t+1}-x_1^t}_1\;\le\;3\eta\,(v^t+v^{t-1}):
\end{equation}
the input of player~$2$ moves only when player~$1$ moves, and player~$1$ moves only when their current losses have variance. This is the only place where the number of players matters. With three or more players, $\ell_2^t$ would also depend on the strategy of player~$3$, whose movement is governed by player~$3$'s variance, a quantity that does not appear in player~$1$'s potential inequality.

\paragraph{A Bernstein inequality for exponential sums.}
\begin{lemma}[Interior Bernstein inequality, $L^2$ form]\label{lem:bernstein}
Let $\cK$ be a real Hilbert space, let $N(s)=\sum_{k=1}^nw_ke^{\lambda_ks}$ with $w_k\in\cK$ and $\lambda_k\in\R$, and let $I=[\alpha+W,\alpha+2W]$ be the middle third of $J=[\alpha,\alpha+3W]$ for some $\alpha\in\R$ and $W>0$. Then
\begin{equation}\label{eq:bernstein}
 \Bigl(\int_I\norm{N'(s)}^2ds\Bigr)^{1/2}\;\le\;\frac{2n}{W}\Bigl(\int_J\norm{N(s)}^2ds\Bigr)^{1/2}.
\end{equation}
The same inequality holds for $\cK=\R^n$ equipped with the variance seminorm $\norm\cdot_p$ of any fixed probability vector $p$.
\end{lemma}

\begin{proof}
Borwein and Erd\'elyi \cite{BE96} proved that every real function $g(u)=\mu_0+\sum_{k=1}^n\mu_ke^{\kappa_ku}$ satisfies $|g'(0)|\le\frac{2n-1}{W}\max_{|u|\le W}|g(u)|$. Let $\psi\in L^2(I;\cK)$ have norm one and set $g(u)=\int_I\ip{N(s+u)}{\psi(s)}ds$. Expanding $N$, we see that $g$ is a real exponential sum with exponents $\lambda_1,\dots,\lambda_n$ (and no constant term, which we may regard as $\mu_0=0$). For $|u|\le W$ the translate $I+u$ is contained in $J$, so the Cauchy--Schwarz inequality gives $|g(u)|\le\norm N_{L^2(J;\cK)}$. Since $g'(0)=\int_I\ip{N'(s)}{\psi(s)}ds$, the scalar inequality gives $\int_I\ip{N'}{\psi}\le\frac{2n}W\norm N_{L^2(J;\cK)}$, and taking the supremum over $\psi$ proves \eqref{eq:bernstein}. 

\emph{The variance seminorm.} Let $p\in\simplex_n$ be fixed and recall that
\[
 \norm v_p^2=\Var_p(v)=\sum_{h=1}^np(h)\bigl(v(h)-\ip pv\bigr)^2\qquad(v\in\R^n).
\]
This is only a seminorm on $\R^n$: it vanishes on every vector that is constant on the support of $p$, so $\R^n$ with $\norm\cdot_p$ is not a Hilbert space and \eqref{eq:bernstein} cannot be invoked for it directly. It is, however, the Euclidean norm of a \emph{linear} image of $v$. Write $P=\operatorname{diag}(p)$ and define the centering map
\[
 \Pi_p:\R^n\to\R^n,\qquad \Pi_pv=P^{1/2}\bigl(v-\ip pv\,\mathbf 1\bigr),
\]
that is, $(\Pi_pv)(h)=\sqrt{p(h)}\,\bigl(v(h)-\ip pv\bigr)$. Then
\[
 \norm{\Pi_pv}_2^2=\sum_{h=1}^np(h)\bigl(v(h)-\ip pv\bigr)^2=\norm v_p^2\qquad(v\in\R^n),
\]
so $\Pi_p$ carries the seminorm $\norm\cdot_p$ isometrically onto the standard norm of $\R^n$. Two properties of $\Pi_p$ are all that the argument needs. First, $\Pi_p$ is linear: the centering $v\mapsto v-\ip pv\mathbf1$ is the linear map $\mathrm{Id}-\mathbf1p^\top$, and $P^{1/2}$ is a fixed diagonal matrix. Second, $\Pi_p$ does not depend on the time variable $s$, because $p$ is fixed. Together these mean that $\Pi_p$ commutes with the two operations we perform on the curve $N$, namely forming it as an exponential sum and differentiating it in $s$: for $N(s)=\sum_{k=1}^nw_ke^{\lambda_ks}$,
\[
 \Pi_pN(s)=\sum_{k=1}^n(\Pi_pw_k)\,e^{\lambda_ks},\qquad
 (\Pi_pN)'(s)=\Pi_p\bigl(N'(s)\bigr)=\sum_{k=1}^n\lambda_k(\Pi_pw_k)\,e^{\lambda_ks}.
\]
Thus $\Pi_pN$ is again an exponential sum with the same exponents $\lambda_1,\dots,\lambda_n$ and at most $n$ terms, now with coefficients $\Pi_pw_k$ in the Hilbert space $\R^n$ equipped with its standard inner product, and its derivative is the image of $N'$ under the same map. (The number of terms does not grow, which matters because the constant $2n$ in \eqref{eq:bernstein} is the number of exponentials.) Applying \eqref{eq:bernstein} with $\cK=\R^n$ to the exponential sum $\Pi_pN$ and translating back through the isometry gives
\[
 \Bigl(\int_I\norm{N'(s)}_p^2\,ds\Bigr)^{1/2}
 =\Bigl(\int_I\norm{(\Pi_pN)'(s)}_2^2\,ds\Bigr)^{1/2}
 \le\frac{2n}W\Bigl(\int_J\norm{\Pi_pN(s)}_2^2\,ds\Bigr)^{1/2}
 =\frac{2n}W\Bigl(\int_J\norm{N(s)}_p^2\,ds\Bigr)^{1/2},
\]
which is \eqref{eq:bernstein} for the variance seminorm. Equivalently, $\Pi_p$ identifies the quotient of $\R^n$ by the kernel of $\norm\cdot_p$ with a subspace of $\R^n$; the explicit map makes visible that the exponential-sum structure survives the passage to the quotient.

Note that the argument uses in an essential way that $p$ does not vary with $s$: with time-dependent weights the map $\Pi_{p(s)}$ would no longer commute with differentiation, and $\Pi_{p(s)}N(s)$ would not be an exponential sum. This is why the weights are frozen at $p=x_1^{t_0}$ in the proof of Lemma~\ref{lem:local}, and the true weights $x_1^{\lfloor s\rfloor}$ are restored only afterwards, through the ratio bound of Lemma~\ref{lem:movement}.
\end{proof}

The inequality is \emph{interior}: it bounds the derivative only on the middle third, and no bound of this kind holds up to the endpoints. This is why the windows used in proving the properties of the temporal approximation framework in continuous time have length $3W$ and only their middle thirds are used.

\subsection{Proof of Theorem~\ref{thm:two}}\label{sec:two-main}

As stated in the proof overview, we use the temporal approximation framework analogously on the \emph{interpolation of the iterates in continuous time}. As we prove in Section~\ref{sec:two-transfer}, by reducing the resulting bounds on the continuous terms to the discrete terms, we obtain a sufficient variance bound.

\paragraph{Defining the temporal approximation framework and the resulting variance bound.}
If player~$2$'s losses stayed equal to $\ell_2^{t_0}$ from some round $t_0$ on, their strategy would follow $\tilt{x_2^{t_0}}{-\eta(s-t_0)\ell_2^{t_0}}$, and player~$1$'s losses $A\cdot \tilt{x_2^{t_0}}{-\eta(s-t_0)\ell_2^{t_0}}$ would be a ratio of two sums of $n$ exponentials in $s$. Lemma~\ref{lem:bernstein} would then bound the increments of player~$1$'s losses over the middle third of a window of length $3W$ by $O(n/W)$ times their size over the window. Player~$2$'s losses do move, but by the feedback estimate \eqref{eq:feedback} only as fast as player~$1$'s current losses have variance, which is what the potential inequality allows us to charge. The following lemma makes this precise. Its statement involves only the discrete trajectory; continuous time enters only its proof, given in Section~\ref{sec:two-transfer}.

\begin{lemma}[Variance bound resulting from the temporal approximation framework]\label{lem:transfer}
There are absolute constants $C$ and $K_0$ such that for every integer $W\ge K_0n$ and every $\eta$ with $0<\eta W\le10^{-3}$,
\begin{equation}\label{eq:transfer}
 \sum_{t=1}^T\Var_{x_1^t}(\ell_1^t-\ell_1^{t-1})\;\le\;\frac14+C\Bigl[\Bigl(\frac{n^2}{W^2}+\eta^4n^2W^2\Bigr)\Bigl(\sum_{t=1}^T\Var_{x_1^t}(\ell_1^{t-1})+1\Bigr)+\eta^2W\Bigr].
\end{equation}
\end{lemma}

In the bracket, the factor $n^2/W^2$ comes from the Bernstein inequality of Lemma~\ref{lem:bernstein}, the factor $\eta^4n^2W^2$ is the freezing error, paid by the variance of the current losses through the feedback estimate \eqref{eq:feedback}, and the term $\eta^2W$ is the cost of the two ends of the horizon.

Tuning $\eta,W$ we may reduce the variance bound to the necessary constants:

\begin{lemma}[Two-player variation--variance inequality]\label{lem:smooth}
There is an absolute constant $c>0$ such that if both players run Optimistic Hedge with $0<\eta\le c/n$, then
\begin{equation}\label{eq:smooth}
 \sum_{t=1}^T\Var_{x_1^t}(\ell_1^t-\ell_1^{t-1})\;\le\;\frac1{64}\sum_{t=1}^T\Var_{x_1^t}(\ell_1^{t-1})+\frac12 .
\end{equation}
\end{lemma}

\begin{proof}
Let $C$ and $K_0$ be the constants of Lemma~\ref{lem:transfer}. Let $K\ge K_0$ be an integer, to be chosen below, and put $W=Kn$. If $\eta\le c/n$ with $cK\le10^{-3}$, Lemma~\ref{lem:transfer} applies and gives
\[
 \sum_{t=1}^T\Var_{x_1^t}(\ell_1^t-\ell_1^{t-1})\;\le\;\frac14+C\bigl(K^{-2}+c^4K^2\bigr)\Bigl(\sum_{t=1}^T\Var_{x_1^t}(\ell_1^{t-1})+1\Bigr)+Cc^2K ,
\]
because $n^2/W^2=K^{-2}$, $\eta^4n^2W^2\le c^4K^2$ and $\eta^2W\le c^2K/n\le c^2K$. Choose $K$ so large that $C/K^2\le\frac1{128}$, and then $c$ so small that $Cc^4K^2\le\frac1{128}$, $Cc^2K\le\frac18$, $cK\le10^{-3}$ and $c\le\frac1{100}$; the last condition gives $\eta\le\frac1{100}$, as required by Lemmas~\ref{lem:movement} and~\ref{lem:potential}. The coefficient of $\sum_{t=1}^T\Var_{x_1^t}(\ell_1^{t-1})$ is then at most $\frac1{64}$ and the additive part at most $\frac14+\frac1{64}+\frac18<\frac12$.
\end{proof}

\begin{proof}[Proof of Theorem~\ref{thm:two}]
Let $\eta\le c/n$ with $c$ as in Lemma~\ref{lem:smooth}; in particular $\eta\le1/100$. For player~$1$, Lemma~\ref{lem:potential} and \eqref{eq:smooth} give
\[
 \Reg_1(T)\;\le\;\frac{\log n}{\eta}+\frac{2\eta}{3}\Bigl(\frac1{64}\sum_{t=1}^T\Var_{x_1^t}(\ell_1^{t-1})+\frac12\Bigr)-\frac\eta3\sum_{t=1}^T\Var_{x_1^t}(\ell_1^{t-1})\;\le\;\frac{\log n}{\eta}+\frac\eta3\;\le\;\frac{\log n}{\eta}+\eta ,
\]
since $\sum_{t=1}^T\Var_{x_1^t}(\ell_1^{t-1})\ge0$ has a negative coefficient. The same argument with the roles of the players exchanged bounds $\Reg_2(T)$.
\end{proof}

\subsection{Proof of Lemma~\ref{lem:transfer}: interpolation and windows}\label{sec:two-transfer}

Throughout this subsection $W\ge1$ is an integer and $0<\eta W\le10^{-3}$; in particular $\eta\le1/100$, so Lemma~\ref{lem:movement} applies. We first build the continuous objects and prove two lemmas about them, and then sum over windows.

\paragraph{The interpolation.}
Continuous time enters only through the following interpolation of player~$2$'s trajectory. For every integer $j\ge1$ and $s\in[j,j+1]$ let $g^j=2\ell_2^j-\ell_2^{j-1}$ be the score direction of round $j$ and
\begin{equation}\label{eq:interp}
 Y(s)=\tilt{x_2^j}{-\eta(s-j)g^j},\qquad L(s)=AY(s).
\end{equation}
Then $Y(j)=x_2^j$, $Y(j+1)=x_2^{j+1}$ and $L(j)=\ell_1^j$; the curves are smooth on each unit interval, which is all that the following integrals require. The proof works with two continuous energies of the interpolated losses, the \emph{position energy} $\int_1^T\Var_{x_1^{\lfloor s\rfloor}}(L(s))\,ds$ and the \emph{derivative energy} $\int_1^T\Var_{x_1^{\lfloor s\rfloor}}(L'(s))\,ds$, in both of which the variance weights are those of player~$1$ in the round containing $s$. They are the continuous counterparts of the variance of the current losses and of the variance of the increments.

\begin{lemma}[Discrete--continuous comparisons]\label{lem:comparison}
For every horizon $T$,
\begin{equation}\label{eq:comp1}
 \sum_{t=1}^T\Var_{x_1^t}(\ell_1^t-\ell_1^{t-1})\;\le\;\frac14+e^{3\eta}\int_1^T\Var_{x_1^{\lfloor s\rfloor}}\bigl(L'(s)\bigr)ds
\end{equation}
and
\begin{equation}\label{eq:comp2}
 \int_1^T\Var_{x_1^{\lfloor s\rfloor}}\bigl(L(s)\bigr)ds\;\le\;2\sum_{t=1}^T\Var_{x_1^t}(\ell_1^t)+2\int_1^T\Var_{x_1^{\lfloor s\rfloor}}\bigl(L'(s)\bigr)ds .
\end{equation}
\end{lemma}

\begin{proof}
The $t=1$ term on the left of \eqref{eq:comp1} is $\Var_{x_1^1}(\ell_1^1)\le\frac14$. For $j\ge1$ we have $\ell_1^{j+1}-\ell_1^j=\int_j^{j+1}L'(s)ds$, and Jensen's inequality for the Hilbert seminorm $\norm\cdot_{x_1^j}$ gives $\Var_{x_1^j}(\ell_1^{j+1}-\ell_1^j)\le\int_j^{j+1}\Var_{x_1^j}(L'(s))ds$. Since $x_1^{j+1}\le e^{3\eta}x_1^j$ coordinatewise, $\Var_{x_1^{j+1}}(\ell_1^{j+1}-\ell_1^j)\le e^{3\eta}\Var_{x_1^j}(\ell_1^{j+1}-\ell_1^j)$, and summing over $1\le j\le T-1$ proves \eqref{eq:comp1}. For \eqref{eq:comp2}, write $L(s)=\ell_1^j+\int_j^sL'(u)du$ for $s\in[j,j+1]$; the triangle inequality, $(\alpha+\beta)^2\le2\alpha^2+2\beta^2$ and the Cauchy--Schwarz inequality on the unit interval give $\int_j^{j+1}\Var_{x_1^j}(L(s))ds\le2\Var_{x_1^j}(\ell_1^j)+2\int_j^{j+1}\Var_{x_1^j}(L'(s))ds$, and summing over $1\le j\le T-1$ proves \eqref{eq:comp2}.
\end{proof}

The second comparison \eqref{eq:comp2} carries the derivative energy on its right-hand side, which is the quantity we are trying to bound. This is harmless, because after the windows are summed the derivative energy appears on the right only with a small coefficient and can be absorbed into the left-hand side (proof of Lemma~\ref{lem:transfer}).

\paragraph{Windows and the frozen curve.}
Fix an integer $W\ge1$ and an integer anchor $t_0\ge2$ with $t_0+3W\le T$ (so that the sums below, which start at $r=t_0-1$, involve only rounds $r\ge1$). Let $J_{t_0}=[t_0,t_0+3W]$ be the window and $I_{t_0}=[t_0+W,t_0+2W]$ its middle third, and measure the movement of player~$2$'s losses over the window by
\[
 \delta_{t_0}=\sum_{r=t_0-1}^{t_0+3W-2}\norm{\ell_2^{r+1}-\ell_2^r}_\infty,\qquad u_{t_0}=\sum_{r=t_0-1}^{t_0+3W-2}\norm{\ell_2^{r+1}-\ell_2^r}_\infty^2,\qquad\text{so that }\delta_{t_0}^2\le3W\,u_{t_0} .
\]
Freezing player~$2$'s losses at $\ell_2^{t_0}$ defines the comparison curves $\bar Y(s)=\tilt{x_2^{t_0}}{-\eta(s-t_0)\ell_2^{t_0}}$ and $\bar L(s)=A\bar Y(s)$ for $s\in J_{t_0}$. For a curve $F$ with values in $\R^n$, an interval $I$ and a probability vector $p$, we write $\norm F_{L^2(I;p)}=\bigl(\int_I\norm{F(s)}_p^2\,ds\bigr)^{1/2}$.

\begin{lemma}[Local energy estimate]\label{lem:local}
There is an absolute constant $C$ such that for all integers $W\ge1$ and all $\eta$ with $0<\eta W\le10^{-3}$, every anchor $t_0$ as above satisfies
\begin{equation}\label{eq:local}
 \int_{I_{t_0}}\Var_{x_1^{\lfloor s\rfloor}}\bigl(L'(s)\bigr)ds\;\le\;C\frac{n^2}{W^2}\int_{J_{t_0}}\Var_{x_1^{\lfloor s\rfloor}}\bigl(L(s)\bigr)ds+C\eta^2n^2W^2\,u_{t_0} .
\end{equation}
\end{lemma}

The first term is what the frozen curve satisfies by Lemma~\ref{lem:bernstein}; the second is the price of freezing.

\begin{proof}
\emph{The frozen curve.} Write $\bar L=N/Z$ with $N(s)=\sum_{b=1}^nx_2^{t_0}(b)\,e^{-\eta(s-t_0)\ell_2^{t_0}(b)}\,A_{\cdot b}$ and $Z(s)=\sum_{b=1}^nx_2^{t_0}(b)e^{-\eta(s-t_0)\ell_2^{t_0}(b)}$, where $A_{\cdot b}$ is the $b$-th column of $A$. Both are sums of $n$ real exponentials in $s$. On $J_{t_0}$ we have $e^{-3\eta W}\le Z\le1$ and $|Z'(s)|\le\eta Z(s)$, so the quotient rule $\bar L'=N'/Z-\bar L\,Z'/Z$ and Lemma~\ref{lem:bernstein} (with $\alpha=t_0$, in the seminorm $\norm\cdot_p$ of an arbitrary fixed $p$) give
\begin{equation}\label{eq:frozen}
 \norm{\bar L'}_{L^2(I_{t_0};p)}\le e^{3\eta W}\norm{N'}_{L^2(I_{t_0};p)}+\eta\norm{\bar L}_{L^2(I_{t_0};p)}
 \le\Bigl(\frac{2ne^{3\eta W}}{W}+\eta\Bigr)\norm{\bar L}_{L^2(J_{t_0};p)}\le C\frac nW\norm{\bar L}_{L^2(J_{t_0};p)},
\end{equation}
where we used $\norm N_{L^2(J_{t_0};p)}\le\norm{\bar L}_{L^2(J_{t_0};p)}$ (as $Z\le1$) and $\eta W\le10^{-3}$.

\emph{Freezing error in position.} For $j\in[t_0,t_0+3W-1]$, since $g^j-\ell_2^{t_0}=(\ell_2^j-\ell_2^{t_0})+(\ell_2^j-\ell_2^{j-1})$, we have $\norm{g^j-\ell_2^{t_0}}_\infty\le2\delta_{t_0}$. Composing the tilts in \eqref{eq:interp} shows that for $s\in[j,j+1]$, $Y(s)=\tilt{x_2^{t_0}}{-\eta(\sum_{r=t_0}^{j-1}g^r+(s-j)g^j)}$, so the scores defining $Y(s)$ and $\bar Y(s)$ differ by at most $\eta\cdot3W\cdot2\delta_{t_0}$ in $\ell_\infty$. The Lipschitz bound of Lemma~\ref{lem:tilt} and $\norm{Av}_\infty\le\norm v_1$ give
\begin{equation}\label{eq:freezepos}
 \norm{Y(s)-\bar Y(s)}_1\le6\eta W\delta_{t_0},\qquad\norm{L(s)-\bar L(s)}_\infty\le6\eta W\delta_{t_0}\qquad(s\in J_{t_0}).
\end{equation}

\emph{Freezing error in derivative.} Let $\Sigma(p)=\operatorname{diag}(p)-pp^\top$, the derivative of the tilt with respect to its score. Then $Y'(s)=-\eta\,\Sigma(Y(s))\,g^j$ and $\bar Y'(s)=-\eta\,\Sigma(\bar Y(s))\,\ell_2^{t_0}$ for $s\in(j,j+1)$. For probability vectors $p,q$ and any $v$,
\[
 \norm{\Sigma(p)v}_1=\sum_hp(h)\bigl|v(h)-\ip pv\bigr|\le\norm v_p\le\norm v_\infty,\qquad
 \norm{(\Sigma(p)-\Sigma(q))v}_1\le3\norm{p-q}_1\norm v_\infty ,
\]
the second because $\Sigma(p)v-\Sigma(q)v=(p-q)\odot v-(p-q)\ip pv-q\ip{p-q}{v}$. Writing $\Sigma(Y)g^j-\Sigma(\bar Y)\ell_2^{t_0}=\Sigma(Y)(g^j-\ell_2^{t_0})+(\Sigma(Y)-\Sigma(\bar Y))\ell_2^{t_0}$ and using \eqref{eq:freezepos},
\begin{equation}\label{eq:freezeder}
 \norm{L'(s)-\bar L'(s)}_\infty\le\norm{Y'(s)-\bar Y'(s)}_1\le\eta\bigl(2\delta_{t_0}+18\eta W\delta_{t_0}\bigr)\le3\eta\delta_{t_0} .
\end{equation}
The absence of a factor $W$ in \eqref{eq:freezeder} is essential; it is why the derivative error is estimated directly rather than inferred from \eqref{eq:freezepos}.

\emph{Assembling the window.} Freeze the variance weights at $p=x_1^{t_0}$. By \eqref{eq:frozen}, \eqref{eq:freezepos}, \eqref{eq:freezeder} and $\norm\cdot_p\le\norm\cdot_\infty$,
\begin{align*}
 \norm{L'}_{L^2(I_{t_0};p)}&\le\norm{\bar L'}_{L^2(I_{t_0};p)}+\sqrt W\cdot3\eta\delta_{t_0}
 \le C\frac nW\bigl(\norm L_{L^2(J_{t_0};p)}+\sqrt{3W}\cdot6\eta W\delta_{t_0}\bigr)+3\eta\sqrt W\delta_{t_0}\\
 &\le C'\frac nW\norm L_{L^2(J_{t_0};p)}+C'\eta n\sqrt W\,\delta_{t_0} .
\end{align*}
By Lemma~\ref{lem:movement}, along the window the true weights $x_1^{\lfloor s\rfloor}$ and $x_1^{t_0}$ differ coordinatewise by a factor at most $e^{9\eta W}\le e^{0.009}$, so the variances with the true weights are within a constant factor of those with $p=x_1^{t_0}$, on both sides of the inequality. Squaring, using $(\alpha+\beta)^2\le2\alpha^2+2\beta^2$ and $\delta_{t_0}^2\le3Wu_{t_0}$, and enlarging the constant, we obtain \eqref{eq:local}.
\end{proof}

\paragraph{Summing over windows.}
\begin{proof}[Proof of Lemma~\ref{lem:transfer}]
Let $C$ be the constant of Lemma~\ref{lem:local}. Let $\mathcal T=\{2+kW:\ k\ge0\text{ an integer},\ 2+(k+3)W\le T\}$ be the set of anchors. The middle thirds $I_{t_0}=[t_0+W,t_0+2W]$, $t_0\in\mathcal T$, are consecutive and have disjoint interiors; the windows $J_{t_0}$ overlap at most threefold, as do the index ranges $\{t_0-1,\dots,t_0+3W-2\}$ of the sums $u_{t_0}$, which all lie within $\{1,\dots,T-1\}$. By \eqref{eq:feedback} and $(\alpha+\beta)^2\le2\alpha^2+2\beta^2$,
\[
 \sum_{t_0\in\mathcal T}u_{t_0}\le3\sum_{t=1}^{T-1}\norm{\ell_2^{t+1}-\ell_2^t}_\infty^2\le3\sum_{t=1}^{T-1}9\eta^2(v^t+v^{t-1})^2\le108\,\eta^2\sum_{t=1}^T\Var_{x_1^t}(\ell_1^t) .
\]
Summing \eqref{eq:local} over $t_0\in\mathcal T$ and using the threefold overlap of the windows,
\[
 \sum_{t_0\in\mathcal T}\int_{I_{t_0}}\Var_{x_1^{\lfloor s\rfloor}}(L'(s))\,ds\;\le\;3C\frac{n^2}{W^2}\int_1^T\Var_{x_1^{\lfloor s\rfloor}}(L(s))\,ds+108C\eta^4n^2W^2\sum_{t=1}^T\Var_{x_1^t}(\ell_1^t) .
\]
The part of $[1,T]$ not covered by the middle thirds consists of an initial segment of length at most $W+1$ and a final segment of length less than $2W$. Since $g^j\in[-1,2]^n$, the interpolation satisfies $\norm{L'(s)}_\infty\le\norm{Y'(s)}_1\le\eta\norm{g^j}_\infty\le2\eta$ almost everywhere, so the uncovered part contributes at most $4W\cdot4\eta^2=16\eta^2W$ to $\int_1^T\Var_{x_1^{\lfloor s\rfloor}}(L'(s))\,ds$; this also covers horizons too short to contain a single window. Altogether,
\[
 \int_1^T\Var_{x_1^{\lfloor s\rfloor}}(L'(s))\,ds\;\le\;C'\frac{n^2}{W^2}\int_1^T\Var_{x_1^{\lfloor s\rfloor}}(L(s))\,ds+C'\eta^4n^2W^2\sum_{t=1}^T\Var_{x_1^t}(\ell_1^t)+16\eta^2W .
\]
\begin{sloppypar}
Now insert the second comparison \eqref{eq:comp2} of Lemma~\ref{lem:comparison}, which bounds the position energy $\int_1^T\Var_{x_1^{\lfloor s\rfloor}}(L(s))\,ds$ by $2\sum_{t=1}^T\Var_{x_1^t}(\ell_1^t)+2\int_1^T\Var_{x_1^{\lfloor s\rfloor}}(L'(s))\,ds$, and choose $K_0$ so large that $W\ge K_0n$ makes $2C'n^2/W^2\le\frac12$. Absorbing the resulting $\frac12\int_1^T\Var_{x_1^{\lfloor s\rfloor}}(L'(s))\,ds$ into the left-hand side gives
\end{sloppypar}
\[
 \int_1^T\Var_{x_1^{\lfloor s\rfloor}}(L'(s))\,ds\;\le\;C''\Bigl[\Bigl(\frac{n^2}{W^2}+\eta^4n^2W^2\Bigr)\sum_{t=1}^T\Var_{x_1^t}(\ell_1^t)+\eta^2W\Bigr].
\]
Finally, by \eqref{eq:shift},
\[
 \sum_{t=1}^T\Var_{x_1^t}(\ell_1^t)\;\le\;e^{3\eta}\sum_{t=1}^T\Var_{x_1^t}(\ell_1^{t-1})+\frac14\;\le\;2\Bigl(\sum_{t=1}^T\Var_{x_1^t}(\ell_1^{t-1})+1\Bigr),
\]
and by \eqref{eq:comp1},
\[
 \sum_{t=1}^T\Var_{x_1^t}(\ell_1^t-\ell_1^{t-1})\;\le\;\frac14+e^{3\eta}\int_1^T\Var_{x_1^{\lfloor s\rfloor}}(L'(s))\,ds\;\le\;\frac14+2\int_1^T\Var_{x_1^{\lfloor s\rfloor}}(L'(s))\,ds ;
\]
combining the last three displays proves \eqref{eq:transfer}.
\end{proof}
\section{Proof of the exponential regret lower bound for three or more players}\label{sec:lowerproof}

This section constructs the three-player game of Theorem~\ref{thm:lower}. Players $1$ and $2$ play a symmetric game against each other, start identically and remain identical, and their common trajectory converges to a periodic orbit. Player $3$ is the \emph{observer}; her losses depend on player~$1$'s action only, and she does not influence players $1$ and $2$. We first analyse a single eight-action block of players $1$ and $2$ (Section~\ref{sec:lower-first}), then the linear response of a block driven by its predecessor (Section~\ref{sec:lower-second}), then the observer (Section~\ref{sec:lower-observer}); Section~\ref{sec:lower-joint} assembles all blocks into one cascade and proves its convergence, and Section~\ref{sec:lower-assembly} proves the theorem. Finally, Section~\ref{sec:lower-many} extends the construction to $m\ge3$ players, with a barrier $2^{\Omega(mn)}$ (Theorem~\ref{thm:lower-many}). Figure~\ref{fig:cascade} shows the overall architecture.

\paragraph{Conventions.}
It is convenient to describe all payoffs in this section as \emph{utilities} $U\in[-\frac12,\frac12]$, rather than as losses in $[0,1]$ as in Section~\ref{sec:setting}; the corresponding losses are $\frac12-U\in[0,1]$. Since Optimistic Hedge is invariant under adding a constant to all entries of a loss vector, only the differences between utilities matter, and we write the losses simply as $-U$. Throughout, $n$ is the number of actions of each player, $M\ge4$ is an integer divisible by four (the period), $\lambda=e^{2\pi i/M}$, and $a>0$ is a small amplitude. The letters $\mu,\delta\in\C$ denote payoff parameters that are fixed before play begins. Since the Optimistic Hedge update \eqref{eq:OH} uses the previous loss vector $\ell^{t-1}$ as well as the current one, its state consists of the current strategy \emph{and} the previous round's utilities; we call the latter the \emph{memory} and denote it by $\mem_t$. Starred quantities are exact periodic trajectories.

\subsection{A single oscillating block}\label{sec:lower-first}

\paragraph{Overview.}
We construct a symmetric two-player game whose common strategy is determined by a complex score $z_t$. Our goal is that the score will approximately rotate in a cycle of length $M$ around some center point $\delta$ with a radius of $a$:
\[
 z_t\approx\delta+a e^{2\pi it/M}.
\]

\subsubsection*{Step 1: actions, utilities and the exact score recurrence}
The two players have the same $n$ distinct actions. The eight active actions are named by the complex numbers
\[
 V=\{1,-1,i,-i,1+i,-1-i,1-i,-1+i\}.
\]
The remaining $n-8$ actions have distinct names $b_1,\ldots,b_{n-8}$; we call them \emph{neutral}. These neutral actions will later serve as the actions of the other blocks (Section~\ref{sec:lower-joint}); here they have utility zero and do not control the score of this block. Each action $\tau$ has a complex \emph{coordinate} $w(\tau)$: the complex number naming it if $\tau$ is active, and $w(b_j)=0$ if it is neutral. Below we write $v$ for the coordinate of a player's own action and $w$ for that of the opponent's action. Sums over actions still run over all $n$ actions, so each neutral action contributes separately even though they all have coordinate $0$.

\begin{figure}[t]
\centering
\begin{tikzpicture}[scale=1.15,
  pt/.style={circle,fill,inner sep=1.4pt},
  ax/.style={-{Latex[length=1.6mm]},gray}]
 \foreach \panel/\title/\xs in {0/{linear coefficient $c(w)$}/0, 1/{quadratic coefficient $q(b)$}/4.6}{
  \begin{scope}[xshift=\xs cm]
   \draw[ax] (-1.75,0) -- (1.8,0) node[right,font=\scriptsize]{$\Rea$};
   \draw[ax] (0,-1.75) -- (0,1.8) node[above,font=\scriptsize]{$\Ima$};
   \draw[gray!50,dashed] (-1,-1) rectangle (1,1);
   \node[font=\small] at (0,-2.25) {\title};
  \end{scope}}
 \foreach \x/\y/\l/\pos in {1/0/{1}/{above right}, -1/0/{-1}/{above left}, 0/1/{i}/{above right}, 0/-1/{-i}/{below right}}{
   \node[pt,blue!70!black] at (\x,\y) {};
   \node[font=\scriptsize,blue!70!black,\pos] at (\x,\y) {$\l$};}
 \foreach \x/\y in {1/1,-1/-1,1/-1,-1/1}{
   \node[pt,gray] at (\x,\y) {};
   \node[font=\scriptsize,gray,anchor=south west] at (\x,\y) {$0$};}
 \begin{scope}[xshift=4.6cm]
 \foreach \x/\y/\l/\pos in {1/0/{1}/{above right}, -1/0/{1}/{above left}, 0/1/{-1}/{above right}, 0/-1/{-1}/{below right},
                           1/1/{i/2}/{above right}, -1/-1/{i/2}/{below left}, 1/-1/{-i/2}/{below right}, -1/1/{-i/2}/{above left}}{
   \node[pt,red!70!black] at (\x,\y) {};
   \node[font=\scriptsize,red!70!black,\pos] at (\x,\y) {$\l$};}
 \end{scope}
\end{tikzpicture}
\caption{The eight active actions $V=\{\pm1,\pm i,\pm1\pm i\}$ as points of $\C$, labeled by their payoff coefficients; neutral actions sit at $0$ and have coefficient $0$. Left: $c(w)=w$ on the axis actions, so the expectation of $c$ is linear in the score to first order and produces the rotation of Section~\ref{sec:lower-first}. Right: $q(b)=b^2$ on the axis actions and $b^2/4$ on the diagonals, so the expectation of $q$ is the square of the score's displacement to leading order (Lemma~\ref{lem:forced-signal}).}
\label{fig:actions}
\end{figure}
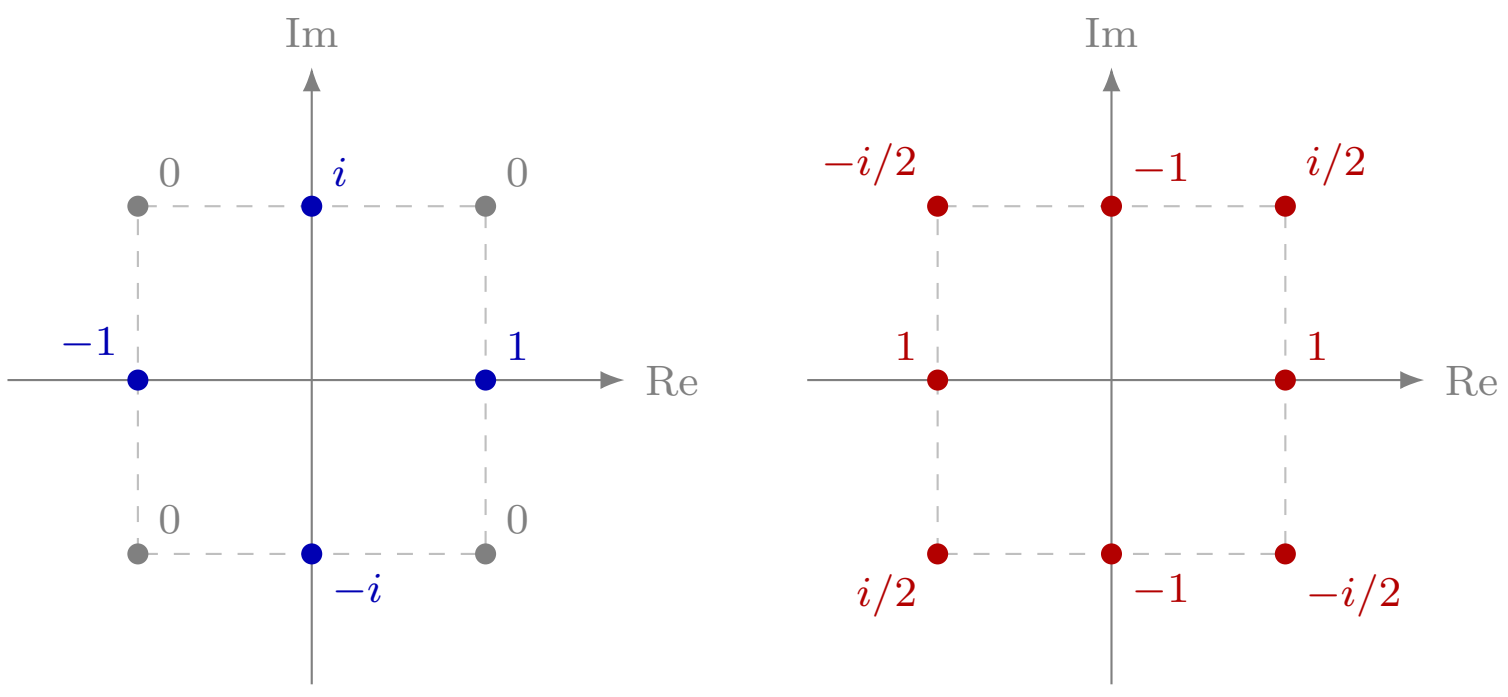

The game will have the following property. There is a complex score $z_t$, with $z_0=0$, such that both players assign to each action $\tau$ the real probability
\begin{equation}\label{eq:pz}
 p_z(\tau)=\frac{\exp(\Rea(\overline{w(\tau)}z))}{Z(z)},\qquad
 Z(z)=\sum_{\tau}\exp(\Rea(\overline{w(\tau)}z)).
\end{equation}
If $z=z_1+iz_2$ and $w=w_1+iw_2$, then $\Rea(\overline wz)=w_1z_1+w_2z_2\in\R$. Hence these are positive real probabilities summing to one; $Z(0)=n$ and $p_0$ is uniform. Figure~\ref{fig:actions} shows the eight active actions.

Define $c(1)=1$, $c(-1)=-1$, $c(i)=i$, $c(-i)=-i$, and $c=0$ on diagonal and neutral actions. Let $\mu\in\C$ be a payoff coefficient and $\delta\in\C$ a shift parameter, to be chosen later and then held fixed throughout play. Both players use the utility table
\begin{equation}\label{eq:gain1}
 U(v,w)=\Rea\!\left(\overline v\,\mu c(w)
                  e^{-\Rea(\overline w\delta)}\right).
\end{equation}
The losses are $-U$. Define the expected complex utility coefficient
\begin{equation}\label{eq:g}
 \begin{split}
 g_{\mu,\delta}(z)
 &=\sum_w p_z(w)\mu c(w)e^{-\Rea(\overline w\delta)}\\
 &=\frac{2\mu}{Z(z)}\left[\sinh(\Rea(z-\delta))
                         +i\sinh(\Ima(z-\delta))\right].
 \end{split}
\end{equation}
Here and below $p_z(w)$ in a sum means the probability of the corresponding action label. In particular $g_{\mu,\delta}(\delta)=0$. The memory is $\mem_t=g_{\mu,\delta}(z_{t-1})$, the utility coefficient of the previous round, with $\mem_0=0$; it enters the update through the term $-\ell^{t-1}$ in \eqref{eq:OH}. The score and memory then follow the recurrence
\begin{equation}\label{eq:oh}
 \begin{split}
 z_{t+1}&=z_t+\eta[2g_{\mu,\delta}(z_t)-\mem_t],\\
 \mem_{t+1}&=g_{\mu,\delta}(z_t),\qquad (z_0,\mem_0)=(0,0).
 \end{split}
\end{equation}
Denote its one-step map on $\C^2\cong\R^4$ by $\Phi_{\mu,\delta}$.

\begin{lemma}[The game realizes the recurrence]\label{lem:block-realization}
For utilities \eqref{eq:gain1}, two identical players using Optimistic Hedge from the uniform distribution have the same strategy $p_{z_t}$ at every round, where $(z_t,\mem_t)$ follows \eqref{eq:oh}. The expected utility of action $v$ against $p_{z_t}$ is $\Rea(\overline v g_{\mu,\delta}(z_t))$.
\end{lemma}
\begin{proof}
The formula for expected utility follows by averaging \eqref{eq:gain1}; the four nonzero coefficients $c(w)$ give \eqref{eq:g}. If $\gamma_t=g_{\mu,\delta}(z_t)$, the utility form of Optimistic Hedge gives
\[
 p_{t+1}(v)\propto p_{z_t}(v)
  e^{\eta\Rea(\overline v(2\gamma_t-\mem_t))}
 \propto e^{\Rea(\overline v[z_t+\eta(2\gamma_t-\mem_t)])}.
\]
Action-independent terms cancel on normalization. Thus the next strategy again has form \eqref{eq:pz}, with precisely the updated score in \eqref{eq:oh}. Initially the strategy is $p_0$ and the memory is zero, since the initial prediction of Optimistic Hedge is the same for all actions. Symmetry and identical initialization keep the two players identical by induction.
\end{proof}

\begin{theorem}[An oscillating block]\label{thm:first}
Let $n\ge9$, let $M\ge4$ be divisible by four, and suppose $0<\eta\le1/100$ and $\eta M\ge16\pi n$. For every sufficiently small $a>0$, there are complex parameters $\mu,\delta$, with $\delta=O(a)$, for which \eqref{eq:gain1} lies in $[-\frac12,\frac12]$ and \eqref{eq:oh} has an orbit $s_t^\ast=(z_t^\ast,\mem_t^\ast)$ of exact period $M$ satisfying
\[
 z_t^\ast=\delta+a\lambda^t+O(a^2),\qquad \lambda=e^{2\pi i/M}.
\]
The remainder is $M$-periodic. The trajectory from $(0,0)$ satisfies
\[
 |z_t-z_t^\ast|+|\mem_t-\mem_t^\ast|\le Ca\rho^t
 \quad(t\ge0)
\]
for some $C<\infty$ and $\rho<1$, independent of $t$ and of sufficiently small $a$.
\end{theorem}

\begin{figure}[tbp]
\centering
\pgfplotsset{oscax/.style={width=0.27\textwidth,height=4.6cm,
   xmin=0,xmax=6.2832,ymin=-1.75,ymax=1.75,axis lines=left,font=\scriptsize,
   xtick={0,1.5708,3.1416,4.7124,6.2832},xticklabels={$0$,$\frac M4$,$\frac M2$,$\frac{3M}4$,$M$},
   ytick={-1.4142,-1,0,1,1.4142},yticklabels={$-\sqrt2$,$-1$,$0$,$1$,$\sqrt2$},
   xlabel={round $t$},ylabel={$[a]$},clip=false,
   legend style={at={(0.5,-0.36)},anchor=north,font=\tiny,legend columns=2,draw=none,column sep=4pt}}}
\begin{tikzpicture}
 \begin{scope}
  \coordinate (d) at (0.15,0.08);
  \draw[-{Latex[length=1.3mm]},gray] (-1.5,0) -- (1.55,0) node[below left=-1pt,font=\tiny]{$\Rea z$};
  \draw[-{Latex[length=1.3mm]},gray] (0,-1.3) -- (0,1.4) node[left,font=\tiny]{$\Ima z$};
  \draw[thick,red!70!black] (d) circle (1.0);
  \draw[-{Latex[length=1.6mm]},thick,red!70!black] (d) ++(35:1.0) arc[start angle=35,end angle=70,radius=1.0];
  \fill (d) circle (0.9pt) node[below right=-1pt,font=\tiny]{$\delta$};
  \draw[red!70!black] (d) -- node[above right=-2pt,font=\tiny,pos=0.6]{$a$} ++(-40:1.0);
  \fill[red!70!black] (d) ++(0:1.0) circle (1.3pt);
  \node[font=\tiny,red!70!black,above right] at ($(d)+(0:1.05)$) {$t=0$};
  \fill[red!70!black] (d) ++(90:1.0) circle (1.3pt);
  \node[font=\tiny,red!70!black,above right] at ($(d)+(90:1.05)$) {$t=\frac M4$};
  \fill[red!70!black] (d) ++(180:1.0) circle (1.3pt);
  \node[font=\tiny,red!70!black,left] at ($(d)+(180:1.05)$) {$t=\frac M2$};
  \fill[red!70!black] (d) ++(270:1.0) circle (1.3pt);
  \node[font=\tiny,red!70!black,below right] at ($(d)+(270:1.05)$) {$t=\frac{3M}4$};
  \fill (0,0) circle (0.9pt);
  \node[font=\tiny,above left=-1pt] at (0,0) {$0$};
  \node[font=\scriptsize] at (0.1,2.0) {score $z_t\approx\delta+a\lambda^t$};
 \end{scope}
 \begin{axis}[oscax,name=pa,anchor=west,at={(3.3cm,0)},title={\scriptsize axis actions: $\log p_t(w)+\log n$}]
  \draw[gray] (axis cs:0,0) -- (axis cs:6.2832,0);
  \addplot[domain=0:6.2832,samples=80,thick,red!75!black] {1.0000*cos(deg(x)-0.0)};
  \addlegendentry{$w=1$}
  \addplot[domain=0:6.2832,samples=80,thick,olive!80!black] {1.0000*cos(deg(x)-90.0)};
  \addlegendentry{$w=i$}
  \addplot[domain=0:6.2832,samples=80,thick,teal!80!black] {1.0000*cos(deg(x)-180.0)};
  \addlegendentry{$w=-1$}
  \addplot[domain=0:6.2832,samples=80,thick,violet!80!black] {1.0000*cos(deg(x)-270.0)};
  \addlegendentry{$w=-i$}
 \end{axis}
 \begin{axis}[oscax,name=pd,anchor=west,at={($(pa.east)+(1.7cm,0)$)},title={\scriptsize diagonal actions: $\log p_t(w)+\log n$}]
  \draw[gray] (axis cs:0,0) -- (axis cs:6.2832,0);
  \addplot[domain=0:6.2832,samples=80,thick,densely dashed,orange!85!black] {1.4142*cos(deg(x)-45.0)};
  \addlegendentry{$w=1+i$}
  \addplot[domain=0:6.2832,samples=80,thick,densely dashed,green!50!black] {1.4142*cos(deg(x)-135.0)};
  \addlegendentry{$w=-1+i$}
  \addplot[domain=0:6.2832,samples=80,thick,densely dashed,blue!75!black] {1.4142*cos(deg(x)-225.0)};
  \addlegendentry{$w=-1-i$}
  \addplot[domain=0:6.2832,samples=80,thick,densely dashed,magenta!80!black] {1.4142*cos(deg(x)-315.0)};
  \addlegendentry{$w=1-i$}
 \end{axis}
\end{tikzpicture}
\caption{The oscillating block of Theorem~\ref{thm:first} over one period of $M$ rounds, to leading order in the amplitude $a$. Left: the common score $z_t$ of players $1$ and $2$ rotates once around the center $\delta$, with radius $a$ and $\lambda=e^{2\pi i/M}$; the uniform start is the point $0$. Middle and right: the log-probabilities of the four axis actions $w\in\{\pm1,\pm i\}$ and of the four diagonal actions $w\in\{\pm1\pm i\}$, in units of $a$. They are linear in the score, $\log p_t(w)=\Rea(\overline wz_t)-\log Z(z_t)$, with $\log Z(z_t)=\log n+O(a^2)$ common to all actions, so each is a cosine $a|w|\cos(2\pi t/M-\arg w)$, up to the constant $\Rea(\overline w\delta)$, which is omitted. Each action is most likely when $z_t-\delta$ points in its direction, so within each group the curves peak a quarter period apart; the diagonal actions have a $\sqrt2$ times larger amplitude.}
\label{fig:ov-oscillator}
\end{figure}
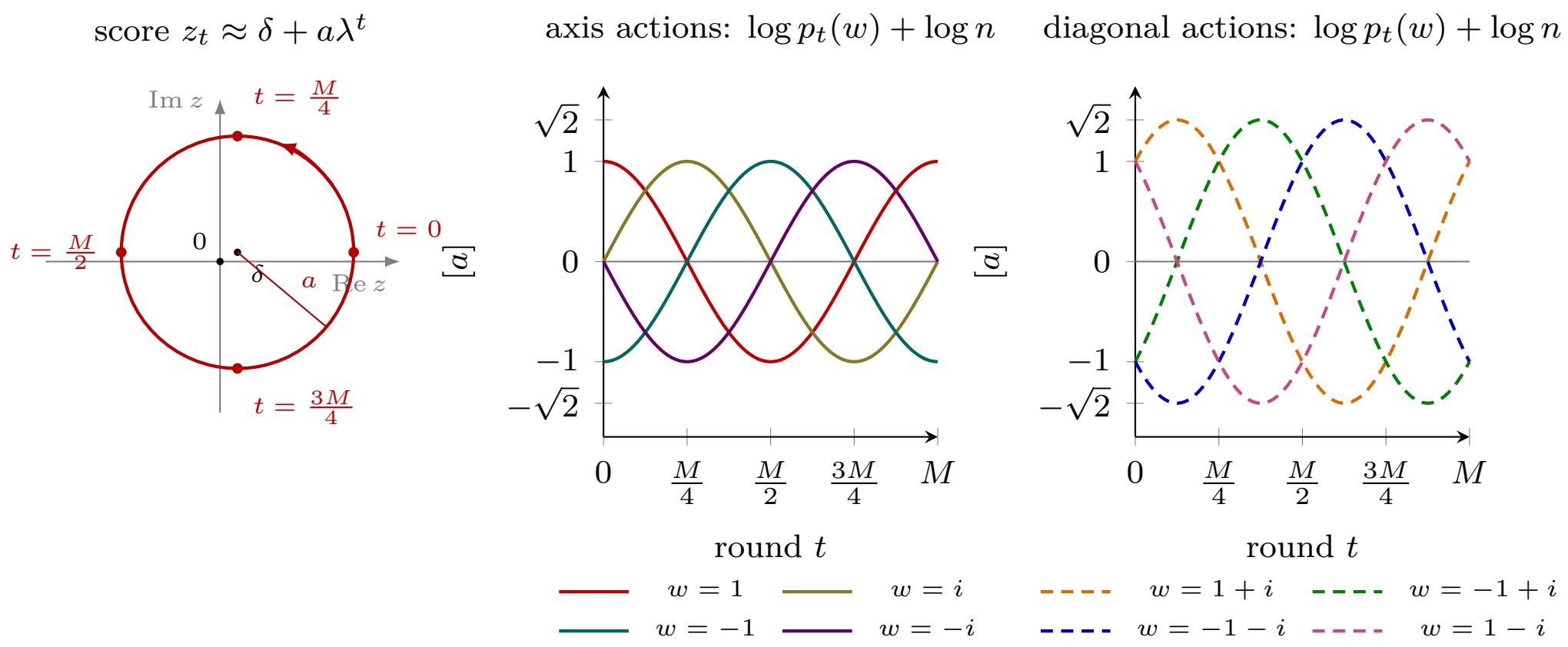

Steps 2 and 3 construct an approximate rotation and close it into an exact cycle by adjusting the utility coefficient $\mu$. Steps 4 and 5 find the manifold of states attracted to this cycle and adjust $\delta$ so that the uniform start lies on it. Step 6 checks the utilities. The assumption $\eta M\ge16\pi n$ says that the rotation is slow: rotating by $2\pi/M$ per round takes a utility coefficient $|\mu|\approx\pi n/(\eta M)$, and bounded utilities cap it (Step~6). Figure~\ref{fig:ov-oscillator} shows the resulting cycle.

For the rest of the proof, $n,\eta,M$ are fixed. Constants and smallness thresholds may depend on them, but are uniform in time and in sufficiently small amplitude and shift. We use the Euclidean norm on $\C^2\cong\R^4$ unless another norm is specified. For a periodic sequence, $O(a^2)$ means a bound uniform over all $M$ phases, hence over all time.

\subsubsection*{Step 2: a circular approximation from the linearization}
We first choose $\mu$ so that the linearized score rotates once in $M$ rounds. Define
\begin{equation}\label{eq:beta}
 \beta=\frac{\lambda(\lambda-1)}{2\lambda-1},\quad
 \chi=\frac{\lambda-1}{2\lambda-1},\quad
 \mu_0=\frac{n\beta}{2\eta},\quad
 \mu_\ast(\delta)=\frac{Z(\delta)\beta}{2\eta}.
\end{equation}
Here $\beta$ selects the rotating mode, $\chi$ will be the second, contracting eigenvalue of the linearized update, and $\mu_\ast(\delta)$ is the reference utility coefficient at center $\delta$, with $\mu_\ast(0)=\mu_0$.

\begin{lemma}[Linear rotation]\label{lem:block-approx}
For small $\delta$ and $\mu=\mu_\ast(\delta)$, the derivative of $\Phi_{\mu,\delta}$ at its fixed point $(\delta,0)$ is
\[
 A=\begin{pmatrix}1+2\beta&-\eta\\ \beta/\eta&0\end{pmatrix},
\]
with eigenvalues $\lambda$ and $\chi$, where $|\chi|<1$. In particular the trial states $\widetilde s_t=(\widetilde z_t,\widetilde\mem_t)$, with $\widetilde z_t=\delta+a\lambda^t$ and $\widetilde\mem_t=\frac{a\beta}{\eta}\lambda^{t-1}$, satisfy $\Phi_{\mu,\delta}(\widetilde s_t)-\widetilde s_{t+1}=O(a^2)$.
\end{lemma}
\begin{proof}
Taylor's theorem at $z=\delta$ gives $g_{\mu,\delta}(\delta+u)=\frac{2\mu}{Z(\delta)}u+O(|u|^2)$, and $2\eta\mu_\ast(\delta)/Z(\delta)=\beta$; this gives $A$. Its characteristic polynomial is $(r-\lambda)(r-\chi)$, and its $\lambda$-eigenvector is $(1,\beta/(\eta\lambda))$, so $\widetilde s_t-(\delta,0)$ follows the linearized update exactly and the Taylor remainder gives the residual. Finally $|\chi|<1$, because $|2\lambda-1|^2-|\lambda-1|^2=3-2\Rea\lambda>0$.
\end{proof}
The residual is $O(a^2)$ but not zero, and along the neutral rotation such errors could accumulate. Step 3 removes them.

\subsubsection*{Step 3: an exact cycle by the implicit function theorem}
We let both the orbit's shape and $\mu$ vary, keeping the amplitude $a$ and the center $\delta$ as prescribed parameters.
\begin{lemma}[Exact periodic family]\label{lem:block-periodic}
For every fixed $n,\eta,M$ as in Theorem~\ref{thm:first}, there is a neighborhood of $(a,\delta)=(0,0)$ on which real-analytic functions $h_t(a,\delta)$ and $\mu(a,\delta)$ exist, with $h_{t+M}=h_t$, such that
\[
 z_t^\ast=\delta+ah_t,\qquad
 \mem_t^\ast=g_{\mu,\delta}(z_{t-1}^\ast)
\]
satisfy the recurrence exactly. They obey $h_t(0,\delta)=\lambda^t$, $\mu(0,\delta)=\mu_\ast(\delta)$ and $h_t(a,\delta)=\lambda^t+O(a)$, uniformly for small $\delta$. For $a\ne0$ the orbit has exact period $M$.
\end{lemma}
We use the following theorem, reading complex variables as pairs of real ones.

\begin{theorem}[Implicit function theorem]\label{thm:block-ift}
Let $F(x,p)$ be $C^r$, $r\ge1$, near $(x_0,p_0)$, with equal numbers of real equations and real unknowns $x$. If $F(x_0,p_0)=0$ and $D_xF(x_0,p_0)$ is invertible, then for all $p$ near $p_0$ there is a unique nearby solution $x=x(p)$, which is $C^r$. If $F$ is real analytic, so is $x(p)$.
\end{theorem}
The unknowns are $x=(h_0,\ldots,h_{M-1},\mu)$ and the parameters are $p=(a,\delta)$. Define the residuals
\begin{equation}\label{eq:periodic}
 \begin{split}
 \mathcal R_t&=h_{t+1}-h_t-\frac\eta a
 [2g_{\mu,\delta}(\delta+ah_t)-g_{\mu,\delta}(\delta+ah_{t-1})],\\
 N&=\frac1M\sum_{t=0}^{M-1}h_t\lambda^{-t}-1,
 \qquad t\bmod M.
 \end{split}
\end{equation}
These are $M+1$ complex equations in $M+1$ complex unknowns. The condition $N=0$ sets the coefficient of $\lambda^t$ in $h$ to $1$; this removes the freedom to rescale or rotate $h$, and makes $a$ the radius.

\begin{lemma}[Properties of the periodic residuals]\label{lem:block-periodic-properties}
The map $(\mathcal R_0,\ldots,\mathcal R_{M-1},N)$ extends real analytically across $a=0$, vanishes at $a=\delta=0$, $h_t=\lambda^t$, $\mu=\mu_0$, and has an invertible real Jacobian with respect to $(h,\mu)$ there. At $a=0$ it also vanishes for every small $\delta$ when $h_t=\lambda^t$ and $\mu=\mu_\ast(\delta)$.
\end{lemma}
\begin{proof}
\emph{Smoothness and the known zero.} Since $g_{\mu,\delta}(\delta)=0$, the quotient has the analytic extension
\[
 \frac{g_{\mu,\delta}(\delta+ah)}a
 =\int_0^1Dg_{\mu,\delta}(\delta+sah)[h]\,ds.
\]
The base-point identities follow from $\lambda-1=\beta(2-\lambda^{-1})$ and $2\eta\mu_\ast(\delta)/Z(\delta)=\beta$.

\emph{Invertibility.} At the base point the derivative in $h$ of the first $M$ residuals is
\[
 (\mathcal L\dot h)_t=\dot h_{t+1}-\dot h_t-\beta(2\dot h_t-\dot h_{t-1}).
\]
On the Fourier mode $\zeta^t$, $\zeta^M=1$, its multiplier is
\[
 L(\zeta)=\zeta-1-\beta(2-\zeta^{-1})
          =\frac{(\zeta-\lambda)(\zeta-\chi)}\zeta.
\]
The $M$ sequences $\zeta^t$, where $\zeta$ ranges over the $M$th roots of unity, form a basis of the complex $M$-periodic sequences. Since $|\chi|<1$, only $\zeta=\lambda$ has zero multiplier. Changing $h$ can therefore correct every Fourier component of the residual except this one. The derivative in $\mu$ of these residuals is
\[
 -\frac{2\eta}{n}(2-\lambda^{-1})\lambda^t\,d\mu,
\]
a nonzero multiple of the missing mode, since $|2-\lambda^{-1}|\ge1$. To solve the linearized system for arbitrary target residuals, first use their $\lambda$-coefficient to determine $d\mu$. Next divide each other Fourier coefficient by its nonzero multiplier $L(\zeta)$ to determine the nonresonant part of $\dot h$. Finally the derivative of $N$, namely $M^{-1}\sum_t\dot h_t\lambda^{-t}$, specifies the remaining coefficient of $\dot h$. Every target thus has exactly one preimage. The Jacobian is invertible over $\C$ at this base point, and hence over $\R$.
\end{proof}
\begin{proof}[Proof of Lemma~\ref{lem:block-periodic}]
Apply Theorem~\ref{thm:block-ift} using Lemma~\ref{lem:block-periodic-properties}. Local uniqueness and the explicit solutions at $a=0$ identify $h_t(0,\delta)$ and $\mu(0,\delta)$. Smoothness gives $h_t=\lambda^t+O(a)$. The equations $\mathcal R_t=0$ are exactly the score recurrence with memory $g(z_{t-1}^\ast)$, so both score and memory repeat after $M$ steps, with no accumulated update error. The normalized first Fourier coefficient is $1$; any sequence of a proper period dividing $M$ has zero coefficient at the primitive mode $\lambda^t$. Thus $h$, and hence $s^\ast$ for $a\ne0$, has exact period $M$.
\end{proof}

\subsubsection*{Step 4: a differentiable strong stable manifold}
\begin{figure}[t]
\centering
\begin{tikzpicture}[scale=1.0,
  pt/.style={circle,fill,inner sep=1.5pt},
  arr/.style={-{Latex[length=2mm]},thick}]
 \begin{scope}
  \draw[gray!70,thick] (0,0) ellipse (1.6 and 0.6);
  \node[pt] (s0) at (1.6,0) {};
  \node[font=\scriptsize,below right] at (s0) {$s_0^\ast$};
  \draw[blue!70!black,thick] (1.2,-1.6) .. controls (1.5,-0.6) and (1.7,0.6) .. (2.25,1.6)
     node[above,font=\scriptsize]{$W^{ss}$};
  \node[pt,red!70!black] (u) at (2.8,-0.9) {};
  \node[font=\scriptsize,red!70!black,right] at (u) {$0$ (uniform start)};
  \draw[arr,gray] (0.7,0.75) arc[start angle=60,end angle=20,x radius=1.6,y radius=0.6];
  \node[font=\scriptsize,gray] at (-0.3,0.95) {cycle};
  \node[font=\small] at (0.9,-2.2) {shift $\delta$ not tuned: $0\notin W^{ss}$};
 \end{scope}
 \begin{scope}[xshift=7.2cm]
  \draw[gray!40,dashed] (0,0) ellipse (1.6 and 0.6);
  \draw[gray!70,thick] (0.45,-0.3) ellipse (1.6 and 0.6);
  \draw[arr,gray!70] (0,0) -- (0.42,-0.28) node[midway,above right=-2pt,font=\scriptsize]{$\delta$};
  \node[pt] (s0) at (2.05,-0.3) {};
  \node[font=\scriptsize,left] at (1.98,0.05) {$s_0^\ast$};
  \draw[blue!70!black,thick] (1.25,-2.0) .. controls (1.75,-1.0) and (2.15,0.3) .. (2.7,1.3)
     node[above,font=\scriptsize]{$W^{ss}$};
  \node[pt,red!70!black] (u) at (1.45,-1.62) {};
  \node[font=\scriptsize,red!70!black,right] at (u) {$0$ (uniform start)};
  \draw[red!60!black,thick,-{Latex[length=1.6mm]}]
     (1.38,-1.2) .. controls (1.55,-0.85) and (1.75,-0.55) .. (1.9,-0.38);
  \node[font=\small] at (0.9,-2.2) {$\delta=\delta(a)$: $0\in W^{ss}$};
 \end{scope}
\end{tikzpicture}
\caption{Schematic of Steps 4 and 5 in the four-dimensional state space of $(z,\mem)$. The cycle of Step 3 is neutrally stable along the center directions $E_c$, so a generic start does not converge to a prescribed phase. The states that do converge to the phase $s_0^\ast$ form the strong stable manifold $W^{ss}$, a two-dimensional graph $\xi\mapsto s_0^\ast+\xi+\Gamma(a,\delta,\xi)$ over $E_s$. Changing $\delta$ translates the cycle, and with it $W^{ss}$, in a center direction (Lemma~\ref{lem:block-matching-properties}); its two real coordinates are tuned so that $W^{ss}$ passes through the uniform start.}
\label{fig:manifold}
\end{figure}
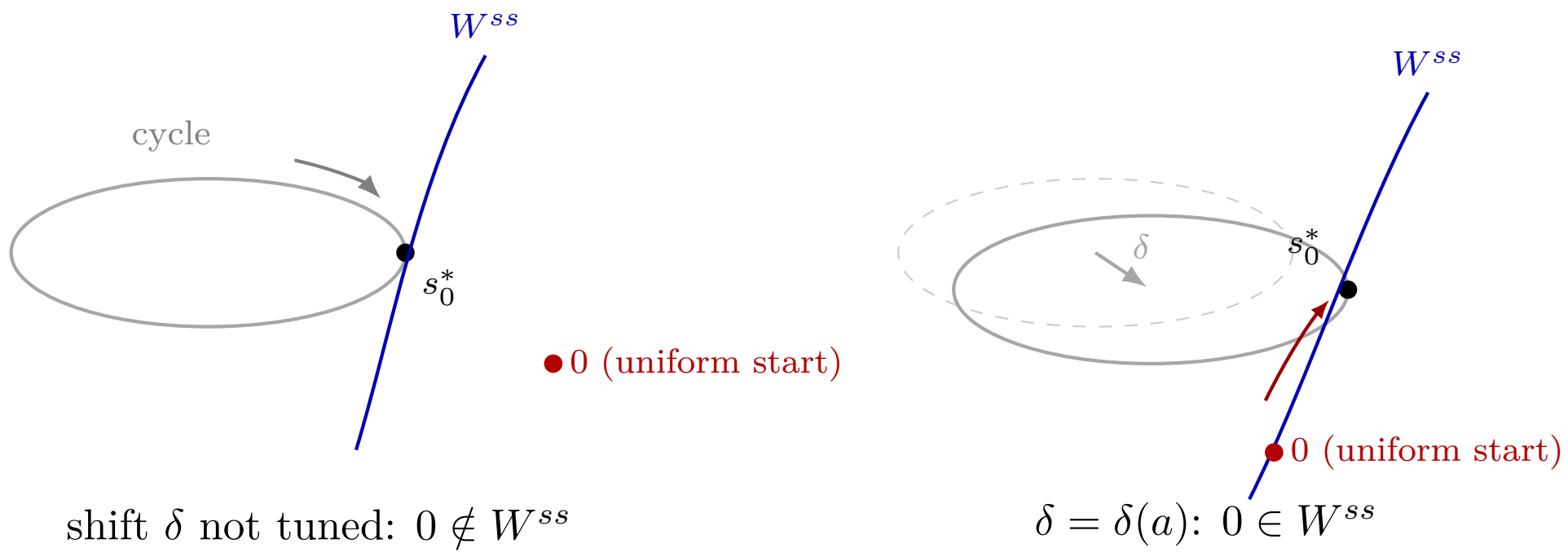

A cycle need not attract the uniform start. We identify the states it does attract; Figure~\ref{fig:manifold} sketches this step and the next.
\begin{definition}[Strong stable manifold of a cycle]
For a specified phase $s_0^\ast$ of an $M$-cycle, a local strong stable manifold is a smooth manifold through $s_0^\ast$ whose points stay near the cycle and satisfy
\[
 \|\Phi^t(s)-s_t^\ast\|\le C\rho^t\|s-s_0^\ast\|\quad(t\ge0)
\]
for a prescribed exponential rate $\rho<1$.
\end{definition}
Let $E_c,E_s\subset\C^2$ be the fixed eigenspaces of the base matrix $A$ for $\lambda,\chi$, with spectral projections $P_c,P_s$ along the other space. Explicitly,
\[
 E_c=\operatorname{span}_{\C}(1,\chi/\eta),\qquad
 E_s=\operatorname{span}_{\C}(1,\lambda/\eta).
\]
Here $c$ refers to the neutral rotating directions and $s$ to the strictly stable ones; each has real dimension two. The splitting is fixed and does not vary with $a,\delta$, so that the manifolds for all parameters are graphs over the same space.

\begin{lemma}[Stable manifolds for the periodic family]\label{lem:block-stable}
For all sufficiently small $a,\delta$, the cycle of Lemma~\ref{lem:block-periodic} has a local two-real-dimensional strong stable manifold of the form
\[
 W^{ss}_{a,\delta}
 =\{s_0^\ast+\xi+\Gamma(a,\delta,\xi):\xi\in E_s\text{ small}\},
\]
where $\Gamma$ takes values in $E_c$ and is jointly $C^1$. It satisfies $\Gamma(a,\delta,0)=0$ for all parameters and $D_\xi\Gamma(0,0,0)=0$. The convergence rate and constant can be chosen uniformly on a smaller parameter neighborhood.
\end{lemma}
We obtain this graph from the parameter-dependent strong stable manifold theorem (see, e.g., \cite{Shub87}).

\begin{theorem}[Parameter-dependent strong stable manifold theorem]\label{thm:block-stable}
Let $G_p$ be a jointly $C^r$ family of local diffeomorphisms, $r\ge2$, with a $C^r$ family of fixed points $\bar s_p$. Suppose, at a base parameter, the derivative at $\bar s_p$ has an invariant splitting into a strongly stable space and a complementary space, with all eigenvalue moduli in the first strictly below $\bar\rho<1$ and all those in the second strictly above $\bar\rho$. On a sufficiently small parameter neighborhood there are local strong stable manifolds through $\bar s_p$, of the dimension of the strongly stable space, depending jointly $C^1$ on the parameter and local manifold coordinate. They are tangent to the corresponding strong stable eigenspaces. After shrinking neighborhoods, their forward iterates converge to $\bar s_p$ at a uniform rate $C\bar\rho^{\,k}$. They can be expressed as graphs over the fixed base strongly stable space, along its fixed complement.
\end{theorem}

\begin{lemma}[Properties needed for the stable manifold theorem]\label{lem:block-stable-properties}
For $p=(a,\delta)$ put $G_p=\Phi_{\mu(a,\delta),\delta}^M$ and $\bar s_p=s_0^\ast(a,\delta)$. These are smooth real maps with $G_p(\bar s_p)=\bar s_p$. At $p=0$ their derivative is $A^M$, with complex eigenvalues $1$ and $\chi^M$. They are local diffeomorphisms near the cycle. For any fixed $\bar\rho$ with $|\chi|^M<\bar\rho<1$, the two real strongly stable directions remain spectrally separated from the two remaining directions for small parameters.
\end{lemma}
\begin{proof}
We apply the theorem to the return map after one period. Analyticity and the fixed-point identity follow from Step 3 and the analytic formula for $\Phi$. At zero amplitude $\Phi$ fixes the origin and has derivative $A$, so the chain rule gives $DG_0(0)=A^M$. Its spectrum has the stated gap because $\lambda^M=1$ and $|\chi|<1$. Also $\det_{\C}A=\beta\ne0$, so the corresponding real derivative is invertible. Continuity of derivatives, eigenvalue multisets and the separated spectral subspaces preserves local invertibility and the strict gap for all small parameters.
\end{proof}
\begin{proof}[Proof of Lemma~\ref{lem:block-stable}]
Apply Theorem~\ref{thm:block-stable} using Lemma~\ref{lem:block-stable-properties}. Projection onto $E_s$ is an invertible coordinate on each nearby manifold, giving the displayed graph in fixed coordinates. Every manifold contains its reference point, so $\Gamma(a,\delta,0)=0$. At the base parameter it is tangent to $E_s$, so $D_\xi\Gamma(0,0,0)=0$. The return-map estimate is $C\bar\rho^{\,k}$ after $k$ periods. The finitely many intermediate maps $\Phi^j$, $0\le j<M$, have uniformly bounded local derivatives; hence the estimate holds at every round with $\rho=\bar\rho^{1/M}<1$ and an adjusted constant.
\end{proof}

\subsubsection*{Step 5: tune the manifold to contain the uniform start}
We tune the two real coordinates of $\delta$ so that the uniform start $(0,0)$ lies on the stable manifold.
\begin{lemma}[Matching the prescribed initial state]\label{lem:block-matching}
For every sufficiently small real $a$ there is a $C^1$ choice $\delta=\delta(a)=O(a)$ for which $0\in W^{ss}_{a,\delta(a)}$.
\end{lemma}
Membership of the uniform state $s_0=0$ is exactly
\begin{equation}\label{eq:fiber}
 P_c(-s_0^\ast)=\Gamma(a,\delta,P_s(-s_0^\ast)).
\end{equation}
These are two real equations, which we solve for $\delta\in\C\cong\R^2$ by the implicit function theorem, with $a$ as the parameter.

\begin{lemma}[Properties of the matching equation]\label{lem:block-matching-properties}
The difference between the two sides of \eqref{eq:fiber}, denoted
\[
 \Delta(a,\delta)=P_c(-s_0^\ast)-\Gamma(a,\delta,P_s(-s_0^\ast))\in E_c,
\]
is $C^1$, satisfies $\Delta(0,0)=0$, and has an invertible real derivative in $\delta$ at $(0,0)$.
\end{lemma}
\begin{proof}
The key point is transversality: changing $\delta$ moves the cycle in a center direction to first order, whereas the graph correction has zero first derivative at the base point. Steps 3 and 4 give smoothness and the base-point identity. At $a=0$, $s_0^\ast(0,\delta)=(\delta,0)$. We claim that
\[
 D_\delta\Delta(0,0)[d\delta]=-P_c(d\delta,0).
\]
To verify the claim, apply the chain rule to the graph term in $\Delta$. Its direct parameter derivative vanishes because $\Gamma(a,\delta,0)$ is identically zero. The derivative through the argument $P_s(-s_0^\ast)$ also vanishes, because it is multiplied by $D_\xi\Gamma(0,0,0)=0$. Only the derivative of the center displacement remains. To see invertibility explicitly, decompose any $(z,\mem)$ as
\[
 (z,\mem)=u_c(1,\chi/\eta)+u_s(1,\lambda/\eta),\qquad
 u_c=\frac{\lambda z-\eta\mem}{\lambda-\chi}.
\]
In the scalar coordinate $u_c$ on $E_c$, the displayed derivative is multiplication by $-\lambda/(\lambda-\chi)\ne0$. Thus it is an invertible map of two-dimensional real spaces.
\end{proof}
\begin{proof}[Proof of Lemma~\ref{lem:block-matching}]
Apply Theorem~\ref{thm:block-ift} to $\Delta(a,\delta)=0$ using Lemma~\ref{lem:block-matching-properties}. It gives $\delta(0)=0$ and a $C^1$ solution for all sufficiently small signed $a$, in particular for small $a>0$. Differentiability gives $\delta(a)=O(a)$. Equation~\eqref{eq:fiber} then puts the uniform initial state on the strong stable manifold. Its displacement from the cycle is $O(a)$, and hence lies in the required local neighborhood for small $a$.
\end{proof}

\subsubsection*{Step 6: conclude exponential convergence to an exact cycle}
\begin{proof}[Proof of Theorem~\ref{thm:first}]
Choose $a>0$ small, put $\delta=\delta(a)$ from Step 5, and set $\mu=\mu(a,\delta(a))$ from Step 3. Step 1 identifies Optimistic Hedge with the full score-and-memory recurrence from $0$. Step 3 supplies an exact period-$M$ orbit with $z_t^\ast=\delta+a\lambda^t+O(a^2)$. Steps 4 and 5 imply that the uniform start converges to the specified phase of this orbit, with error at most $Ca\rho^t$ because $\|s_0^\ast\|=O(a)$.

It remains to check bounded utilities; this is where the assumption on $\eta M$ is used. Since $|\beta|\le|\lambda-1|\le2\pi/M$, the assumption on $\eta M$ gives
\[
 |\mu_0|\le\frac{\pi n}{\eta M}\le\frac1{16}.
\]
For $v\in V$, $|v|\le\sqrt2$; for every opponent action, $|c(w)|\le1$ and $e^{-\Rea(\overline w\delta)}\le e^{\sqrt2|\delta|}$. Thus
\[
 |U(v,w)|\le\sqrt2|\mu|e^{\sqrt2|\delta|}<\tfrac12
\]
for sufficiently small $a$, by continuity from $(\mu_0,0)$. Neutral actions have utility $0$. All utilities therefore lie in $[-\frac12,\frac12]$, completing the proof.
\end{proof}

\subsection{A block driven by a periodic predecessor}\label{sec:lower-second}

We now build a block that responds to its predecessor at twice the frequency. The \emph{source} is the preceding block, with score $x_t$ on the orbit $x_t^\ast=\delta+a\lambda^t+O(a^2)$ of Theorem~\ref{thm:first}; the new block has score $y_t$. Block number and round number are different indices: the source's strategy at round $t$ determines an input used to update the new block for round $t+1$. In this subsection the new block is carried by a separate pair of identically initialized players, with its own normalizer, and we work out its linear response. This is also exactly how the first block of each further pair of players is driven in the $m$-player game of Section~\ref{sec:lower-many}. Section~\ref{sec:lower-joint} puts all blocks into the same two players and handles the nonlinear terms and the convergence from the uniform start for all blocks at once.

\subsubsection*{Step 1: extract a signal rotating at twice the frequency}
\begin{figure}[tbp]
\centering
\pgfplotsset{dblax/.style={width=0.4\textwidth,height=4.6cm,
   xmin=0,xmax=6.2832,ymin=-1.3,ymax=1.3,ytick={-1,0,1},axis lines=left,font=\scriptsize,
   xtick={0,1.5708,3.1416,4.7124,6.2832},xticklabels={$0$,$\frac M4$,$\frac M2$,$\frac{3M}4$,$M$},
   xlabel={round $t$},clip=false,
   legend style={at={(0.5,-0.36)},anchor=north,font=\tiny,legend columns=2,draw=none,column sep=4pt}}}
\begin{tikzpicture}
 \begin{axis}[dblax,name=src,ylabel={$[a/n]$},
   title={\scriptsize source block: $p_t(b)-\frac1n$, one cycle per period}]
  \draw[gray] (axis cs:0,0) -- (axis cs:6.2832,0);
  \addplot[domain=0:6.2832,samples=80,thick,red!75!black] {cos(deg(x))};
  \addlegendentry{$b=1$}
  \addplot[domain=0:6.2832,samples=80,thick,teal!80!black] {-cos(deg(x))};
  \addlegendentry{$b=-1$}
  \addplot[domain=0:6.2832,samples=80,thick,olive!80!black] {sin(deg(x))};
  \addlegendentry{$b=i$}
  \addplot[domain=0:6.2832,samples=80,thick,violet!80!black] {-sin(deg(x))};
  \addlegendentry{$b=-i$}
 \end{axis}
 \begin{axis}[dblax,at={(src.east)},anchor=west,xshift=1.6cm,ylabel={$[a^2/n]$},
   title={\scriptsize combination: two cycles per period}]
  \draw[gray] (axis cs:0,0) -- (axis cs:6.2832,0);
  \addplot[domain=0:6.2832,samples=80,thin,red!60] {cos(deg(x))^2};
  \addlegendentry{$p_t(1)+p_t(-1)-\frac2n$}
  \addplot[domain=0:6.2832,samples=80,thin,olive!70] {sin(deg(x))^2};
  \addlegendentry{$p_t(i)+p_t(-i)-\frac2n$}
  \addplot[domain=0:6.2832,samples=80,very thick,black] {cos(2*deg(x))};
  \addlegendentry{$\Rea Q_t$: difference of the two}
  \addplot[domain=0:6.2832,samples=80,very thick,black,densely dashed] {sin(2*deg(x))};
  \addlegendentry{$\Ima Q_t$: from the diagonals}
 \end{axis}
\end{tikzpicture}
\caption{Frequency doubling as a linear combination of probabilities, drawn for $\delta=0$ and to leading order in $a$. Left: while the source score $x_t\approx a\lambda^t$ goes around once, each of its action probabilities oscillates once, and opposite actions oscillate in antiphase. Right: summing opposite actions cancels these first-order oscillations and leaves the second-order terms $\frac{a^2}n\cos^2(2\pi t/M)$ and $\frac{a^2}n\sin^2(2\pi t/M)$. Their difference, which is $\Rea Q_t$ for the weights $q(\pm1)=1$ and $q(\pm i)=-1$, is $\frac{a^2}n\cos(4\pi t/M)$, a full cycle every half period. The diagonal actions, with weights $\pm\frac i2$, give $\Ima Q_t=\frac{a^2}n\sin(4\pi t/M)$ in the same way, so $Q_t\approx\frac{a^2}n\lambda^{2t}$ rotates at twice the source's speed. Note the different units: the doubled signal is a factor $a$ smaller.}
\label{fig:ov-doubling}
\end{figure}
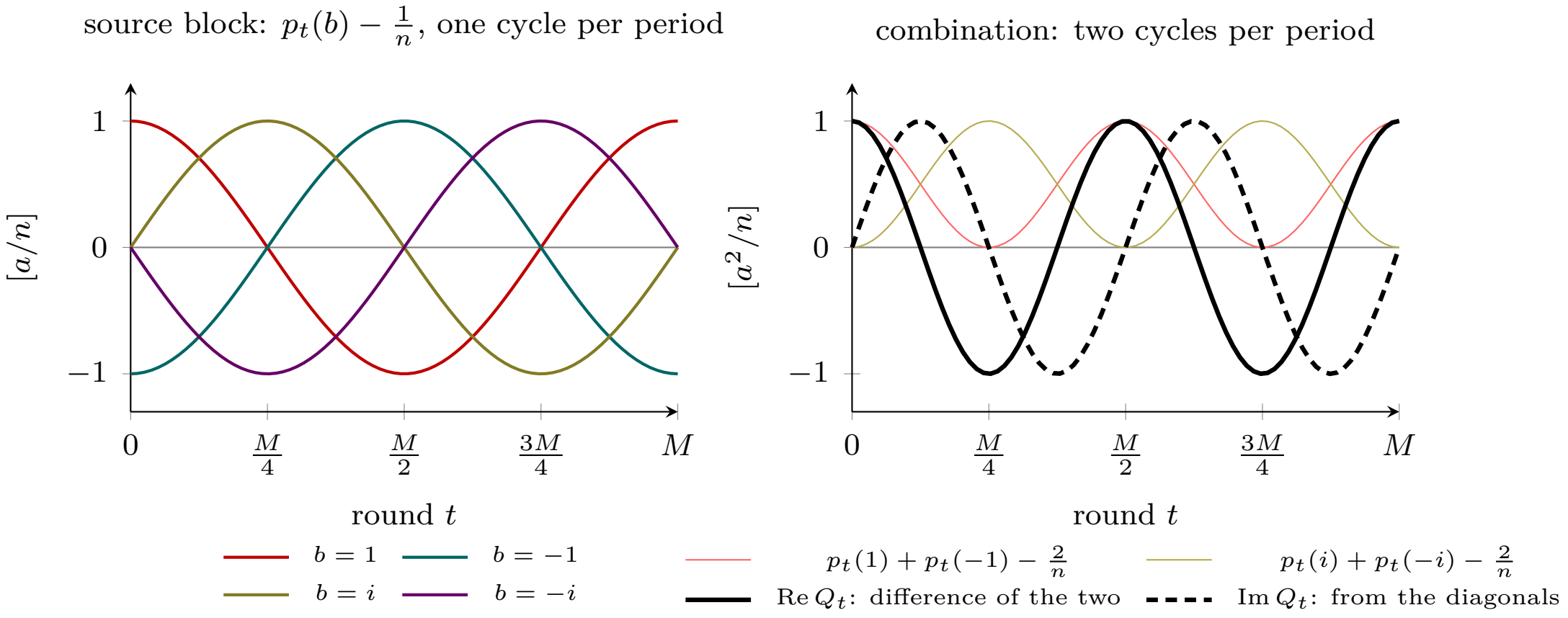

We assign fixed numbers to the source's actions whose expectation is approximately the square of its displacement from $\delta$.
\begin{lemma}[Quadratic extraction]\label{lem:forced-signal}
There is a complex-valued function $r_\delta$, fixed before play begins, on the source's actions such that its expectation $Q_\delta(x)$ under $p_x$ satisfies, for $\delta,u=O(a)$,
\begin{equation}\label{eq:Qtaylor}
 Q_\delta(\delta+u)=\frac{u^2}{n}+O(a^4).
\end{equation}
In particular, on the source orbit,
\[
 Q_\delta(x_t^\ast)=\frac{a^2}{n}\lambda^{2t}+O(a^3),
\]
with an $M$-periodic remainder uniform in $t$.
\end{lemma}
Represent the source's active actions by $b\in V=\{\pm1,\pm i,1+i,-1-i,1-i,-1+i\}$; its other $n-8$ actions are distinct labels with coordinate zero. Define
\[
 q(\pm1)=1,\quad q(\pm i)=-1,\quad
 q(1+i)=q(-1-i)=i/2,\quad
 q(1-i)=q(-1+i)=-i/2,
\]
and $q=0$ on neutral actions. Thus $q(b)=b^2$ on the four axis actions and $q(b)=b^2/4$ on the four diagonal ones (Figure~\ref{fig:actions}); the factor $\frac14$ compensates for the larger coordinates of the diagonal actions. For a source score $x$, recall that
\[
 p_x(b)=\frac{e^{\Rea(\overline b x)}}{Z(x)},\qquad
 Z(x)=\sum_b e^{\Rea(\overline b x)},
\]
where the sum is over distinct action labels, including all neutral actions. Set
\[
 r_\delta(b)=q(b)e^{-\Rea(\overline b\delta)},\qquad
 Q_\delta(x)=\mathbb E_{B\sim p_x}[r_\delta(B)]
 =\sum_b p_x(b)r_\delta(b).
\]
The factor involving $\delta$ shifts the numerator to the source's displacement $x-\delta$; the denominator remains $Z(x)$.
\begin{proof}
Write $u=u_1+iu_2$. Pairing opposite actions in the expectation gives
\begin{equation}\label{eq:Q}
 Q_\delta(\delta+u)
 =\frac{2\cosh u_1-2\cosh u_2+2i\sinh u_1\sinh u_2}{Z(\delta+u)}.
\end{equation}
The horizontal and vertical actions produce $u_1^2-u_2^2$ to second order; the diagonal actions produce $2iu_1u_2$. Thus the numerator is
\[
 u_1^2-u_2^2+2iu_1u_2+O(|u|^4)=u^2+O(|u|^4).
\]
Symmetry of the action set gives $Z(x)=n+O(|x|^2)$ near zero, proving \eqref{eq:Qtaylor}. On the source orbit, $u=a\lambda^t+O(a^2)$, so $u^2=a^2\lambda^{2t}+O(a^3)$. This proves the stated input estimate. Exact periodicity follows because $Q_\delta$ is a fixed function of the exactly periodic source score.
\end{proof}
Figure~\ref{fig:ov-doubling} shows the mechanism in terms of probabilities: opposite actions oscillate in antiphase, so their first-order oscillations cancel in $Q_\delta$, and only the second-order part, at twice the frequency, survives. Complex scores are essential here: a real oscillation $a\cos(\omega t)$ squares to $\frac{a^2}2(1+\cos2\omega t)$, whose constant part would shift the center of the next block.

\subsubsection*{Step 2: why the rotating input alone is insufficient}
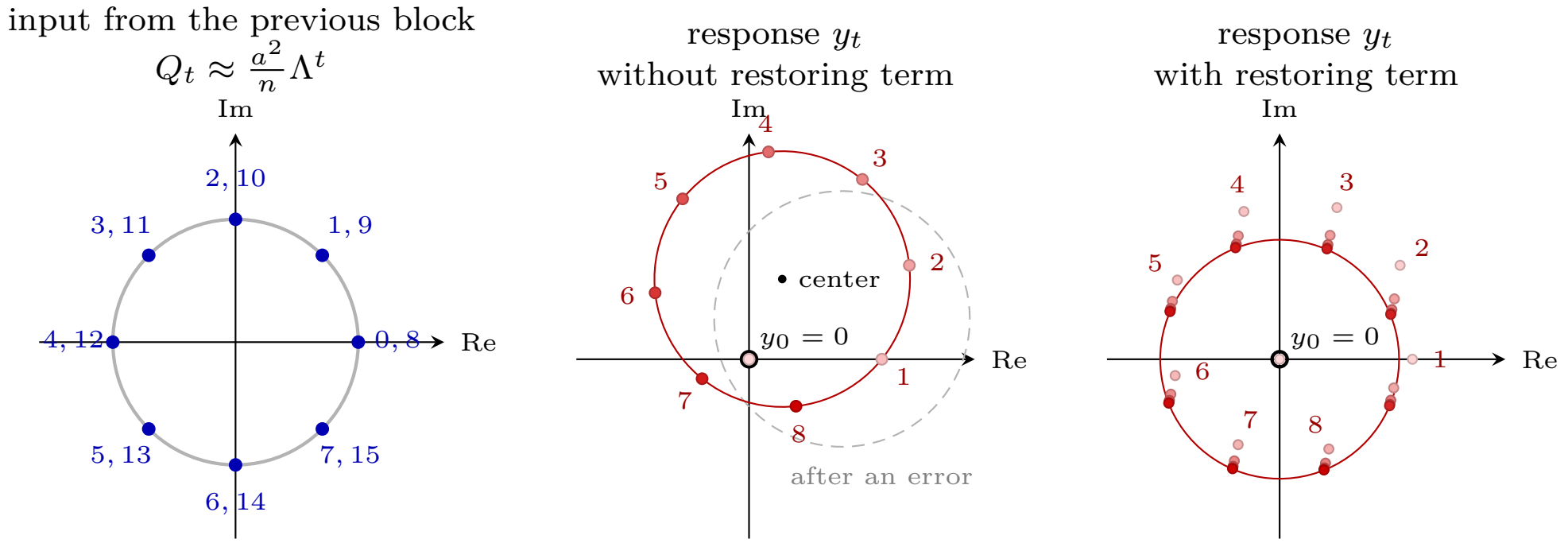
\begin{figure}[tbp]
\centering
\pgfplotsset{rspax/.style={width=0.3\textwidth,height=0.3\textwidth,axis equal image,
   axis lines=middle,font=\scriptsize,xtick=\empty,ytick=\empty,clip=false,
   xlabel={\tiny$\Rea$},ylabel={\tiny$\Ima$},
   xlabel style={anchor=west},ylabel style={anchor=south},
   title style={align=center,font=\scriptsize},
   colormap={fade}{color=(red!15) color=(red!80!black)}}}
\begin{tikzpicture}
 \begin{axis}[rspax,name=inp,xmin=-1.6,xmax=1.7,ymin=-1.6,ymax=1.7,
   title={input from the previous block\\ $Q_t\approx\frac{a^2}n\Lambda^t$}]
  \addplot[domain=0:360,samples=90,gray!60,thick] ({0.000+1.000*cos(x)},{0.000+1.000*sin(x)});
  \addplot[only marks,mark=*,mark size=1.4pt,blue!70!black] coordinates {(1.000,0.000) (0.707,0.707) (0.000,1.000) (-0.707,0.707) (-1.000,0.000) (-0.707,-0.707) (-0.000,-1.000) (0.707,-0.707)};
  \node[font=\tiny,blue!70!black] at (axis cs:1.320,0.000) {$0,8$};
  \node[font=\tiny,blue!70!black] at (axis cs:0.933,0.933) {$1,9$};
  \node[font=\tiny,blue!70!black] at (axis cs:0.000,1.320) {$2,10$};
  \node[font=\tiny,blue!70!black] at (axis cs:-0.933,0.933) {$3,11$};
  \node[font=\tiny,blue!70!black] at (axis cs:-1.320,0.000) {$4,12$};
  \node[font=\tiny,blue!70!black] at (axis cs:-0.933,-0.933) {$5,13$};
  \node[font=\tiny,blue!70!black] at (axis cs:-0.000,-1.320) {$6,14$};
  \node[font=\tiny,blue!70!black] at (axis cs:0.933,-0.933) {$7,15$};
 \end{axis}
 \begin{axis}[rspax,name=noc,at={(inp.east)},anchor=west,xshift=1.1cm,xmin=-2.6,xmax=3.4,ymin=-2.7,ymax=3.4,
   title={response $y_t$\\ without restoring term}]
  \addplot[domain=0:360,samples=90,gray!60,densely dashed] ({1.400+1.925*cos(x)},{0.607+1.925*sin(x)});
  \addplot[domain=0:360,samples=90,red!70!black,thin] ({0.500+1.925*cos(x)},{1.207+1.925*sin(x)});
  \addplot[scatter,only marks,mark=*,mark size=1.3pt,point meta=explicit] coordinates {(0.000,0.000) [0] (2.000,0.000) [1] (2.414,1.414) [2] (1.707,2.707) [3] (0.293,3.121) [4] (-1.000,2.414) [5] (-1.414,1.000) [6] (-0.707,-0.293) [7] (0.707,-0.707) [8]};
  \fill (axis cs:0.500,1.207) circle (1pt) node[right,font=\tiny]{center};
  \node[font=\tiny,gray,anchor=north] at (axis cs:2.000,-1.368) {after an error};
  \draw[black,thick] (axis cs:0,0) circle (1.8pt);
  \node[font=\tiny,anchor=south west,inner sep=1pt] at (axis cs:0.08,0.08) {$y_0=0$};
  \node[font=\tiny,red!60!black] at (axis cs:2.327,-0.263) {$1$};
  \node[font=\tiny,red!60!black] at (axis cs:2.832,1.459) {$2$};
  \node[font=\tiny,red!60!black] at (axis cs:1.970,3.034) {$3$};
  \node[font=\tiny,red!60!black] at (axis cs:0.248,3.539) {$4$};
  \node[font=\tiny,red!60!black] at (axis cs:-1.327,2.678) {$5$};
  \node[font=\tiny,red!60!black] at (axis cs:-1.832,0.955) {$6$};
  \node[font=\tiny,red!60!black] at (axis cs:-0.970,-0.620) {$7$};
  \node[font=\tiny,red!60!black] at (axis cs:0.752,-1.125) {$8$};
 \end{axis}
 \begin{axis}[rspax,name=wc,at={(noc.east)},anchor=west,xshift=1.1cm,xmin=-2.6,xmax=3.4,ymin=-2.7,ymax=3.4,
   title={response $y_t$\\ with restoring term}]
  \addplot[domain=0:360,samples=90,red!70!black,thin] ({0.000+1.799*cos(x)},{0.000+1.799*sin(x)});
  \addplot[scatter,only marks,mark=*,mark size=1.1pt,point meta=explicit] coordinates {(0.000,0.000) [0] (2.000,0.000) [1] (1.814,1.414) [2] (0.863,2.283) [3] (-0.538,2.224) [4] (-1.540,1.192) [5] (-1.573,-0.246) [6] (-0.625,-1.286) [7] (0.741,-1.351) [8] (1.718,-0.432) [9] (1.728,0.909) [10] (0.760,1.865) [11] (-0.623,1.856) [12] (-1.615,0.872) [13] (-1.638,-0.526) [14] (-0.682,-1.530) [15] (0.691,-1.564) [16] (1.674,-0.617) [17] (1.690,0.747) [18] (0.727,1.724) [19] (-0.652,1.733) [20] (-1.640,0.764) [21] (-1.660,-0.619) [22] (-0.701,-1.612) [23] (0.675,-1.635) [24] (1.660,-0.679) [25] (1.677,0.693) [26] (0.716,1.676) [27] (-0.661,1.692) [28] (-1.648,0.729) [29] (-1.667,-0.651) [30] (-0.707,-1.639) [31] (0.669,-1.659) [32] (1.655,-0.700) [33] (1.673,0.675) [34] (0.712,1.661) [35] (-0.665,1.678) [36] (-1.651,0.717) [37] (-1.670,-0.661) [38] (-0.709,-1.648) [39] (0.667,-1.667) [40]};
  \draw[black,thick] (axis cs:0,0) circle (1.8pt);
  \node[font=\tiny,anchor=south west,inner sep=1pt] at (axis cs:0.08,0.08) {$y_0=0$};
  \node[font=\tiny,red!60!black] at (axis cs:2.420,0.000) {$1$};
  \node[font=\tiny,red!60!black] at (axis cs:2.145,1.672) {$2$};
  \node[font=\tiny,red!60!black] at (axis cs:1.011,2.676) {$3$};
  \node[font=\tiny,red!60!black] at (axis cs:-0.637,2.633) {$4$};
  \node[font=\tiny,red!60!black] at (axis cs:-1.872,1.449) {$5$};
  \node[font=\tiny,red!60!black] at (axis cs:-1.158,-0.181) {$6$};
  \node[font=\tiny,red!60!black] at (axis cs:-0.441,-0.908) {$7$};
  \node[font=\tiny,red!60!black] at (axis cs:0.539,-0.983) {$8$};
 \end{axis}
\end{tikzpicture}
\caption{How the next block uses the doubled signal, for linearized dynamics without errors, drawn for $M=16$; the response is shown in rounds $t\ge0$, from light (early) to dark (late), starting from the uniform state $y_0=0$ (open circle); the numbers mark the rounds $t=1,\dots,8$ of the first period of the input. Left: the input $Q_t\approx\frac{a^2}n\Lambda^t$, $\Lambda=\lambda^2$, produced by the previous block (Figure~\ref{fig:ov-doubling}); it repeats every $M/2$ rounds. Middle: if the block only integrates this input ($\kappa=0$), its score $y_t$ immediately goes around a circle at the input's speed, but the center of that circle is set by the initial state rather than by the game. Nothing pulls the score back, so any error moves the circle permanently (dashed), and errors accumulate. Right: with the restoring self-interaction $-\kappa c(w)$, all state differences contract, and the score converges from the uniform start to a single circle centered at $0$, which it then follows with a fixed phase lag behind the input. The restoring strength is exaggerated ($\gamma=\eta d=0.15$) to make the convergence visible within a few periods.}
\label{fig:ov-response}
\end{figure}

Generating a rotation is not the same as making it attract trajectories.
\begin{lemma}[No attraction without feedback]\label{lem:forced-no-feedback}
If the new block's utility coefficient is a prescribed sequence $f_t$, independent of its score, two trajectories with the same initial memory and the same input retain their initial score difference forever. Moreover, if $f_t$ is $M$-periodic, then after the initial memory is consistent with the input, the score displacement over a period is
\[
 \eta\sum_{j=0}^{M-1}f_j.
\]
\end{lemma}
\begin{proof}
Utilities $\Rea(\overline v\nu r_\delta(b))$ give $f_t=\nu Q_\delta(x_t)$. Once the memory equals $f_{t-1}$, the score update is
\[
 y_{t+1}-y_t=\eta(2f_t-f_{t-1}).
\]
Its increment does not depend on $y_t$, so subtracting two updates preserves the score difference, including the first update when the initial memories agree. Summing over a full period with consistent memory gives the displayed displacement, since the shifted sum of $f_{t-1}$ equals the sum of $f_t$.
\end{proof}
For a pure rotating input $f_t=\phi\Lambda^t$, where $\Lambda=\lambda^2\ne1$, the recurrence admits the exact rotating solution
\[
 y_t=\psi\Lambda^t,\qquad
 \psi=\frac{\eta\phi(2-\Lambda^{-1})}{\Lambda-1}.
\]
Here $\phi$ is the input amplitude, $\psi$ is the response coefficient, and the initial memory is chosen consistently with the periodic input. However, every constant translate is also a solution: there is no preferred center and no attraction. Further, small corrections to the input can have nonzero mean and cause drift. We therefore add feedback that removes score offsets rather than merely integrating the source signal. Figure~\ref{fig:ov-response} compares the two cases for the linearized dynamics.

\subsubsection*{Step 3: add restoring utilities and compute the linear rotation}
\begin{lemma}[Stable linear response]\label{lem:forced-linear}
Fix $\kappa=\nu=1/64$, put $d=2\kappa/n$, $\gamma=\eta d$, and $\Lambda=\lambda^2$. For the linear utility coefficient
\[
 g_t=-d y_t+\frac{\nu a^2}{n}\Lambda^t,
\]
the score-and-memory recurrence \eqref{eq:oh} has an exact rotating solution with score $y_t=c_1a^2\Lambda^t$, where
\[
 c_1=\frac{(\eta\nu/n)(2-\Lambda^{-1})}
 {\Lambda-1+\gamma(2-\Lambda^{-1})}\ne0.
\]
Its consistent memory is $\mem_t=(\nu/n-dc_1)a^2\Lambda^{t-1}$. Every other solution approaches this one exponentially.
\end{lemma}
To realize the restoring term, the new players have the same action labels and probabilities $p_y$ as in \eqref{eq:pz}. Write $v$ for the new player's action, $w$ for her paired opponent's action, and $b$ for the source player's action, all at the same round. Recall $c(w)=w$ for $w\in\{\pm1,\pm i\}$ and $c(w)=0$ otherwise. If the source chooses $b$, define
\begin{equation}\label{eq:gain2}
 U(v;w,b)=\Rea\!\left(\overline v[-\kappa c(w)+\nu r_\delta(b)]\right).
\end{equation}
These utilities lie in $[-\frac12,\frac12]$ for small $a$: $|v|\le\sqrt2$, $|c(w)|\le1$, and $|r_\delta(b)|\le e^{\sqrt2|\delta|}$.

The negative sign penalizes alignment with the opponent's current-block action. By symmetry the two new players stay identical. Averaging over their paired opponent and the source yields expected utility $\Rea(\overline v g_t(y))$, with
\begin{equation}\label{eq:g2}
 g_t(y)=-\kappa\frac{2[\sinh(\Rea y)+i\sinh(\Ima y)]}{Z(y)}
          +\nu Q_\delta(x_t).
\end{equation}
The exponential-family calculation in Lemma~\ref{lem:block-realization} now gives exactly
\[
 y_{t+1}=y_t+\eta(2g_t(y_t)-\mem_t),\qquad
 \mem_{t+1}=g_t(y_t),\qquad (y_0,\mem_0)=(0,0).
\]
Near zero the first term of \eqref{eq:g2} is $-dy+O(|y|^3)$. With the source on its orbit, Lemma~\ref{lem:forced-signal} therefore gives
\[
 g_t(y)=-dy+\frac{\nu a^2}{n}\Lambda^t+O(|y|^3+a^3).
\]
\begin{proof}[Proof of Lemma~\ref{lem:forced-linear}]
A rotating sequence keeps its form under time shifts: shifting one round multiplies it by $\Lambda$. This reduces the recurrence to an equation for one complex coefficient. Substitute $y_t=c_1a^2\Lambda^t$ and the consistent memory into the linear recurrence. Canceling $a^2\Lambda^t$ gives
\[
 c_1(\Lambda-1)=-\gamma c_1(2-\Lambda^{-1})
                  +\frac{\eta\nu}{n}(2-\Lambda^{-1}),
\]
which gives the stated coefficient. To justify division and attraction, subtract two solutions. Their state difference evolves by
\[
 \mathsf B=\begin{pmatrix}1-2\gamma&-\eta\\-\gamma/\eta&0\end{pmatrix}.
\]
Its eigenvalues are $r_\pm=(1-2\gamma\pm\sqrt{1+4\gamma^2})/2$. For $0<\gamma<2/3$, one has $0<r_+<1$ and $-1<r_-<0$; our parameter choices satisfy this restriction. Thus powers of $\mathsf B$ decay exponentially. A zero denominator in the formula for $c_1$ would make $\Lambda$, of modulus one, an eigenvalue of $\mathsf B$, a contradiction. The numerator is nonzero since $|\Lambda^{-1}|=1$. Hence the response is a nonzero rotation and all state differences decay.
\end{proof}

Lemma~\ref{lem:forced-linear} is the linear part of the argument. The nonlinear terms, the exact periodic response and the convergence from the uniform start are handled in Section~\ref{sec:lower-joint}, jointly for all blocks; no stable-manifold matching is needed for these blocks, because all their linearized directions contract.

\subsection{The observer}\label{sec:lower-observer}

We now use the oscillation of players $1$ and $2$ to give player $3$ linear regret. Player $3$ has two groups of actions with opposite utility advantages. Each fixed action has average utility $0$, but the changing strategy of Optimistic Hedge responds with harmful timing to a sufficiently fast oscillation. By Remark~\ref{rem:criterion} at the end of this subsection, a mode of frequency $\omega$ increases regret exactly when $\cos\omega<\frac12$; this is why the slow first block is useless here and the last block runs at $\omega=\pi/2$.

Throughout this subsection, assume that players $1$ and $2$ have an exact $M$-periodic orbit approached exponentially from initialization. Suppose its last block, block $k$, has score
\[
 y_{k,t}^\ast=c_k\,a^K\lambda^{Kt}+O(a^{K+1}),\qquad
 K=2^{k-1},\quad c_k\ne0,\quad \lambda^K=e^{i\pi/2}.
\]
These are hypotheses here; Theorem~\ref{thm:cascade} in Section~\ref{sec:lower-joint} provides them. All constants below may depend on the fixed game parameters. We choose $a$ first and then $\eps$.

\subsubsection*{Step 1: the signal and player 3's utilities}
We turn the last block's rotation into a mean-zero signal and use it to define player $3$'s utilities.
\begin{lemma}[The observer's signal]\label{lem:observer-signal}
There is a bounded real function $F$ on player $1$'s actions such that $f_t=\mathbb E[F(W_t)]$ satisfies
\begin{equation}\label{eq:f}
 f_{t+M}^\ast=f_t^\ast,\qquad
 \sum_{t=0}^{M-1}f_t^\ast=0,\qquad
 |f_t-f_t^\ast|\le C\rho^t,\quad \rho<1,
\end{equation}
and, for some fixed phase $\theta$,
\begin{equation}\label{eq:fquarter}
 f_t^\ast=\alpha\cos(\pi t/2+\theta)+O(a^{K+1}),
 \qquad \alpha=\Theta(a^K)>0.
\end{equation}
In particular, the partial sums of $f_t$ are bounded uniformly in time.
\end{lemma}
\begin{proof}
Let $H$ equal $1$ on action $(k,1)$, $-1$ on action $(k,-1)$, and zero otherwise. Thus its expectation is the difference between these two action probabilities. Define the fixed constant
\[
 b=\frac1M\sum_{t=0}^{M-1}\mathbb E_{p_t^\ast}[H],\qquad F=H-b.
\]
Subtracting $b$ makes the orbit average exactly zero; this is a fixed utility-table adjustment, not a time-dependent rule. On the orbit the uncentered expectation is
\[
 \frac{2\sinh(\Rea y_{k,t}^\ast)}{Z_t^\ast}.
\]
All scores are $O(a)$, so the shared normalizer satisfies $Z_t^\ast=n+O(a^2)$. Expanding the numerator gives the leading term $(2/n)\Rea(c_ka^K e^{i\pi t/2})$, with error $O(a^{K+1})$. Its leading term has mean zero because $M$ is divisible by four. Centering changes only the error term. This proves \eqref{eq:fquarter}, with $\alpha=2|c_k|a^K/n$ and $\theta=\arg c_k$.

The expectation of a fixed bounded function is smooth in the scores, so exponential approach of the players' state to its orbit gives the last bound in \eqref{eq:f}. A periodic mean-zero sequence has bounded partial sums, and adding an exponentially decaying sequence preserves this property.
\end{proof}

\begin{lemma}[Fixed actions have bounded excess utility]\label{lem:observer-benchmark}
Split player $3$'s $n$ actions into nonempty groups $+$ and $-$. Give every action in these groups, respectively, utilities
\[
 U_3(+,w)=\eps F(w),\qquad
 U_3(-,w)=-\eps F(w),
\]
where $w$ is player $1$'s action. For sufficiently small $\eps>0$, these utilities lie in $[-\frac12,\frac12]$, and every fixed action has cumulative loss $O(1)$.
\end{lemma}
\begin{proof}
Boundedness of $F$ ensures valid utilities if $\eps\|F\|_\infty\le\frac12$. Expected losses are $\mp\eps f_t$. Their cumulative sums are bounded by Lemma~\ref{lem:observer-signal}. These utilities do not affect players $1$ and $2$, whose utilities are independent of player $3$.
\end{proof}
Our main claim is the following.
\begin{proposition}[Linear regret of the observer]\label{prop:observer}
Under the orbit hypotheses above, the utilities of Lemma~\ref{lem:observer-benchmark} give, for sufficiently small $a>0$ and then sufficiently small $\eps>0$,
\[
 \Reg_3(T)=c_{\mathrm{obs}}T+O(1),\qquad c_{\mathrm{obs}}>0.
\]
\end{proposition}

\subsubsection*{Step 2: the observer's response}
\begin{lemma}[Log-odds update]\label{lem:observer-odds}
Let $r_t=\log(\Pr_t(+)/\Pr_t(-))$, and let $r_0=\log(|+|/|-|)$ at uniform initialization. With $f_{-1}=0$, Optimistic Hedge gives
\begin{equation}\label{eq:r}
 \begin{split}
 r_{t+1}&=r_t+2\eta\eps(2f_t-f_{t-1}),\\
 r_t&=r_0+2\eta\eps\left(\sum_{s<t}f_s+f_{t-1}\right).
 \end{split}
\end{equation}
\end{lemma}
\begin{proof}
Actions in the same group have identical utility sequences, so their relative weights remain constant. The utility difference between groups is $2\eps f_t$; taking the log ratio in the Optimistic Hedge update gives the first identity. Summing it gives the second, using zero initial utility memory.
\end{proof}
Figure~\ref{fig:ov-observer} in Section~\ref{sec:ov-lower} shows the resulting timing mistake for equal groups and the signal $\alpha,0,-\alpha,0,\dots$: at round two the observer still prefers $+$ although it is now bad. The next steps prove that this effect persists for every phase $\theta$, unequal groups and the error terms of the actual signal.

\subsubsection*{Step 3: regret through the signal's timing}
\begin{lemma}[Regret and signal correlation]\label{lem:observer-correlation}
Write $S_t=\sum_{s<t}f_s$. Then
\begin{equation}\label{eq:reg3}
 \Reg_3(T)=\eps|S_T|-\eps\sum_{t<T}f_t\tanh(r_t/2),
\end{equation}
and
\[
 \Reg_3(T)=-\eta\eps^2\sech^2(r_0/2)
 \sum_{t<T}f_t(S_t+f_{t-1})+O(1)+O(\eta^2\eps^3T).
\]
\end{lemma}
\begin{proof}
The probability imbalance between groups is $\tanh(r_t/2)$, so the learner's expected loss is $-\eps f_t\tanh(r_t/2)$. The best fixed action has cumulative loss $-\eps|S_T|$. Subtraction proves \eqref{eq:reg3}.

By bounded partial sums and \eqref{eq:r}, $r_t-r_0=O(\eta\eps)$. Taylor expansion gives
\[
 \tanh(r_t/2)=\tanh(r_0/2)
 +\tfrac12\sech^2(r_0/2)(r_t-r_0)
 +O((r_t-r_0)^2).
\]
The constant term sums to $O(1)$ because $S_T$ is bounded. Substituting \eqref{eq:r} into the linear term gives the stated correlation, and the quadratic remainder contributes $O(\eta^2\eps^3T)$. Thus the sign is controlled, to leading order, by how the current signal correlates with accumulated past signals and the latest observation.
\end{proof}

\subsubsection*{Step 4: the leading oscillation gives positive regret}
\begin{lemma}[Positive regret rate]\label{lem:observer-positive}
For the signal in \eqref{eq:fquarter}, the limiting regret rate exists and satisfies
\begin{equation}\label{eq:slope}
 \lim_{T\to\infty}\frac{\Reg_3(T)}T
 =\eta\eps^2\sech^2(r_0/2)
 \left(\frac{\alpha^2}{4}+O(a^{2K+1})\right)
 +O(\eta^2\eps^3).
\end{equation}
The rate is positive for small $a$ and then small $\eps$, and regret differs from this rate times $T$ by $O(1)$.
\end{lemma}
\begin{proof}
We can compute the required correlation without introducing Fourier notation. Since $S_{t+1}=S_t+f_t$,
\[
 2f_tS_t=S_{t+1}^2-S_t^2-f_t^2.
\]
Summing telescopes. Boundedness of $S_T$ therefore yields
\[
 -\sum_{t<T}f_t(S_t+f_{t-1})
 =\frac12\sum_{t<T}f_t^2-\sum_{t<T}f_tf_{t-1}+O(1).
\]
The summable transient in \eqref{eq:f} changes the right-hand side by $O(1)$. Its average consequently equals the orbit average of $\tfrac12(f_t^\ast)^2-f_t^\ast f_{t-1}^\ast$.

For $\bar f_t=\alpha\cos(\pi t/2+\theta)$, averaging over four rounds gives
\[
 \frac14\sum_{t=0}^3\bar f_t^2=\frac{\alpha^2}{2},\qquad
 \frac14\sum_{t=0}^3\bar f_t\bar f_{t-1}=0.
\]
Thus the squared-signal term contributes $\alpha^2/4$, while the lag-one correlation contributes zero, regardless of $\theta$. Replacing $\bar f_t$ by $f_t^\ast=\bar f_t+O(a^{K+1})$ changes either average by $O(a^{2K+1})$, since $\alpha=\Theta(a^K)$. Lemma~\ref{lem:observer-correlation} gives the expansion in \eqref{eq:slope}. Choose $a$ so that the parenthesis is positive, then $\eps$ so that the cubic error is smaller than its positive quadratic term. Unequal group sizes merely change the strictly positive factor $\sech^2(r_0/2)$.

It remains to justify the exact rate and the $O(1)$ remainder. The actual signal equals a mean-zero periodic signal plus an exponentially decaying error. Thus its partial sums, and then $r_t$ by \eqref{eq:r}, approach periodic sequences with exponentially decaying errors; the partial-sum limit includes a constant shift from the transient. Smoothness of $\tanh$ shows that $\eps f_t\tanh(r_t/2)$ also approaches a periodic sequence with a summable error. Its cumulative sum is its period average times $T$, plus $O(1)$. The fixed-action correction $\eps|S_T|$ remains bounded. This proves the exact rate and bounded remainder, and justifies taking the limit in the expansion.
\end{proof}
\begin{proof}[Proof of Proposition~\ref{prop:observer}]
Combine Lemmas~\ref{lem:observer-benchmark}, \ref{lem:observer-correlation} and~\ref{lem:observer-positive}.
\end{proof}

\begin{remark}[Which oscillations hurt the observer]\label{rem:criterion}
The argument of Steps 3 and 4 applies to any mean-zero periodic signal $f$, up to a summable transient. By Lemma~\ref{lem:observer-correlation}, the regret rate is $-\eta\eps^2\sech^2(r_0/2)$ times the period average of $f_t(S_t+f_{t-1})$. The map $f\mapsto S_t+f_{t-1}$ multiplies the Fourier mode $e^{i\omega t}$ by $1/(e^{i\omega}-1)+e^{-i\omega}$, whose real part is $\cos\omega-\frac12$, so by Parseval's identity
\begin{equation}\label{eq:ov-slope}
 \lim_{T\to\infty}\frac{\Reg_3(T)}{T}\;=\;\eta\eps^2\sech^2(r_0/2)\sum_{\omega\ne0}\Bigl(\tfrac12-\cos\omega\Bigr)\,|\hat f(\omega)|^2+O(\eta^2\eps^3).
\end{equation}
The integrator $1/(e^{i\omega}-1)$ is the Hedge part of the update, and its real part $-\frac12$ makes every frequency contribute positively. The term $e^{-i\omega}$ is the optimistic correction. For slow signals $\cos\omega\approx1$ and the rate is negative: the observer anticipates the next loss and beats the best fixed action, which is the mechanism behind the upper bounds. For $\omega>\pi/3$ the correction is out of phase and the regret grows linearly. This is why the first block, of frequency $2\pi/M\approx0$, cannot be used directly, and why the cascade ends at $\omega=\pi/2$, where $\frac12-\cos\omega=\frac12$.
\end{remark}

\subsection{The cascade with a shared normalizer}\label{sec:lower-joint}

In the three-player game all $k$ blocks are carried by players~$1$ and~$2$ themselves. Player~$1$ (and symmetrically player~$2$) has the actions $(j,w)$ for $1\le j\le k$ and $w\in V$, together with $n-8k\ge1$ neutral actions, and her mixed strategy is parametrized by $k$ complex scores through
\[
 p(j,w)=\frac{e^{\Rea(\overline w z_j)}}{Z(z_1,\dots,z_k)},\qquad Z(z_1,\dots,z_k)=\sum_{j=1}^k\sum_{w\in V}e^{\Rea(\overline w z_j)}+(n-8k).
\]
The utility of action $(j,v)$ against the opponent's action $(j',w)$ is given by \eqref{eq:gain1} if $j=j'=1$, by the self-interaction $\Rea(\overline v[-\kappa c(w)])$ if $j=j'\ge2$, by the coupling $\Rea\!\left(\overline v\nu q(w)e^{-\Rea(\overline w\delta_{j-1})}\right)$ if $j'=j-1$, where $\delta_1=\delta$ and $\delta_j=0$ for $j\ge2$, and equals $0$ otherwise; neutral actions have utility $0$ (Figure~\ref{fig:utility-table}). Here $\kappa=\nu=1/64$ as in Lemma~\ref{lem:forced-linear}. Unlike in Section~\ref{sec:lower-second}, the paired opponent of block $j$ and its source are now the same player.

For $u=u_1+iu_2\in\C$ write
\[
 S(u)=\sinh u_1+i\sinh u_2,\qquad
 \Psi(u)=\sum_{w\in V}q(w)e^{\Rea(\overline wu)}=2\cosh u_1-2\cosh u_2+2i\sinh u_1\sinh u_2,
\]
and, for $z=(z_1,\dots,z_k)$,
\begin{equation}\label{eq:gjoint}
 g_1(z)=\frac{2\mu S(z_1-\delta)}{Z(z)},\qquad
 g_j(z)=\frac{-2\kappa S(z_j)+\nu\Psi(z_{j-1}-\delta_{j-1})}{Z(z)}\quad(2\le j\le k).
\end{equation}
These are \eqref{eq:g} and \eqref{eq:g2} with the shared normalizer in place of the block-specific ones; for instance $p(j-1,w)q(w)e^{-\Rea(\overline w\delta_{j-1})}=q(w)e^{\Rea(\overline w(z_{j-1}-\delta_{j-1}))}/Z(z)$. Since an opponent action lies in exactly one block, and expected utility is linear in the opponent's mixed strategy, the expected utility of action $(j,v)$ is $\Rea(\overline vg_j(z))$. The calculation of Lemma~\ref{lem:block-realization} therefore applies to every block at once: the strategies keep the displayed form, the two players remain identical, and their joint state $s=(z_j,\mem_j)_{j\le k}\in\C^{2k}$ evolves by
\begin{equation}\label{eq:joint}
 z_{j,t+1}=z_{j,t}+\eta\bigl[2g_j(z_t)-\mem_{j,t}\bigr],\qquad\mem_{j,t+1}=g_j(z_t),\qquad s_0=0.
\end{equation}
Denote this one-step map by $\Phi_{\mu,\delta}$. Write $K_j=2^{j-1}$ and $\Lambda_j=\lambda^{K_j}$, so that $\Lambda_1=\lambda$ and $\Lambda_k=e^{i\pi/2}$ when $M=2^{k+1}$.

\begin{theorem}[The cascade]\label{thm:cascade}
Let $k\ge1$, $n\ge8k+1$ and $M=2^{k+1}$, and suppose $0<\eta\le1/100$ and $\eta M\ge16\pi n$. For every sufficiently small $a>0$ there are $\mu,\delta\in\C$, with $\delta=O(a)$, for which all utilities lie in $[-\frac12,\frac12]$ and \eqref{eq:joint} has an orbit $s_t^\ast=(z_{j,t}^\ast,\mem_{j,t}^\ast)_{j\le k}$ of exact period $M$ with
\[
 z_{1,t}^\ast=\delta+a\lambda^t+O(a^2),\qquad
 z_{j,t}^\ast=c_ja^{K_j}\Lambda_j^t+O(a^{K_j+1})\quad(2\le j\le k),
\]
where the constants $c_j\ne0$ do not depend on $a$ and the remainders are $M$-periodic. The trajectory from the uniform start satisfies
\[
 \|s_t-s_t^\ast\|\le Ca\rho^t\qquad(t\ge0)
\]
for some $C<\infty$ and $\rho<1$, independent of $t$ and of sufficiently small $a$.
\end{theorem}

\paragraph{Feedback through the normalizer.}
Sections~\ref{sec:lower-first} and~\ref{sec:lower-second} gave each block its own normalizer, so information flowed only forward, from block $j-1$ to block $j$. The shared normalizer couples the blocks in both directions: later blocks act back on earlier ones, including block~$1$, through $Z$. This feedback is harmless for three reasons. First, it does not change the form of the strategies: block $j$'s probabilities remain proportional to $e^{\Rea(\overline wz_j)}$, and $Z$ only rescales each coefficient $g_j$. Second, it is small: block $j$ contributes $\sum_{w\in V}e^{\Rea(\overline wz_j)}-8=3|z_j|^2+O(|z_j|^4)$ to $Z$, so the blocks $j\ge2$ change $Z$ only by $O(a^4)$, and this change is nearly constant in time, because its leading term depends only on $|z_j|$. Third, $Z-n$ is quadratic in the scores, so the feedback disappears from the linearizations used below. The forward structure therefore survives only in these linearizations, and the proof applies the implicit function theorem and the stable manifold theorem to all blocks jointly rather than block by block.

\begin{figure}[t]
\centering
\begin{tikzpicture}[x=1.25cm,y=-0.8cm,
  cell/.style={font=\scriptsize,align=center}]
 \foreach \j/\l in {1/1,2/2,3/3,4/\cdots,5/k}{
   \node[cell] at (\j,0.15) {$j'=\l$};
   \node[cell,anchor=east] at (0.4,\j) {$j=\l$};}
 \node[cell,anchor=east] at (0.4,6) {neutral};
 \node[cell] at (6,0.15) {neutral};
 \node[font=\scriptsize] at (3.5,-0.55) {opponent's action $(j',w)$};
 \node[font=\scriptsize,rotate=90] at (-0.75,3.5) {own action $(j,v)$};
 \fill[blue!12] (0.5,0.5) rectangle (1.5,1.5);
 \node[cell] at (1,1) {\eqref{eq:gain1}};
 \foreach \j in {2,3,5}{
   \fill[gray!15] (\j-0.5,\j-0.5) rectangle (\j+0.5,\j+0.5);
   \node[cell] at (\j,\j) {$-\kappa c$};}
 \foreach \j in {2,3}{
   \fill[red!12] (\j-1.5,\j-0.5) rectangle (\j-0.5,\j+0.5);
   \node[cell] at (\j-1,\j) {$\nu q$};}
 \fill[red!12] (3.5,4.5) rectangle (4.5,5.5);
 \node[cell] at (4,5) {$\nu q$};
 \node[cell] at (4,4) {$\ddots$};
 \foreach \i in {0,...,6}{\draw[gray!60] (0.5,\i+0.5) -- (6.5,\i+0.5); \draw[gray!60] (\i+0.5,0.5) -- (\i+0.5,6.5);}
\end{tikzpicture}
\caption{Block structure of the utility table of players $1$ and $2$ in the cascade. Each entry gives the utility $\Rea(\overline v\,\cdot\,)$ of own action $(j,v)$ against the opponent's action $(j',w)$: the oscillator \eqref{eq:gain1} in block $(1,1)$ (blue), the restoring self-interaction $-\kappa c(w)$ on the rest of the diagonal (gray), and the coupling $\nu q(w)e^{-\Rea(\overline w\delta_{j-1})}$ to the preceding block on the subdiagonal (red). Blank entries, including all neutral rows and columns, have utility $0$.}
\label{fig:utility-table}
\end{figure}

\begin{proof}[Proof of Theorem~\ref{thm:cascade}]
For $k=1$ this is Theorem~\ref{thm:first}, so let $k\ge2$. We follow Steps 3--6 of Section~\ref{sec:lower-first}, with all blocks at once.

\emph{Rescaled residuals.} Each block is measured in units of its expected amplitude. The unknowns are $M$-periodic complex sequences $h_j=(h_{j,t})_{t\bmod M}$, $1\le j\le k$, and $\mu$; the parameters are $(a,\delta)$. Put
\[
 z_{1,t}=\delta+ah_{1,t},\qquad z_{j,t}=a^{K_j}h_{j,t}\quad(2\le j\le k),
\]
and define, for $1\le j\le k$ and $t\bmod M$,
\[
 \mathcal R_{j,t}=h_{j,t+1}-h_{j,t}-\frac{\eta}{a^{K_j}}\bigl[2g_j(z_t)-g_j(z_{t-1})\bigr],
\]
together with the normalization $N$ of \eqref{eq:periodic} applied to $h_1$. This gives $kM+1$ complex equations in $kM+1$ complex unknowns. A zero with $a\ne0$ is an exact $M$-periodic orbit of \eqref{eq:joint}, with memories $\mem_{j,t}^\ast=g_j(z_{t-1}^\ast)$.

\emph{Analyticity across $a=0$.} The functions $S$ and $\Psi$ are real analytic, $S(0)=0$ with derivative the identity, and $\Psi(u)=u^2+O(|u|^4)$ by the proof of Lemma~\ref{lem:forced-signal}. Hence $S(\epsilon v)/\epsilon$ and $\Psi(\epsilon v)/\epsilon^2$ are real analytic in $(\epsilon,v)$, for real $\epsilon$, with values $v$ and $v^2$ at $\epsilon=0$. Now $z_{1,t}-\delta=ah_{1,t}$, $z_{j-1,t}-\delta_{j-1}=a^{K_{j-1}}h_{j-1,t}$ for $j\ge3$, and $K_j=2K_{j-1}$. So each quotient $g_j(z_t)/a^{K_j}$, and hence each residual, extends real analytically across $a=0$. At $a=0$ all scores except $z_1=\delta$ vanish, so $Z(z_t)=Z(\delta)$, the single-block normalizer of \eqref{eq:pz}, and
\[
 \frac{g_1(z_t)}a\Big|_{a=0}=\frac{2\mu h_{1,t}}{Z(\delta)},\qquad
 \frac{g_j(z_t)}{a^{K_j}}\Big|_{a=0}=\frac{-2\kappa h_{j,t}+\nu h_{j-1,t}^2}{Z(\delta)}\quad(j\ge2).
\]
In particular, at $a=0$ the residuals of block $1$ are those of \eqref{eq:periodic}, and those of block $j\ge2$ involve only $h_j$ and $h_{j-1}$. The normalizer feedback has disappeared: it enters $g_i$ through the scores $z_j=a^{K_j}h_j$ only at positive order in $a$.

\emph{Known zeros at $a=0$.} For each small $\delta$ put $h_{1,t}=\lambda^t$, $\mu=\mu_\ast(\delta)$, and $h_{j,t}=c_j(\delta)\Lambda_j^t$ for $j\ge2$, where $c_1(\delta)=1$ and
\[
 c_j(\delta)=\frac{(\eta\nu/Z(\delta))(2-\Lambda_j^{-1})}{\Lambda_j-1+\gamma_\delta(2-\Lambda_j^{-1})}\,c_{j-1}(\delta)^2,\qquad
 \gamma_\delta=\frac{2\eta\kappa}{Z(\delta)}.
\]
Block $1$ solves its equations by Lemma~\ref{lem:block-periodic-properties}. For $j\ge2$, $h_{j-1,t}^2=c_{j-1}^2\Lambda_{j-1}^{2t}=c_{j-1}^2\Lambda_j^t$, so the equations of block $j$ at $a=0$ are those of Lemma~\ref{lem:forced-linear} with $\Lambda=\Lambda_j$, $n$ replaced by $Z(\delta)$, and input coefficient $c_{j-1}^2$; the computation there gives $c_j$. Its denominator equals $(\Lambda_j-r_+)(\Lambda_j-r_-)/\Lambda_j$, where $r_\pm$ are the eigenvalues of $\mathsf B$ with $\gamma=\gamma_\delta\in(0,2/3)$; it is nonzero because $|r_\pm|<1=|\Lambda_j|$. Its numerator is nonzero, so $c_j(\delta)\ne0$ by induction. These sequences are $M$-periodic because $\Lambda_j^M=1$.

\begin{sloppypar}
\emph{Invertible Jacobian.} Order the unknowns as $(h_1,\mu,h_2,\dots,h_k)$ and the equations as $(\mathcal R_1,N,\mathcal R_2,\dots,\mathcal R_k)$. By the formulas at $a=0$, the Jacobian at $(a,\delta)=(0,0)$ and the known zero is block lower-triangular. Its first diagonal block is the Jacobian of Lemma~\ref{lem:block-periodic-properties}, which is invertible. For $j\ge2$ the diagonal block is
\end{sloppypar}
\[
 \dot h\mapsto\bigl(\dot h_{t+1}-\dot h_t+\gamma(2\dot h_t-\dot h_{t-1})\bigr)_t,\qquad\gamma=\frac{2\eta\kappa}n,
\]
which multiplies the Fourier mode $\zeta^t$, $\zeta^M=1$, by $(\zeta-r_+)(\zeta-r_-)/\zeta\ne0$. Hence the Jacobian is invertible. Theorem~\ref{thm:block-ift} gives real-analytic $h_j(a,\delta)$ and $\mu(a,\delta)$ near $(0,0)$, and local uniqueness identifies them at $a=0$ with the known zeros. Thus $h_{j,t}(a,\delta)=c_j(\delta)\Lambda_j^t+O(a)$, uniformly in $t$ and small $\delta$. For $a\ne0$ the orbit has exact period $M$, because block $1$ does (proof of Lemma~\ref{lem:block-periodic}).

\emph{Strong stable manifold.} At $a=\delta=0$ and $\mu=\mu_0$ the origin of $\C^{2k}$ is fixed by $\Phi_{\mu_0,0}$. Since $Z-n$ and $\Psi$ vanish to second order at the origin, neither the normalizer nor the coupling contributes to the derivative there, and
\[
 D\Phi_{\mu_0,0}(0)=\operatorname{diag}(A,\mathsf B,\dots,\mathsf B).
\]
This is invertible, since $\det A=\beta\ne0$ and $\det\mathsf B=-\gamma\ne0$. Its $M$th power has the complex eigenvalues $1$ and $\chi^M$ on block $1$ and $r_\pm^M$ on every other block. Fix $\bar\rho$ with $\max\{|\chi|,|r_+|,|r_-|\}^M<\bar\rho<1$. The strongly stable space is $E_s\oplus(\C^2)^{k-1}$, of real dimension $4k-2$, and its complement is $E_c$ in the coordinates of block $1$, of real dimension $2$. The proofs of Lemmas~\ref{lem:block-stable-properties} and~\ref{lem:block-stable} now apply verbatim to $G_p=\Phi^M_{\mu(p),\delta}$ and its fixed points $s_0^\ast(p)$, $p=(a,\delta)$. They give a local strong stable manifold
\[
 W^{ss}_{a,\delta}=\bigl\{s_0^\ast+\xi+\Gamma(a,\delta,\xi):\xi\in E_s\oplus(\C^2)^{k-1}\text{ small}\bigr\},
\]
with $\Gamma$ valued in $E_c$, jointly $C^1$, $\Gamma(a,\delta,0)=0$ and $D_\xi\Gamma(0,0,0)=0$, whose points converge to the orbit at a uniform rate $C\rho^t$ in every round.

\emph{Matching the uniform start.} Let $P_c$ and $P_s$ be the projections of the joint splitting. The uniform start $0$ lies on $W^{ss}_{a,\delta}$ exactly when $\Delta(a,\delta)=P_c(-s_0^\ast)-\Gamma(a,\delta,P_s(-s_0^\ast))=0$. At $a=0$ every score except $z_1=\delta$ vanishes, and every memory vanishes because each $g_j$ vanishes there; so $s_0^\ast(0,\delta)=((\delta,0),0,\dots,0)$. As in Lemma~\ref{lem:block-matching-properties}, $D_\delta\Delta(0,0)[d\delta]=-P_c((d\delta,0),0,\dots,0)$, which is multiplication by $-\lambda/(\lambda-\chi)\ne0$ in the coordinate $u_c$ on $E_c$. Theorem~\ref{thm:block-ift} gives a $C^1$ solution $\delta=\delta(a)=O(a)$. Since $\|s_0^\ast\|=O(a)$, the trajectory from the uniform start satisfies $\|s_t-s_t^\ast\|\le Ca\rho^t$.

\emph{Leading terms.} With $\delta=\delta(a)=O(a)$ we have $c_j(\delta(a))=c_j(0)+O(a)$. Setting $c_j=c_j(0)\ne0$ gives $z_{j,t}^\ast=a^{K_j}h_{j,t}=c_ja^{K_j}\Lambda_j^t+O(a^{K_j+1})$ for $j\ge2$, and $z_{1,t}^\ast=\delta+a\lambda^t+O(a^2)$.

\emph{Bounded utilities.} The utilities of block $1$ are bounded as in the proof of Theorem~\ref{thm:first}. An action $(j,v)$ with $j\ge2$ receives a nonzero utility only against opponent actions in block $j$, where it is at most $\sqrt2\kappa$ in absolute value, or in block $j-1$, where it is at most $\sqrt2\nu e^{\sqrt2|\delta|}$, since $|q|\le1$. Both are below $\frac12$ for small $a$. Neutral actions and all remaining pairs have utility $0$.
\end{proof}

\subsection{Proof of Theorem~\ref{thm:lower}}\label{sec:lower-assembly}

We first prove the second part of the theorem, linear regret for $\eta\ge C_0n2^{-n/8}$, and then deduce the first.
\begin{proof}[Proof of linear regret for $\eta\ge C_0n2^{-n/8}$]
Let $\eta_0=1/100$ and $C_0=16\pi$. Given $n\ge C_0$ (so that $n\ge9$) and $\eta\in[C_0n2^{-n/8},\eta_0]$, put $k=\lfloor(n-1)/8\rfloor\ge1$ and $M=2^{k+1}$, which is divisible by four. Since $k\ge n/8-1$,
\[
 \frac{16\pi n}{M}=16\pi n\,2^{-k-1}\le16\pi n\,2^{-n/8}\le\eta ,
\]
so Theorem~\ref{thm:cascade} applies. The $k$ blocks use $8k\le n-1$ active actions of each of players $1$ and $2$, leaving at least one neutral action; the last block oscillates at frequency $2^{k-1}\cdot2\pi/M=\pi/2$. By Theorem~\ref{thm:cascade}, players $1$ and $2$ converge exponentially to a joint periodic orbit whose last block satisfies the hypothesis of Section~\ref{sec:lower-observer}, and Proposition~\ref{prop:observer} furnishes the observer with linear regret. Her $n$ actions are split into two nonempty groups, so $G\in\cG(3,n)$.
\end{proof}

\begin{proof}[Proof of the exponential bound]
Let $0<\eta\le\eta_0$. If $\eta\ge C_0n2^{-n/8}$, the game just constructed has $\sup_T\Reg_3(T)=\infty$. If $\eta<C_0n2^{-n/8}$, let $G$ be the three-player game in which players~$1$ and~$2$ have constant losses and player~$3$'s loss is $0$ on one action and $1$ on the others, irrespective of the opponents' actions. By Proposition~\ref{prop:constloss}, $\sup_T\Reg_3(T)\ge\log n/\eta-1>2^{n/8}\log n/(C_0n)-1$. The bound on $\inf_{0<\eta\le\eta_0}\sup_{T\ge1}\,\sup_{G\in\cG(3,n)}\,\max_i\Reg_i(T)$ follows by taking the infimum over $\eta\in(0,\eta_0]$.
\end{proof}

The exponential form of the threshold is forced by the mechanism: the first block's period is at least of order $n/\eta$, each halving of the period costs eight actions, and the period must reach four.

\begin{remark*}
The bound with $\inf_\eta$ in Theorem~\ref{thm:lower} concerns $\inf_\eta\sup_T$, the natural order for a horizon-independent guarantee. The stronger statement with $\sup_T\inf_\eta$, in which the learning rate may be tuned to the horizon, holds as well provided the constants $c(G,\eta)$ and $C(G,\eta)$ in Theorem~\ref{thm:lower} are bounded away from zero and from infinity, respectively, uniformly for $\eta$ in compact subintervals of $(0,\eta_0]$; this uniformity is what the implicit-function arguments of Sections~\ref{sec:lower-first}--\ref{sec:lower-second} provide when $\eta$ is carried along as a parameter, and we omit the details.
\end{remark*}

\subsection{More than three players}\label{sec:lower-many}

The construction extends to $m\ge3$ players by splitting all players except the observer into pairs, each pair carrying its own segment of the cascade.

\begin{theorem}[An exponential barrier for $m$ players]\label{thm:lower-many}
Let $C_0$ and $\eta_0$ be as in Theorem~\ref{thm:lower}, let $m\ge3$, and put $P=\lfloor(m-1)/2\rfloor$. For every $n\ge C_0$ and every learning rate $\eta\in(0,\eta_0]$ there is an $m$-player game $G\in\cG(m,n)$ in which, when all players run Optimistic Hedge with learning rate $\eta$ starting from the uniform distribution,
\[
 \sup_T\,\Reg_m(T)\ \ge\ \frac{2^{1+P(n-8)/8}\log n}{C_0\,n}-1\ =\ 2^{\Omega(mn)} .
\]
Moreover, if $\eta\ge C_0\,n\,2^{-1-P(n-8)/8}$, then $G$ can be chosen so that player $m$'s regret grows linearly: $\Reg_m(T)\ge c\,T-C$ for all $T$, for some $c=c(G,\eta)>0$ and $C=C(G,\eta)$.
\end{theorem}
For $m\in\{3,4\}$ we have $P=1$ and $2^{1+(n-8)/8}=2^{n/8}$, so the bound is that of Theorem~\ref{thm:lower}.

\paragraph{The game.}
Write $k_0=\lfloor(n-1)/8\rfloor$ and $k=Pk_0$. For $1\le p\le P$, players $2p-1$ and $2p$ form \emph{pair} $p$, which carries the blocks $j$ with $(p-1)k_0<j\le pk_0$. Blocks are numbered consecutively across pairs, so the predecessor of block $j$ is block $j-1$ also when the two lie in different pairs. Each player of pair $p$ has the actions $(j,w)$, for the $k_0$ blocks $j$ of her pair and $w\in V$, together with $n-8k_0\ge1$ neutral actions. Her mixed strategy is parametrized by the scores of these blocks as in Section~\ref{sec:lower-joint}, with the normalizer of her own pair,
\[
 Z_p(z)=\sum_{(p-1)k_0<j\le pk_0}\ \sum_{w\in V}e^{\Rea(\overline wz_j)}+(n-8k_0).
\]
Within a pair, the utility of action $(j,v)$ against the partner's action $(j',w)$ is as in Section~\ref{sec:lower-joint}, except that the first block of each pair $p\ge2$ has only the self-interaction $\Rea(\overline v[-\kappa c(w)])$, against $j'=j$, and no coupling to the partner. Instead, this block $j=(p-1)k_0+1$ is driven by the last block of pair $p-1$ exactly as in Section~\ref{sec:lower-second}: each player of pair $p$ receives the additional utility $\Rea\bigl(\overline v\,\nu r_{\delta_{j-1}}(b)\bigr)$ from action $(j,v)$ when player $2p-3$ plays $(j-1,b)$, and no additional utility otherwise, where $r_\delta$ is the function of Lemma~\ref{lem:forced-signal}. Thus the utility of a player of pair $p\ge2$ is the sum of a term that depends on her partner's action and a term that depends on the action of player $2p-3$. Player $m$ is the observer of Section~\ref{sec:lower-observer}, with her signal taken from the last block $k$ of player $2P-1$ instead of player $1$. If $m$ is even, player $m-1$ has constant losses and influences nobody.

\paragraph{The cascade across pairs.}
Both players of a pair start at the uniform distribution and receive the same utilities, so they remain identical. As in Section~\ref{sec:lower-joint}, the joint state $s=(z_j,\mem_j)_{j\le k}$ of all pairs follows \eqref{eq:joint} with $k$ blocks, where now, for a block $j$ of pair $p$,
\begin{equation}\label{eq:gmany}
 g_j(z)=\begin{cases}
 \dfrac{2\mu S(z_1-\delta)}{Z_1(z)}, & j=1,\\[2.5ex]
 \dfrac{-2\kappa S(z_j)+\nu\Psi(z_{j-1}-\delta_{j-1})}{Z_p(z)}, & j\ge2\text{ not the first block of pair }p,\\[2.5ex]
 \dfrac{-2\kappa S(z_j)}{Z_p(z)}+\dfrac{\nu\Psi(z_{j-1}-\delta_{j-1})}{Z_{p-1}(z)}, & j\text{ the first block of pair }p\ge2.
 \end{cases}
\end{equation}
The last case is \eqref{eq:g2}: the source's signal is normalized by the source's own normalizer. A pair is influenced only by its own blocks and by the last block of the preceding pair. Hence the feedback through the normalizer described in Section~\ref{sec:lower-joint} stays inside each pair, and between pairs information flows strictly forward.

\begin{theorem}[The cascade across pairs]\label{thm:cascade-many}
The conclusions of Theorem~\ref{thm:cascade} hold for the dynamics \eqref{eq:joint} with the coefficients \eqref{eq:gmany}, for all $P\ge1$, $k_0\ge1$ and $n\ge8k_0+1$, with $k=Pk_0$ blocks and $M=2^{k+1}$, under the same conditions $0<\eta\le1/100$ and $\eta M\ge16\pi n$.
\end{theorem}

\begin{proof}
The proof of Theorem~\ref{thm:cascade} applies, with $k$ blocks, after three changes, all due to the separate normalizers. For a block $j\ge2$, let $Z^{\mathrm{own}}_j$ and $Z^{\mathrm{src}}_j$ be the values at $a=0$ of the normalizers of the pairs carrying blocks $j$ and $j-1$. Since at $a=0$ all scores except $z_1=\delta$ vanish, each of these equals $Z(\delta)$ for pair $1$ and $n$ for every other pair.

\emph{Residuals at $a=0$.} For $j\ge2$,
\[
 \frac{g_j(z_t)}{a^{K_j}}\Big|_{a=0}=-\frac{2\kappa h_{j,t}}{Z^{\mathrm{own}}_j}+\frac{\nu h_{j-1,t}^2}{Z^{\mathrm{src}}_j},
\]
so the residuals of block $j$ still involve only $h_j$ and $h_{j-1}$.

\emph{Known zeros.} They are $h_{j,t}=c_j(\delta)\Lambda_j^t$ with $c_1=1$ and
\[
 c_j(\delta)=\frac{(\eta\nu/Z^{\mathrm{src}}_j)(2-\Lambda_j^{-1})}{\Lambda_j-1+\gamma_j(2-\Lambda_j^{-1})}\,c_{j-1}(\delta)^2,\qquad\gamma_j=\frac{2\eta\kappa}{Z^{\mathrm{own}}_j}\in(0,\tfrac23),
\]
which are nonzero by the argument given there.

\emph{Bounded utilities.} An action of the first block of a pair $p\ge2$ receives a utility of absolute value at most $\sqrt2\kappa$ from its partner and at most $\sqrt2\nu e^{\sqrt2|\delta_{j-1}|}$ from player $2p-3$; their sum is below $\frac12$ for small $a$.

Everything else is unchanged. At $(a,\delta)=(0,0)$ all normalizers equal $n$, so the Jacobian is block lower-triangular with the same diagonal blocks as before. Moreover, $Z_p-n$ and $\Psi$ vanish to second order at the origin, so $D\Phi_{\mu_0,0}(0)=\operatorname{diag}(A,\mathsf B,\dots,\mathsf B)$. Finally, the matching equation still involves only the shift $\delta$ of block $1$.
\end{proof}

\begin{proof}[Proof of Theorem~\ref{thm:lower-many}]
Let $\eta_0=1/100$, $C_0=16\pi$, $n\ge C_0$ and $k_0=\lfloor(n-1)/8\rfloor\ge1$. Since $k_0\ge(n-8)/8$, the $k=Pk_0$ blocks give $M=2^{k+1}\ge2^{1+P(n-8)/8}$.

If $\eta\ge C_0n2^{-1-P(n-8)/8}$, then $\eta M\ge16\pi n$, so Theorem~\ref{thm:cascade-many} applies. The last block oscillates at frequency $2^{k-1}\cdot2\pi/M=\pi/2$, and Proposition~\ref{prop:observer} gives player $m$ linear regret; in particular $\sup_T\Reg_m(T)=\infty$.

If $\eta<C_0n2^{-1-P(n-8)/8}$, let players $1,\dots,m-1$ have constant losses, and let player $m$'s loss be $0$ on one action and $1$ on the others, irrespective of the opponents' actions. By Proposition~\ref{prop:constloss}, $\sup_T\Reg_m(T)\ge\log n/\eta-1>2^{1+P(n-8)/8}\log n/(C_0n)-1$.

Finally, $n\ge C_0>16$ gives $n-8\ge n/2$, and $P\ge(m-1)/3$ for $m\ge3$, so the exponent $1+P(n-8)/8$ is at least $(m-1)n/48$.
\end{proof}

\appendix
\section{Elementary lemmas on exponential weights}\label{app:elementary}

This appendix collects the facts about exponential tilting used throughout, the proof of the potential inequality, and the regret of Optimistic Hedge against a constant loss vector. For a probability vector $p$ and a real vector $v$ of the same dimension, the variance has the variational characterization
\begin{equation}\label{eq:var}
 \Var_p(v)\;=\;\sum_ap(a)\bigl(v(a)-\ip pv\bigr)^2\;=\;\min_{c\in\R}\sum_ap(a)\bigl(v(a)-c\bigr)^2,
\end{equation}
so $\norm v_p=\sqrt{\Var_p(v)}$ is a Hilbert seminorm whose kernel is the constant vectors when $p$ has full support.

\begin{lemma}[Variance facts]\label{lem:varfacts}
Let $p,p'$ be probability vectors and $u,v$ real vectors. If $p'(a)\le\Gamma p(a)$ for all $a$, then $\Var_{p'}(v)\le\Gamma\Var_p(v)$. Moreover $\Var_p(u+v)\le2\Var_p(u)+2\Var_p(v)$, and $\Var_p(v)\le b^2/4$ whenever all entries of $v$ lie in an interval of length $b$; in particular $\Var_p(v)\le\norm v_\infty^2$.
\end{lemma}

\begin{proof}
The first claim follows from \eqref{eq:var} with $c=\ip pv$, the second from the seminorm triangle inequality and $(\alpha+\beta)^2\le2\alpha^2+2\beta^2$, and the third from \eqref{eq:var} with $c$ the midpoint of the interval.
\end{proof}

Recall that $\tilt pz$ denotes the probability vector with entries $p(a)e^{z(a)}/\sum_cp(c)e^{z(c)}$, that $\osc(z)=\max_az(a)-\min_az(a)$, and that $\norm z_p=\sqrt{\Var_p(z)}$. For probability vectors $p,q$ with $q(a)>0$ whenever $p(a)>0$, $\KL(p\,\|\,q)=\sum_{a:p(a)>0}p(a)\log\frac{p(a)}{q(a)}$.

\begin{lemma}[Exponential tilting]\label{lem:tilt}
Let $p\in\simplex_n$, let $z\in\R^n$ with $\osc(z)\le r$, and let $q=\tilt pz$. Then:
\begin{enumerate}[label=(\roman*)]
\item\label{it:density} $e^{-r}\le q(a)/p(a)\le e^{r}$ for every $a$ in the support of $p$;
\item\label{it:varcomp} $e^{-r}\Var_p(w)\le\Var_q(w)\le e^r\Var_p(w)$ for every $w\in\R^n$;
\item\label{it:movement} $\norm{\tilt pz-p}_1\le e^{r/2}\norm z_p$;
\item\label{it:lipschitz} $\norm{\tilt pz-\tilt pw}_1\le\norm{z-w}_\infty$ for every $w\in\R^n$;
\item\label{it:kl} $\tfrac12e^{-r}\Var_p(z)\le\KL(p\,\|\,\tilt pz)\le\tfrac12e^{r}\Var_p(z)$.
\end{enumerate}
\end{lemma}

\begin{proof}
\ref{it:density}: the normalizer $\sum_cp(c)e^{z(c)}$ lies between $e^{\min z}$ and $e^{\max z}$, and $q(a)/p(a)$ is $e^{z(a)}$ divided by it. \ref{it:varcomp}: by the variational formula \eqref{eq:var}, if $q\le\Gamma p$ coordinatewise then $\Var_q(w)\le\sum_aq(a)(w(a)-\ip pw)^2\le\Gamma\Var_p(w)$; apply this with $\Gamma=e^r$ and symmetrically. \ref{it:movement}: let $p_s=\tilt p{sz}$ for $0\le s\le1$. Differentiating, $\frac{d}{ds}p_s(a)=p_s(a)\bigl(z(a)-\ip{p_s}z\bigr)$, so by the Cauchy--Schwarz inequality $\norm{\frac{d}{ds}p_s}_1\le\sum_ap_s(a)|z(a)-\ip{p_s}z|\le\norm z_{p_s}\le e^{r/2}\norm z_p$, using \ref{it:varcomp} for $p_s=\tilt p{sz}$ with $\osc(sz)\le r$; integrate over $s\in[0,1]$. \ref{it:lipschitz}: apply the same computation along the segment $w+s(z-w)$, $0\le s\le1$: the derivative of $\tilt p{w+s(z-w)}$ has $\ell_1$ norm at most $\norm{z-w}_{\tilt p{w+s(z-w)}}\le\norm{z-w}_\infty$. \ref{it:kl}: let $\phi(s)=\log\sum_ap(a)e^{sz(a)}$. Then $\phi'(s)=\ip{p_s}z$, $\phi''(s)=\Var_{p_s}(z)$, and a direct computation gives $\KL(p\,\|\,\tilt pz)=\phi(1)-\phi(0)-\phi'(0)=\int_0^1(1-s)\Var_{p_s}(z)\,ds$ by Taylor's formula with integral remainder. Apply \ref{it:varcomp} to $p_s$ and use $\int_0^1(1-s)ds=\frac12$.
\end{proof}

\begin{proof}[Proof of Lemma~\ref{lem:potential}]
Throughout the proof we drop the player index and write $\ell^t=\ell_i^t$ and $x^t=x_i^t$. Let $S^t=\sum_{s\le t}\ell^s$, so that $x^t=\tilt{u}{-\eta(S^{t-1}+\ell^{t-1})}$ with $u$ the uniform distribution, and let $\Phi_t=\log\bigl(\frac1n\sum_ae^{-\eta S^t(a)}\bigr)$, so that $\Phi_0=0$. Write $Z_t=\sum_ae^{-\eta(S^{t-1}(a)+\ell^{t-1}(a))}$, so that $x^t(a)=e^{-\eta(S^{t-1}(a)+\ell^{t-1}(a))}/Z_t$. Since $S^t=(S^{t-1}+\ell^{t-1})+(\ell^t-\ell^{t-1})$ and $S^{t-1}=(S^{t-1}+\ell^{t-1})-\ell^{t-1}$,
\[
 e^{\Phi_t}=\frac{Z_t}{n}\,\E_{x^t}e^{-\eta(\ell^t-\ell^{t-1})},\qquad e^{\Phi_{t-1}}=\frac{Z_t}{n}\,\E_{x^t}e^{\eta\ell^{t-1}},
\]
and therefore $\Phi_t-\Phi_{t-1}=\log\E_{x^t}e^{-\eta(\ell^t-\ell^{t-1})}-\log\E_{x^t}e^{\eta\ell^{t-1}}$. For a probability vector $p$ and a vector $v$ let $K_p(v)=\log\E_pe^{v}-\E_pv$ be the cumulant generating function minus its linear term; a direct computation shows $K_p(v)=\KL(p\,\|\,\tilt pv)$, so by Lemma~\ref{lem:tilt}\ref{it:kl}, $\frac12e^{-\osc(v)}\Var_p(v)\le K_p(v)\le\frac12e^{\osc(v)}\Var_p(v)$. Separating the linear terms,
\[
 \eta\ip{x^t}{\ell^t}+\Phi_t-\Phi_{t-1}=K_{x^t}\bigl(-\eta(\ell^t-\ell^{t-1})\bigr)-K_{x^t}\bigl(\eta\ell^{t-1}\bigr).
\]
The two arguments have oscillation at most $2\eta$ and $\eta$, respectively. Summing over $t\le T$, using $\Phi_0=0$ and $\Phi_T\ge\log\bigl(\frac1ne^{-\eta\min_aS^T(a)}\bigr)=-\eta\min_aS^T(a)-\log n$, we obtain
\[
 \eta\Bigl(\sum_{t\le T}\ip{x^t}{\ell^t}-\min_aS^T(a)\Bigr)-\log n\;\le\;\frac{\eta^2e^{2\eta}}2D_i(T)-\frac{\eta^2e^{-\eta}}2V_i(T),
\]
and dividing by $\eta$ gives, for every $\eta>0$,
\[
 \Reg_i(T)\;\le\;\frac{\log n}{\eta}+\frac{\eta}{2}\Bigl(e^{2\eta}D_i(T)-e^{-\eta}V_i(T)\Bigr).
\]
For $\eta\le1/8$ we have $\frac12e^{2\eta}\le\frac23$ and $\frac12e^{-\eta}\ge\frac13$, which gives \eqref{eq:potential}.
\end{proof}

\begin{proposition}[Regret against a constant loss vector]\label{prop:constloss}
Let $n\ge2$ and $\eta>0$, and let a player with $n$ actions run Optimistic Hedge against the constant sequence $\ell^t=\ell$ for $t\ge1$, where $\ell(a^\ast)=0$ for one action $a^\ast$ and $\ell(a)=1$ for all $a\ne a^\ast$. Then $\Reg(T)$ is nondecreasing in $T$ and
\[
 \lim_{T\to\infty}\Reg(T)\;\ge\;\frac{\log n}{\eta}-1 .
\]
In particular, if this player is embedded as player~$3$ of a three-player game whose loss table does not depend on the actions of players~$1$ and~$2$, then $\sup_T\sup_{G\in\cG(3,n)}R(G,T,\eta)\ge\log n/\eta-1$.
\end{proposition}

\begin{proof}
With $\ell^0=0$, the first update multiplies the weights by $e^{-2\eta\ell}$ and every later update by $e^{-\eta\ell}$, so $x^t(a)\propto e^{-\eta t\ell(a)}$ for all $t\ge2$, and $x^1$ is uniform. The total mass on the suboptimal actions in round $t$ is therefore $m_t=(n-1)e^{-\eta t}/(1+(n-1)e^{-\eta t})$ for $t\ge2$ and $m_1=(n-1)/n$, and $\Reg(T)=\sum_{t\le T}m_t$ because the best action has loss zero. Each $m_t$ is positive, so $\Reg(T)$ is nondecreasing, and since $t\mapsto m_t$ is decreasing,
\[
 \sum_{t\ge1}m_t\;\ge\;\int_1^\infty\frac{(n-1)e^{-\eta u}}{1+(n-1)e^{-\eta u}}\,du\;=\;\frac1\eta\log\bigl(1+(n-1)e^{-\eta}\bigr)\;\ge\;\frac1\eta\log\bigl(ne^{-\eta}\bigr)\;=\;\frac{\log n}{\eta}-1 ,
\]
where the last inequality uses $1+(n-1)e^{-\eta}\ge ne^{-\eta}$.
\end{proof}

\section*{AI disclosure}
The proofs originated from human insight and were developed through a combination of human and AI work: most of the proofs contain a combination of ideas from the authors and OpenAI's ChatGPT. Anthropic's Claude was also used to draft and edit parts of the text. The authors verified the correctness and originality of all content, including references, and take full responsibility for the content of the paper.

\bibliographystyle{alpha}
\bibliography{refs}

\end{document}